\documentclass{article} 
\usepackage{iclr2027_conference,times}

\usepackage{amsmath,amsfonts,bm}

\def\eqref#1{equation~\ref{#1}}

\def\1{\bm{1}}

\DeclareMathAlphabet{\mathsfit}{\encodingdefault}{\sfdefault}{m}{sl}
\SetMathAlphabet{\mathsfit}{bold}{\encodingdefault}{\sfdefault}{bx}{n}

\usepackage{hyperref}
\usepackage{url}
\usepackage[utf8]{inputenc} 
\usepackage[T1]{fontenc}    
\usepackage{hyperref}       
\usepackage{url}            
\usepackage{booktabs}       
\usepackage{amsfonts}       
\usepackage{nicefrac}       
\usepackage{microtype}      
\usepackage{xcolor}         
\usepackage{multirow}
\usepackage{soul}
\usepackage{url}
\usepackage{booktabs}
\usepackage{arydshln}
\usepackage{nicefrac}
\usepackage{multirow}
\usepackage{tabularx}
\usepackage{amsmath}
\usepackage{amssymb}
\usepackage{amsfonts}
\usepackage{mathtools}
\usepackage{fontawesome}
\usepackage{tikzducks}
\usepackage{xcolor,colortbl}
\usepackage[export]{adjustbox}
\usepackage{subcaption}
\usepackage{ragged2e}
\usepackage{enumitem}
\usepackage{pdflscape}
\usepackage{makecell}

\newcommand{\model}{\texttt{Lychee-Reranker}}

\title{\href{https://huggingface.co/papers/2606.22807}{Lychee-Reranker}: Fast but Not Late Interaction for Compressed Document Reranking}

\iclrfinalcopy

\author{%
    \textbf{Xinping Zhao$^{1,2}$, Jiaxin Xu$^{1}$, Ziqi Dai$^{1}$, Xin Zhang$^{1}$, Huiyao Chen$^{1,2}$, Shouzheng Huang$^{1}$,} \\
    \textbf{Xianhao Xiong$^{3}$, Danyu Tang$^{1}$, Xinshuo Hu, Guohong Fu$^{3}$, Meishan Zhang$^{1}$, Baotian Hu$^{1,2}$\thanks{Corresponding Author}~~} \\
    $^{1}$Harbin Institute of Technology (Shenzhen);
    $^{2}$Shenzhen Loop Area Institute; $^{3}$Soochow University \\
    \texttt{zhaoxinping@stu.hit.edu.cn, mason.zms@gmail.com,} \\
    \texttt{hubaotian@hit.edu.cn, zhangmin2021@hit.edu.cn}
}

\begin{document}

\maketitle

\begin{abstract}
As retrieval systems scale, effective and efficient reranking becomes increasingly important.
However, most existing encoder- and decoder-based rerankers jointly process every query--passage pair, tightly coupling their online computation and limiting deployment efficiency and flexibility.
We present \model, a \emph{fast but not late-interaction} (FBNL) reranker that decouples query and passage computation while retaining expressive relevance modeling.
Built on an encoder--decoder architecture, \model\ pre-encodes passages using Matryoshka embedding pooling, while its decoder models system and user instructions together with query intent; cross-attention then captures fine-grained relevance between the resulting query context and passage representations.
Together, these designs offer four key advantages: (i) \emph{efficiency} from offline passage encoding, (ii) \emph{expressiveness} from cross-attention, (iii) \emph{compactness} from Matryoshka embedding pooling, and (iv) \emph{test-time compute} through an adjustable compute budget.
We instantiate \model\ in three sizes, \emph{Nano}, \emph{Small}, and \emph{Large}, with 0.27B, 1B, and 4B activated parameters, respectively.
Extensive experiments on BEIR, MIRACL, and LMEB demonstrate strong reranking performance with superior efficiency. 
On BEIR and MIRACL, our models achieve competitive performance in multi-domain and multilingual reranking, on par with strong industrial rerankers such as the Qwen3/BGE-Reranker series.
On LMEB-Dialogue, a compact embedding model paired with our Nano reranker, which has only 0.27B activated parameters, remains competitive with 7--12B embedding models.
\model\ was previously known as \texttt{KaLM-Reranker-V1}.
Data and models are available at \href{https://huggingface.co/collections/KaLM-Embedding/lychee-kalm-reranker-and-jev}{link}.
The code is available at \url{https://github.com/HITsz-TMG/Lychee-Reranker}.
\end{abstract}

\section{Introduction}
\label{sec:introduction}
At scale, neural retrieval systems typically follow a two-stage pipeline: a retriever first recalls a small candidate set from a large corpus, and a reranker then performs fine-grained relevance modeling to produce the final ranking~\citep{nogueira2019passage,karpukhin2020dense}.
The reranked passages can then be provided directly to large language models (LLMs) or further processed through context engineering~\citep{lewis2020retrieval,xu2024recomp}.
Reranking quality is therefore crucial to end-to-end performance in search~\citep{nogueira2019passage}, recommendation~\citep{DBLP:conf/ijcai/LiuXQS00ZT22,DBLP:journals/corr/abs-2412-18378}, and retrieval-augmented generation (RAG)~\citep{lewis2020retrieval,DBLP:conf/emnlp/ZhaoLZHCHZ24,chen2026beyond}.
Most existing rerankers estimate relevance by jointly processing query--passage pairs, thereby coupling query and passage computation~\citep{zhang2025qwen3,wang2025jina,chen2024bge}.
Consequently, passage representations must be recomputed for each query, preventing offline precomputation and incurring substantial online cost at scale.
Late-interaction models offer a more efficient alternative by encoding queries and passages independently~\citep{khattab2020colbert}.
However, their reliance on token-level similarity functions restricts query--passage interaction to predefined scoring operations, limiting the expressiveness of relevance modeling.
More broadly, both paradigms estimate relevance using a fixed computational budget and lack a mechanism for flexibly allocating computation during inference.
This makes it difficult to flexibly adjust test-time compute~\citep{snell2025scaling}, where additional inference-time computation progressively refines relevance modeling.
\begin{figure}[t]
    \centering
    \begin{subfigure}[t]{0.485\linewidth}
        \centering
        \includegraphics[width=\linewidth]{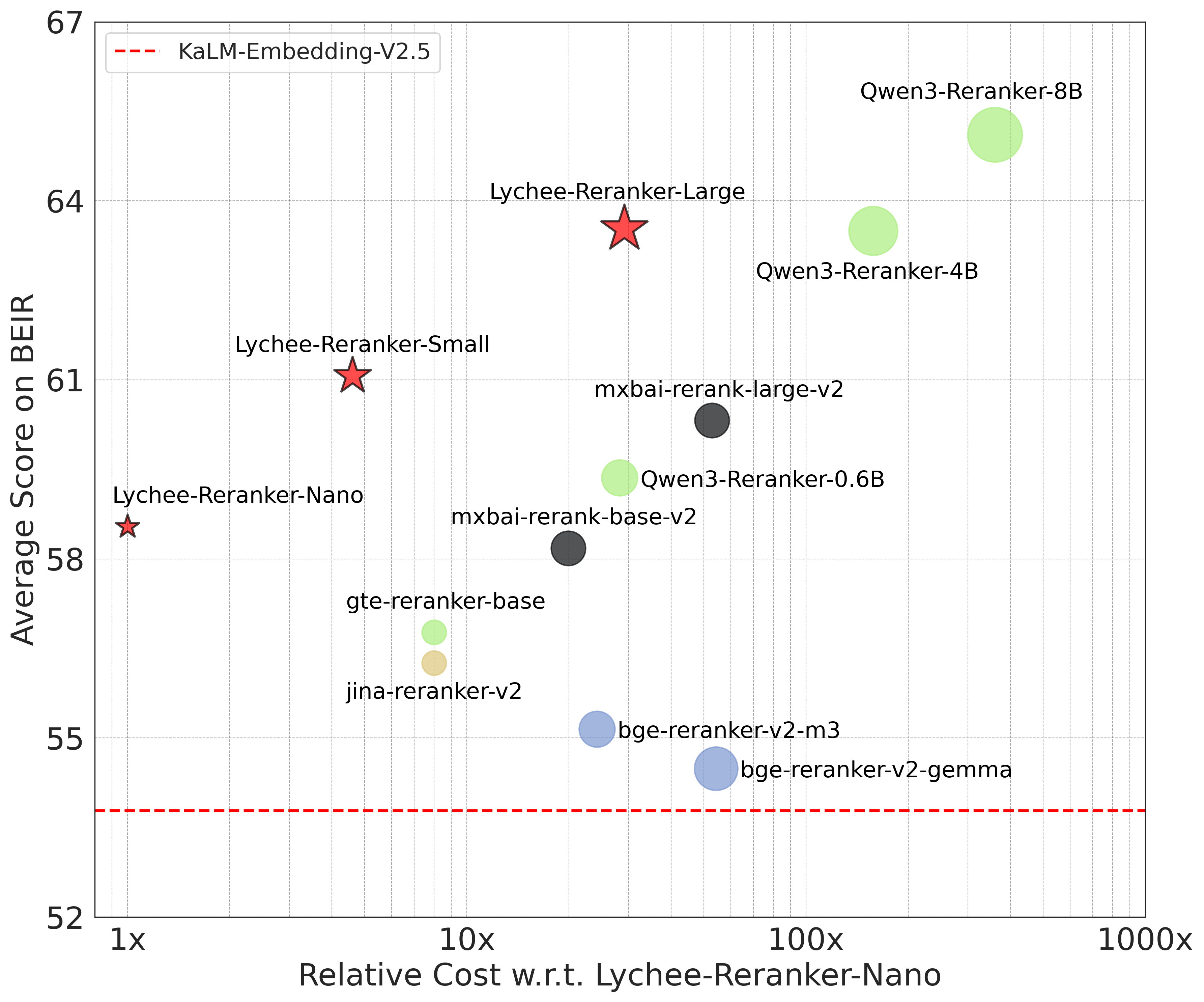}
        \caption{BEIR.}
        \label{fig:beir_scatter}
        \vspace{0.05cm}
    \end{subfigure}
    \hfill
    \begin{subfigure}[t]{0.485\linewidth}
        \centering
        \includegraphics[width=\linewidth]{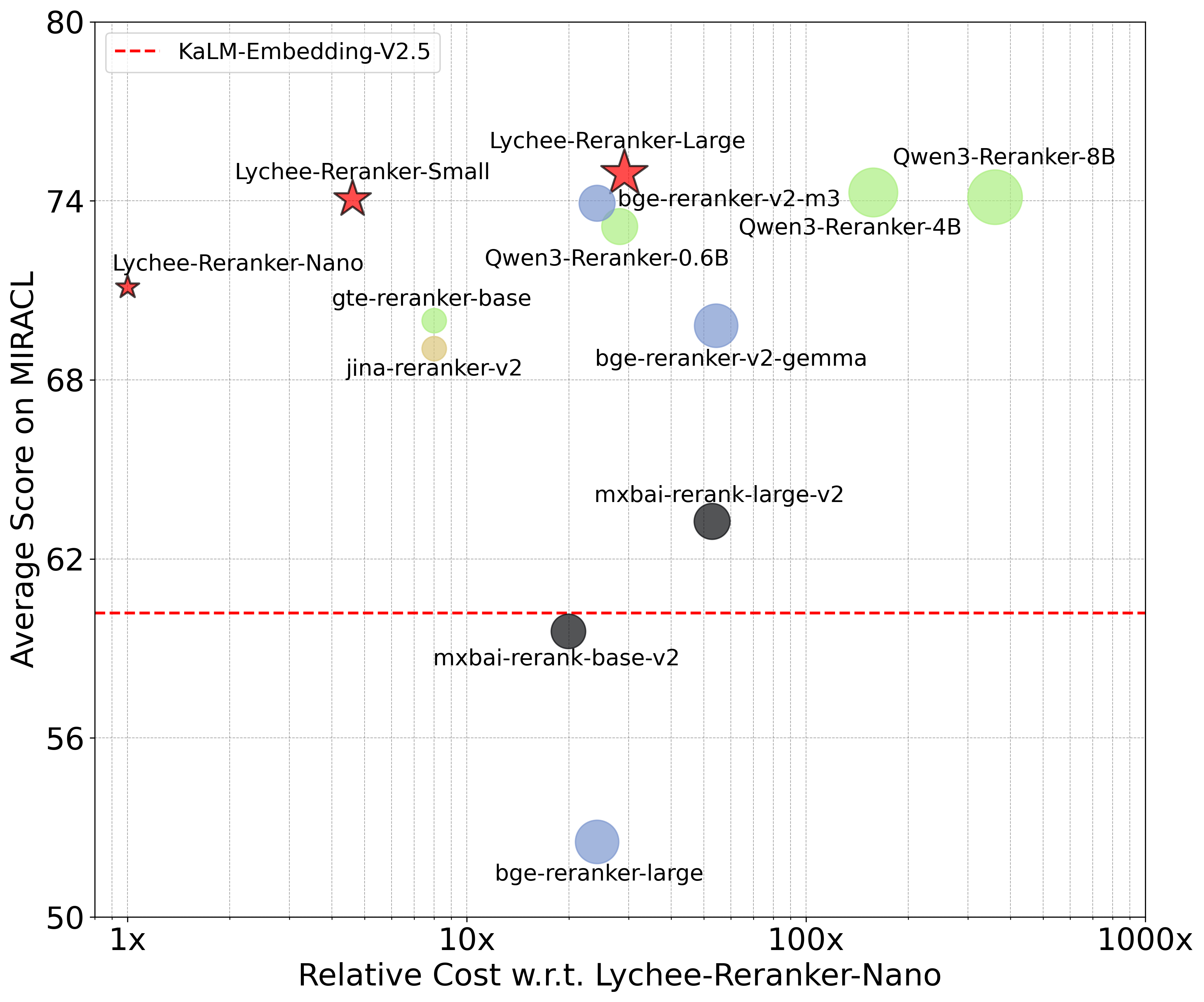}
        \caption{MIRACL.}
        \label{fig:miracle_scatter}
        \vspace{0.05cm}
    \end{subfigure}
    \caption{Comparison between the \model\ series and other competitive reranking models on BEIR and MIRACL \emph{w.r.t.} performance and cost. The cost is estimated following the analysis in Appendix~\ref{app:time_complexity}; the x-axis is plotted on a logarithmic scale. Marker sizes are proportional to model sizes, and points closer to the upper-left corner indicate better performance--cost trade--offs.}
    \label{fig:cost_perf_scatter}
    \vspace{-0.2cm}
\end{figure}
In this work, we introduce the \model\ series, which features a \emph{fast but not late-interaction} (FBNL) reranking design that decouples query and passage computation while preserving expressive relevance modeling. 
\model\ is built on an encoder--decoder architecture, where the encoder is dedicated to producing passage representations that can be precomputed offline and reused across queries. 
To further reduce storage and improve efficiency, \model\ is trained using Matryoshka embedding pooling (MEP), which produces compact passage representations while preserving their effectiveness for reranking.
By varying the MEP compression ratio, \model\ supports flexible test-time compute, allowing additional inference-time computation to progressively improve reranking quality.
The decoder takes the system instruction, user instruction, and query as inputs. 
The decoder then models relevance through merged self-attention and cross-attention over the concatenated query context and pre-encoded passage representations, enabling fine-grained query-conditioned interactions without fully re-encoding each passage for every user request.
Overall, this FBNL design offers four key advantages in practice:
\begin{itemize}[leftmargin=*]
  \item \textbf{Efficiency}: {Passage representations can be precomputed offline, reducing online computation costs.}
  \item \textbf{Expressiveness}: {The decoder supports more expressive relevance modeling than late interaction.}
  \item \textbf{Compactness}: {MEP compresses passage representations, significantly reducing storage overhead.}
  \item \textbf{Test-time compute}: {The MEP compression ratio enables flexible allocation of inference-time computation, allowing the model to achieve better reranking quality with more computation.}
\end{itemize}
We instantiate \model\ in three sizes: \emph{Nano}, \emph{Small}, and \emph{Large}, with 0.27B, 1B, and 4B activated parameters, respectively, to accommodate diverse performance and serving requirements.
Experiments on BEIR~\citep{thakur2021beir}, MIRACL~\citep{zhang2023miracl}, and LMEB~\citep{DBLP:journals/corr/abs-2603-12572} show that \model\ achieves leading reranking performance with substantially higher efficiency across multi-domain, multilingual, and long-horizon memory retrieval settings.
Specifically, as shown in Figure~\ref{fig:cost_perf_scatter}, \model\texttt{-Nano} substantially outperforms gte-reranker-base on BEIR and slightly outperforms it on MIRACL, while being about 8x more efficient.
Across MEP compression ratios from 2x to 128x, performance degrades gradually, with only a very small drop between 2x and 8x.
On LMEB-Dialogue, a compact embedding model paired with the 0.27B Nano reranker remains competitive with 7--12B embedding models.
Last but not least, our models support test-time compute: they cheaply screen many candidates (\emph{e.g.,} 64x compression), then thoroughly examine the most promising ones (\emph{e.g.,} 2x compression).
Results show that this cascaded multi-stage reranking strategy substantially improves efficiency with negligible performance loss.

\section{Related Work}
\textbf{Embedding Models.}
Text embedding models map text into continuous vectors for efficient similarity search~\citep{DBLP:conf/eacl/MuennighoffTMR23,DBLP:conf/iclr/EnevoldsenCKKMS25,zhang2025role}.
From GloVe word vectors~\citep{pennington2014glove} to transformer-based sentence embedders such as Sentence-BERT~\citep{reimers2019sentence}, dense representations have been widely adopted for semantic similarity search.
DPR~\citep{karpukhin2020dense} further demonstrated that dual-encoder passage retrieval can serve as an effective and scalable first-stage retriever for open-domain QA.
Subsequent models such as Contriever~\citep{izacard2022unsupervised}, GTR~\citep{ni2022large}, and E5~\citep{wang2022text} improved generalization through contrastive pretraining, large-scale supervision, and stronger backbones.
Recent LLM-driven embedding models have become dominant, including GTE~\citep{li2023gte,zhang2024mgte}, Qwen3-Embedding~\citep{zhang2025qwen3}, KaLM~\citep{hu2025kalmembedding,zhao2026kalmembeddingv}, BGE~\citep{chen2024bge}, Jina~\citep{sturua2024jina,akram2026jina}, NV-Embed~\citep{lee2025nv}, and F2LLM~\citep{zhang2026f2llm}.
Despite the strong performance of LLM-based embeddings, these models typically adopt a bi-encoder architecture, where queries and documents are encoded separately and matched only through vector similarity.
This design limits fine-grained query--document interaction, making rerankers necessary to further refine the retrieved candidates.

\textbf{Reranking Models.}
Reranking addresses the limited interaction in embedding retrieval by performing deeper query--document relevance modeling.
Early neural rerankers such as BERT-based rerankers concatenate the query and passage and apply full self-attention for relevance modeling~\citep{nogueira2019passage}.
Generative and LLM-based rerankers, including RankGPT~\citep{sun2023chatgpt}, RankVicuna~\citep{pradeep2023rankvicuna}, BGE~\citep{chen2024bge}, Qwen3-Reranker~\citep{zhang2025qwen3}, Jina-Reranker~\citep{wang2025jina}, and GMR~\citep{zhang2026supervised}, further strengthen this paradigm through generative scoring, instruction tuning, and stronger backbones.
Despite their effectiveness, these methods tightly couple query and passage computation, requiring each candidate passage to be processed jointly with each query.
In parallel, late-interaction methods such as ColBERT~\citep{khattab2020colbert} and ColBERTv2~\citep{santhanam2022colbertv2} decouple query and passage encoding while preserving token-level matching.
They improve efficiency, but their interaction is largely limited to similarity aggregation between independently contextualized query and passage token representations.
Overall, reranking methods have developed along multiple directions, including full interaction, generative scoring, and efficient late interaction.
However, prevailing models often face a trade-off between efficiency and effectiveness~\citep{zhang2025qwen3,wang2025jina}.
Our models introduce a new reranking paradigm, fast but not late interaction (FBNL), that offers a more favorable trade-off between efficiency and effectiveness: the encoder produces reusable passage representations, while the decoder performs fine-grained query--passage relevance modeling.

\section{Model Architecture}
\label{sec:archi}

\subsection{Architecture}
\begin{figure}[t]
    \centering
    \includegraphics[width=1.0\linewidth]{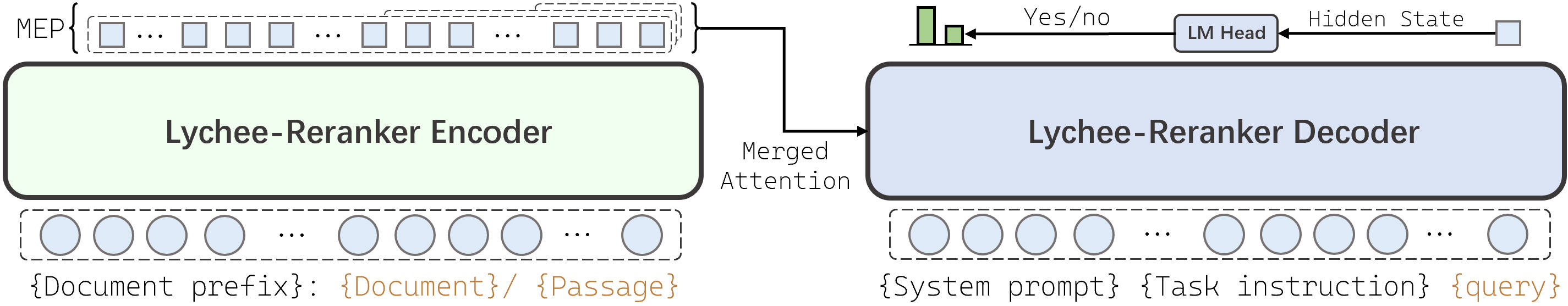}
    \caption{Overview of \model. The encoder produces reusable passage representations compressed by MEP, while the decoder computes accurate relevance scores via merged attention.}
    \label{fig:model_framework}
\end{figure}

\model\ is built on the T5Gemma2 model family~\citep{DBLP:journals/corr/abs-2512-14856} and instantiated at three encoder--decoder scales: 270M--270M, 1B--1B, and 4B--4B.
We initialize the models from T5Gemma2 to inherit its pretrained language modeling capabilities while adopting its encoder--decoder architecture for efficient document reranking.
Table~\ref{tab:model-spec} summarizes the model specifications.
Figure~\ref{fig:model_framework} illustrates our model architecture, which is designed to address key limitations of existing rerankers.
Concretely, existing rerankers~\citep{chen2024bge,wang2025jina,zhang2025qwen3} tightly couple query and document computation, requiring joint processing of each query--document pair and resulting in high serving cost at scale.
Late-interaction methods such as ColBERT~\citep{khattab2020colbert} improve efficiency by encoding queries and documents separately and then performing multi-vector interaction, but cannot capture fine-grained query--document interactions during scoring.
Our \emph{fast but not late-interaction} (FBNL) paradigm differs from both conventional encoder- and decoder-based rerankers and late-interaction models.
Unlike the former, it reuses document representations to improve serving efficiency; unlike the latter, it preserves rich query--document interaction through decoder cross-attention rather than restricting interaction to post-encoding scoring.
These designs, together with Matryoshka embedding pooling in \S\ref{sec:training}, enable efficient and expressive reranking at scale without sacrificing rich query--document interaction.
Formally, given an instruction $I$, a query $q$, and a candidate document $p$, the encoder maps $p$ to a reusable representation:
\begin{equation}
    \mathbf{H}_p = \operatorname{Enc}(p) \in \mathbb{R}^{n \times d},
\end{equation}
where $n$ denotes the document sequence length and $d$ is the hidden dimension.
The decoder takes the system instruction (omitted from the equation for brevity), the task-specific instruction $I$, and the user query $q$ as input. It then attends to $\mathbf{H}_p$ through merged attention to produce vocabulary logits:
\begin{equation}
        \mathbf{Z} = \operatorname{Dec}(I, q; \mathbf{H}_p) \in \mathbb{R}^{|\mathcal{V}|},
\end{equation}
where $\mathcal{V}$ denotes the token vocabulary.
During training and inference, we use \texttt{logits\_to\_keep} to retain only the logits of the final token, substantially reducing GPU memory usage.
More specifically, let $\mathbf{X} \in \mathbb{R}^{m \times d}$ denote the decoder input.
The decoder projects $\mathbf{X}$ into queries and projects the concatenation of the decoder input $\mathbf{X}$ and the encoder output $\mathbf{H}_p$ into keys and values, as follows:
\begin{align}
    \mathbf{Q} = \mathbf{X}\mathbf{W}_Q, \quad
    \mathbf{K} = [\mathbf{X}; \mathbf{H}_p]\mathbf{W}_K, \quad
    \mathbf{V} = [\mathbf{X}; \mathbf{H}_p]\mathbf{W}_V,
\end{align}
where $\mathbf{W}_Q$, $\mathbf{W}_K$, and $\mathbf{W}_V$ denote the projection matrices for queries, keys, and values, respectively.
The merged self-attention and cross-attention output is then computed as:
\begin{equation}
    \mathbf{O}
    = \operatorname{softmax}
      \left(\frac{\mathbf{Q}\mathbf{K}^{\top}}{\sqrt{d_h}} + \mathbf{M}\right)\mathbf{V}\mathbf{W}_O,
\end{equation}
where $d_h$ is the attention head dimension, $\mathbf{M} \in \mathbb{R}^{m \times (m+n)}$ is the attention mask, and $\mathbf{W}_O$ denotes the output projection matrix.
The language-model head projects the decoder hidden state at the first prediction position into a vector of vocabulary logits $\mathbf{Z} \in \mathbb{R}^{|\mathcal{V}|}$.
To compute the relevance score, we compare the next-token likelihoods of \texttt{yes} and \texttt{no}.
We use only the two entries in $\mathbf{Z}$ corresponding to these label tokens, denoted by $z_{\mathrm{yes}}$ and $z_{\mathrm{no}}$, respectively.
We define the relevance score as follows:
\begin{equation}
    \text{score}(q, p)
    = \frac{\exp(z_{\mathrm{yes}})}
           {\exp(z_{\mathrm{yes}}) + \exp(z_{\mathrm{no}})}.
\end{equation}
Here, $\text{score}(q, p)$ denotes the relevance score of document $p$ with respect to query $q$.
Appendix~\ref{app:complexity} presents a comprehensive time- and space-complexity analysis comparing our encoder--decoder reranker, \model, with conventional encoder-based and decoder-based rerankers.

\begin{table}[t]
\centering
\caption{The \model\ series. Activated parameters denote the number of parameters activated during online inference, excluding the offline encoder. ``\#Layers'' is the number of decoder layers;  values in parentheses give the total encoder--decoder layer count used as $L$ in Eq.~\ref{equ:enc_dec_time}. Document token dimension is the size of each encoded document token. MEP denotes Matryoshka embedding pooling, with supported compression ratios. ``Instruction Aware'' indicates support for task-specific instructions. ``Test-Time Compute'' is support for on-demand allocation of compute during inference.}
\label{tab:model-spec}
\resizebox{1.0\textwidth}{!}{%
\begin{tabular}{l|c|c|c|cccccc}
\toprule
\multicolumn{1}{c|}{Models} & \begin{tabular}[c]{@{}c@{}}Activated \\ Params.\end{tabular} & \begin{tabular}[c]{@{}c@{}}Non-Embedding \\ Params.\end{tabular} & \begin{tabular}[c]{@{}c@{}}Embedding \\ Params.\end{tabular} & \#Layers & \begin{tabular}[c]{@{}c@{}}Sequence \\ Length\end{tabular} & \begin{tabular}[c]{@{}c@{}}Document \\ Token Dim.\end{tabular} & \begin{tabular}[c]{@{}c@{}}MEP \\ Support\end{tabular} & \begin{tabular}[c]{@{}c@{}}Instruction \\ Aware\end{tabular} & \begin{tabular}[c]{@{}c@{}}Test-Time \\ Compute\end{tabular} \\
\midrule
\href{https://huggingface.co/KaLM-Embedding/KaLM-Reranker-V1-Nano-R2}{\model\texttt{-Nano}} & 0.27B & 100M & 168M & 18 (36) & 128K & 640 & 1x---128x & Yes & Yes \\
\href{https://huggingface.co/KaLM-Embedding/KaLM-Reranker-V1-Small-R2}{\model\texttt{-Small}} & 1B & 698M & 302M & 26 (52) & 128K & 1152 & 1x---128x & Yes & Yes \\
\href{https://huggingface.co/KaLM-Embedding/KaLM-Reranker-V1-Large-R2}{\model\texttt{-Large}} & 4B & 3209M & 675M & 34 (68) & 128K & 2560 & 1x---128x & Yes & Yes \\
 \bottomrule
\end{tabular}%
}
\vspace{0.05cm}
\vspace{-0.3cm}
\end{table}

\begin{table}[t]
\centering
\caption{Prompt templates for \model. The encoder takes a document and produces reusable representations; the decoder takes an instruction and query to compute the reranking score.}
\label{tab:prompt_template_for_enc_dec}
\footnotesize
\setlength{\tabcolsep}{6pt}
\renewcommand{\arraystretch}{0.935}
\resizebox{1.0\textwidth}{!}{
\begin{tabular}{p{0.96\textwidth}}
\toprule
\multicolumn{1}{c}{\textbf{Prompt Templates for \model}} \\
\midrule
\rowcolor{gray!10}
\textbf{Prompt Template for Encoder} \\
\addlinespace[1pt]
\texttt{<Document>: \textcolor{brown}{\{Document\}}} \\
\midrule
\rowcolor{gray!10}
\textbf{Prompt Template for Decoder} \\
\addlinespace[1pt]
\texttt{<bos><start\_of\_turn>\textcolor{blue}{user}} \\
\texttt{Judge whether the Document meets the requirements based on the Query and the Instruct provided. Note that the answer can only be ``yes'' or ``no''.} \\
\texttt{} \\
\texttt{<Instruct>: \textcolor{brown}{\{Instruction\}}} \\
\texttt{<Query>: \textcolor{brown}{\{Query\}}<end\_of\_turn>} \\
\texttt{<start\_of\_turn>\textcolor{red}{model}\textbackslash{}n\textbackslash{}n\textbackslash{}n\textbackslash{}n} \\
\bottomrule
\end{tabular}
}
\vspace{0.05cm}
\vspace{-0.4cm}
\end{table}

\subsection{Prompt Templates}
Table~\ref{tab:prompt_template_for_enc_dec} summarizes the encoder and decoder prompt templates used by \model.
The encoder receives only the \textcolor{brown}{\bf document} and produces reusable passage representations that can be stored and shared across queries.
Because T5Gemma2 does not support a system role~\citep{google2025gemmaprompt}, we include the system prompt (\emph{i.e.,} ``\texttt{Judge whether ...}'') in the \textcolor{blue}{\bf user} message, together with the \textcolor{brown}{\bf instruction} and \textcolor{brown}{\bf query}.
The \textcolor{red}{\bf model} response then predicts whether the encoded document satisfies the specified requirements.
This separation follows the FBNL design: document representations remain reusable, while the decoder performs instruction- and query-aware interaction through merged self- and cross-attention during scoring.
The decoder output is restricted to the binary label tokens ``\texttt{yes}'' and ``\texttt{no}'', whose logits are used to compute the relevance score.
Instructions are provided in \S\ref{app:inst}.

\section{Model Training}
\label{sec:training}
Appendix~\ref{app:training_data} provides details on the training data, including data sources and processing procedures.
\subsection{Training Objective}
\textbf{Supervised Fine-Tuning (SFT).} Given an instruction $I$, a query $q$, and a document $p$, \model\ predicts a binary relevance label $l \in \{\texttt{yes}, \texttt{no}\}$.
Following the scoring method in \S\ref{sec:archi}, we define the SFT objective as the negative log-likelihood of the ground-truth label:
\begin{equation}
\label{equ:sft}
    \mathcal{L}_{\mathrm{sft}}(I,q,p,l)
    = - \log
    \frac{\exp(z_l)}
         {\exp(z_{\mathrm{yes}}) + \exp(z_{\mathrm{no}})} ,
\end{equation}
where $z_l$ is the logit of the ground-truth label token $l$, with $l=\texttt{yes}$ for relevant documents and $l=\texttt{no}$ for irrelevant ones.
Though reusable passage representations improve serving efficiency, storing the full representation $\mathbf{H}_p \in \mathbb{R}^{n \times d}$ for every passage in a large corpus can incur substantial storage overhead.
To reduce this overhead, we introduce \textbf{Matryoshka Embedding Pooling} (MEP), which compresses passage representations along the sequence dimension while preserving reranking effectiveness.
For a compression ratio $r$, MEP partitions the passage tokens into consecutive groups of at most $r$ tokens and applies mean pooling to each group, yielding a compact token representation:
\begin{equation}
    \mathbf{H}^{(r)}_p[j]
    = \operatorname{MeanPool}\left(
        \mathbf{H}_p[(j-1)r+1:\min(jr,n)]
      \right),
    \quad j=1,\ldots,\left\lceil \frac{n}{r} \right\rceil,
\end{equation}
where $\mathbf{H}^{(r)}_p \in \mathbb{R}^{\lceil n/r \rceil \times d}$ denotes the compressed passage representation and $\lceil\cdot\rceil$ is the ceiling function.
Having established $\mathbf{H}^{(r)}_p$, the decoder computes the relevance score at compression ratio $r$ as follows:
\begin{equation}
    \text{score}^{(r)}(q,p)
    = \frac{\exp(z^{(r)}_{\mathrm{yes}})}
           {\exp(z^{(r)}_{\mathrm{yes}}) + \exp(z^{(r)}_{\mathrm{no}})} .
\end{equation}
During training, we jointly optimize the SFT loss over a set of compression ratios, \emph{e.g.,} $\mathcal{R}=\{2, 8, 32, 128\}$, where $r=1$ corresponds to the uncompressed representation.
The final objective is:
\begin{equation}
\label{equ:sft1}
    \mathcal{L}_{\mathrm{sft}}(I,q,p,l,\mathcal{R})
    = \sum_{r \in \mathcal{R}} \lambda_r
      \left(
      - \log
      \frac{\exp(z^{(r)}_l)}
           {\exp(z^{(r)}_{\mathrm{yes}}) + \exp(z^{(r)}_{\mathrm{no}})}
      \right),
\end{equation}
where $\lambda_r$ controls the loss weight and is set to $1$ by default. This training objective encourages the model to learn robust passage representations that remain effective under different compression ratios.
\textbf{Knowledge Distillation (KD).} We use a stronger teacher reranker to provide soft supervision~\citep{hinton2015distilling,DBLP:journals/tmm/RaoMDQLZT24}. 
Concretely, for each query--document pair $(q,p)$, the teacher produces a soft relevance label $y \in [0,1]$, while the student predicts a score $\hat{y}=\text{score}(q,p)$.
To this end, we minimize the binary cross-entropy between the teacher's soft label and the student's prediction:
\begin{equation}
\label{equ:kd}
    \mathcal{L}_{\mathrm{kd}}
    = \sum\nolimits_{r \in \mathcal{R}} \lambda_r (- y \log \hat{y}^{(r)} - (1-y)\log(1-\hat{y}^{(r)})),
\end{equation}
where $y$ is the teacher's soft label and $\hat{y}^{(r)}$ is the student's prediction at compression ratio $r$. Teacher label construction is defined in \S\ref{app:training_data}.
Compared with the hard binary labels used in Equation~\ref{equ:sft1}, soft labels preserve fine-grained relevance signals and mitigate the impact of potential false hard negatives.
\begin{figure}[t]
    \centering
    \includegraphics[width=0.87\linewidth]{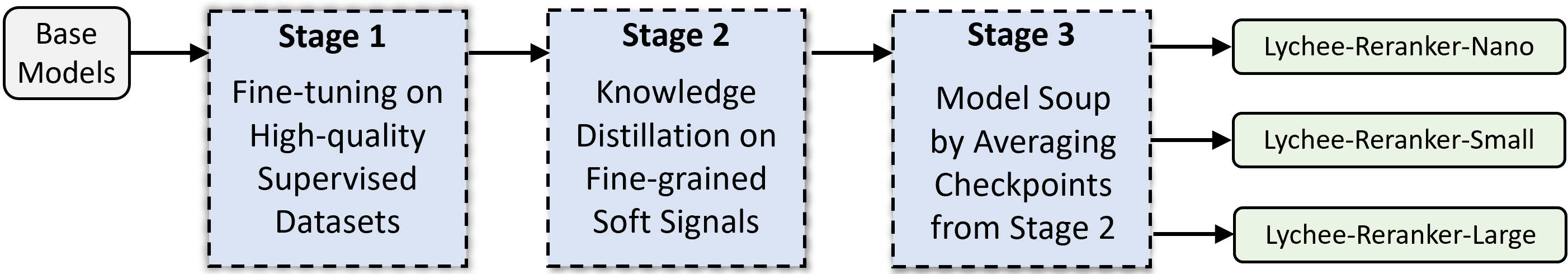}
    \caption{Multi-stage training pipeline of the \model\ series.}
    \label{fig:multi_stage_training}
    \vspace{-0.1cm}
\end{figure}
\subsection{Multi-stage Training}
\label{sec:multi_stage}
Multi-stage training is widely used to improve the generalization and robustness of retrieval models across diverse data distributions~\citep{zhang2025qwen3,akram2026jina,zhao2026kalmembeddingv,zhang2026f2llm}.
To progressively enhance the model's reranking capabilities, we employ the three-stage training pipeline shown in Figure~\ref{fig:multi_stage_training}: \textbf{(1)} fine-tuning, \textbf{(2)} knowledge distillation, and \textbf{(3)} model soup.
\textbf{Fine-tuning on High-Quality Supervised Datasets.}
First, we fine-tune the model on high-quality supervised reranking datasets. Using the SFT objective in Eq.~\ref{equ:sft1}, the model learns from binary hard labels across multiple compression ratios, establishing a robust foundation for document reranking.
\textbf{Knowledge Distillation with Fine-Grained Soft Signals.}
Next, rather than continuing to rely on coarse-grained hard labels, the model distills fine-grained soft knowledge from a strong teacher reranker (\textit{i.e.,} Qwen3-Reranker-8B~\citep{zhang2025qwen3}). The KD objective in Eq.~\ref{equ:kd} enables the student model to capture nuanced semantic differences between positive passages and hard negatives.

\textbf{Model Soup Using Stage-2 Checkpoints.}
Finally, we merge the LoRA adapters into their respective backbones and average the resulting Stage~2 model checkpoints trained with different LoRA configurations. The averaged checkpoint serves as the final model. More details are provided in \S\ref{app:imple}.

\section{Evaluation}
\label{sec:eval}

\subsection{Experimental Settings}
Experimental settings, training and implementation details, and task-specific evaluation instructions for \model\ are provided in Appendices~\ref{app:setup}, \ref{app:imple}, and~\ref{app:inst}, respectively.
Unless otherwise specified, \model\ uses a default compression ratio of $r=4$ in all experiments.
\begin{table*}[t]
    \centering
    \caption{BEIR~\citep{thakur2021beir} reranking results measured by nDCG@10. The abbreviations for the 13 tasks are defined in Table~\ref{tab:inst_beir}. ``N'', ``S'', and ``L'' are abbreviations for Nano, Small, and Large, respectively. Avg. denotes the average performance across the 13 tasks. Cost denotes the relative online computation cost estimated using the time-complexity analysis in Appendix~\ref{app:time_complexity}. The estimate assumes $|q|=32$, $n=1024$, $K=1$, and $r=4$, with model-specific $L$ and $d$ taken from Tables~\ref{tab:model-spec} and~\ref{tab:model_links}. Costs are normalized relative to \model\texttt{-Nano} (1.0x). Within each parameter group, the best results are \textbf{boldfaced}, and the second-best results are \underline{underlined}.}
    \label{tab:main_beir_results}
    \renewcommand\arraystretch{1.15}
    \setlength{\tabcolsep}{3pt}
    \resizebox{\textwidth}{!}{\begin{tabular}{lccc|ccccccccccccc}
        \toprule
        \multicolumn{1}{l}{\textbf{Models}} &
        \multicolumn{1}{c}{\textbf{Size}} &
        \multicolumn{1}{c}{\textbf{Cost}} &
        \multicolumn{1}{c|}{\textbf{Avg.}} &
        \multicolumn{1}{c}{\textbf{AA}} &
        \multicolumn{1}{c}{\textbf{CF}} &
        \multicolumn{1}{c}{\textbf{CQA}} &
        \multicolumn{1}{c}{\textbf{DB}} &
        \multicolumn{1}{c}{\textbf{FV}} &
        \multicolumn{1}{c}{\textbf{FQA}} &
        \multicolumn{1}{c}{\textbf{HQA}} &
        \multicolumn{1}{c}{\textbf{MS}} &
        \multicolumn{1}{c}{\textbf{NF}} &
        \multicolumn{1}{c}{\textbf{NQ}} &
        \multicolumn{1}{c}{\textbf{QR}} &
        \multicolumn{1}{c}{\textbf{SD}} &
        \multicolumn{1}{c}{\textbf{SF}} \\
        \midrule
        \multicolumn{17}{c}{\textbf{First-stage Retriever}} \\ \midrule
         KaLM-Embedding-V2.5 & 0.5B & -- & 53.78 & 59.60 & 32.53 & 45.54 & 42.62 & 83.72 & 46.99 & 70.74 & 40.15 & 36.95 & 57.53 & 88.88 & 19.88 & 73.98 \\ \midrule
         \multicolumn{17}{c}{\textbf{Second-stage Reranker}} \\ \midrule
        \rowcolor{gray!10}
        \multicolumn{17}{l}{\emph{Models with more than 4B parameters}} \\
        Qwen3-Reranker-8B & 8B & 359.8x & 65.11 & 82.92 & 49.63 & 52.49 & 56.07 & 94.96 & 59.72 & 84.24 & 46.84 & 41.71 & 77.13 & 89.88 & 28.97 & 81.85 \\
        \midrule
        \rowcolor{gray!10}
        \multicolumn{17}{l}{\emph{Models with 1B--4B parameters}} \\
        Qwen3-Reranker-4B & 4B & 157.8x & \underline{63.50} & \textbf{79.85} & \textbf{49.92} & \underline{49.94} & \textbf{54.59} & \textbf{94.46} & \underline{56.69} & \underline{83.68} & 44.47 & \underline{41.57} & \underline{74.03} & 88.40 & \textbf{27.14} & 80.73 \\
        \model\texttt{-L} & 4B & \textbf{29.1x} & \textbf{63.53} & \underline{72.90} & \underline{47.94} & \textbf{52.36} & \underline{52.73} & \underline{94.22} & \textbf{59.15} & \textbf{84.23} & \underline{47.38} & \textbf{42.35} & \textbf{75.61} & \textbf{90.72} & \underline{24.89} & \textbf{81.42} \\
        bge-reranker-v2-gemma & 2.5B & 54.2x & 54.49 & 69.34 & 29.60 & 37.28 & 43.02 & 82.87 & 47.21 & 82.09 & 45.38 & 28.59 & 68.80 & 86.64 & 18.86 & 68.66 \\
        mxbai-rerank-large-v2 & 1.5B & {52.8x} & 60.32 & 71.43 & 46.54 & 45.90 & 49.74 & 93.74 & 50.75 & 82.66 & \textbf{47.65} & 37.59 & 70.77 & \underline{89.56} & 16.99 & \underline{80.86} \\
        \midrule
        \rowcolor{gray!10}
        \multicolumn{17}{l}{\emph{Models with 0.5B--1B parameters}} \\
        \model\texttt{-S} & 1B & \textbf{4.6x} & \textbf{61.07} & \underline{69.43} & \underline{45.63} & \underline{47.06} & \textbf{49.77} & \underline{93.36} & \textbf{53.62} & \textbf{83.55} & 46.90 & \textbf{39.89} & \textbf{73.14} & \textbf{90.77} & \textbf{22.15} & \underline{78.72} \\
        bge-reranker-large & 0.6B & 24.2x  & 51.86 & 28.47 & 41.51 & 38.89 & 48.76 & 93.01 & 41.41 & \underline{83.12} & \underline{46.96} & 34.14 & 52.46 & 73.84 & 17.23 & 74.41\\
        bge-reranker-v2-m3 & 0.6B & 24.2x & 55.15 & 41.73 & 37.41 & 39.09 & 47.77 & 90.15 & 45.07 & 82.62 & \textbf{47.79} & 34.54 & \underline{69.25} & \underline{89.14} & 18.04 & 74.35 \\
        Qwen3-Reranker-0.6B & 0.6B & 28.2x & \underline{59.36} & \textbf{73.13} & \textbf{48.84} & \textbf{47.21} & \underline{49.43} & \textbf{93.45} & \underline{47.99} & 81.19 & 42.35 & \underline{39.32} & 64.04 & 84.12 & \textbf{22.15} & 78.46 \\
        mxbai-rerank-base-v2 & 0.5B & 19.9x & 58.18 & 63.41 & 43.38 & 45.79 & 48.97 & 92.71 & 46.43 & 81.45 & 46.79 & 38.30 & 66.83 & 88.14 & 14.93 & \textbf{79.16} \\
        \midrule
        \rowcolor{gray!10}
        \multicolumn{17}{l}{\emph{Models with fewer than 0.5B parameters}} \\
        gte-reranker-base & 0.3B & 8.0x & \underline{56.77} & \underline{58.44} & \textbf{47.04} & 39.21 & 47.31 & \textbf{94.25} & 45.35 & \underline{81.48} & 45.44 & 36.61 & 65.58 & 81.83 & 17.74 & \underline{77.71} \\
        jina-reranker-v2 & 0.3B & 8.0x & 56.26 & 51.59 & 34.40 & \textbf{40.85} & \textbf{49.03} & 92.35 & \underline{46.10} & 79.73 & \textbf{47.71} & \underline{37.12} & \underline{67.16} & \underline{87.80} & \underline{20.01} & 77.50 \\
        \model\texttt{-N} & 0.27B & \textbf{1.0x} & \textbf{58.54} & \textbf{63.97} & \underline{43.25} & \underline{40.84} & \underline{47.93} & \underline{92.95} & \textbf{47.09} & \textbf{82.58} & \underline{45.64} & \textbf{38.72} & \textbf{69.30} & \textbf{89.92} & \textbf{20.80} & \textbf{78.03} \\
        \bottomrule
    \end{tabular}}
    \vspace{-0.3cm}
\end{table*}
\begin{table*}[t]
    \centering
    \caption{MIRACL~\citep{zhang2023miracl} reranking results measured by nDCG@10 across all 18 languages. The language abbreviations are defined in Table~\ref{tab:inst_miracle}. Other settings follow those in Table~\ref{tab:main_beir_results}.}
    \label{tab:main_miracle_results}
    \renewcommand\arraystretch{1.15}
    \setlength{\tabcolsep}{2pt}
    \resizebox{\textwidth}{!}{\begin{tabular}{lccc|*{18}{c}}
        \toprule
        \textbf{Models} & \textbf{Size} & \textbf{Cost} & \textbf{Avg.} &
        \textbf{ar} & \textbf{bn} & \textbf{de} & \textbf{en} & \textbf{es} & \textbf{fa} & \textbf{fi} & \textbf{fr} & \textbf{hi} &
        \textbf{id} & \textbf{ja} & \textbf{ko} & \textbf{ru} & \textbf{sw} & \textbf{te} & \textbf{th} & \textbf{yo} & \textbf{zh} \\
        \midrule
        \multicolumn{22}{c}{\textbf{First-stage Retriever}} \\ \midrule
        KaLM-Embedding-V2.5 & 0.5B & -- & 60.19 &
        69.32 & 63.65 & 53.05 & 53.40 & 51.32 & 49.55 & 70.98 & 52.14 & 44.90 &
        49.89 & 61.37 & 58.03 & 57.87 & 66.54 & 74.53 & 70.19 & 79.20 & 57.48 \\ \midrule
        \multicolumn{22}{c}{\textbf{Second-stage Reranker}} \\ \midrule
        \rowcolor{gray!10}
        \multicolumn{22}{l}{\emph{Models with more than 4B parameters}} \\
        Qwen3-Reranker-8B & 8B & 359.8x & 74.13 &
        81.94 & 83.78 & 64.87 & 67.21 & 61.21 & 63.84 & 82.50 & 63.30 & 66.00 &
        59.17 & 80.01 & 75.05 & 78.87 & 78.95 & 86.20 & 86.66 & 87.47 & 67.30 \\
        \midrule
        \rowcolor{gray!10}
        \multicolumn{22}{l}{\emph{Models with 1B--4B parameters}} \\
        Qwen3-Reranker-4B & 4B & 157.8x & \underline{74.28} &
        \underline{81.35} & \underline{84.94} & \underline{65.45} & \underline{66.19} & \underline{61.25} & \textbf{65.37} & \underline{82.67} & \underline{63.39} & \textbf{66.47} &
        \textbf{60.16} & \underline{79.94} & \underline{73.66} & \underline{77.89} & \underline{78.59} & \textbf{87.56} & \textbf{86.58} & \textbf{87.42} & \textbf{68.20} \\
        \model\texttt{-L} & 4B & \textbf{29.1x} & \textbf{74.92} &
        \textbf{82.56} & \textbf{85.80} & \textbf{65.75} & \textbf{67.65} & \textbf{63.76} & \underline{65.09} & \textbf{83.38} & \textbf{65.99} & \underline{66.30} &
        \underline{59.17} & \textbf{80.27} & \textbf{74.45} & \textbf{79.51} & \textbf{80.56} & \underline{87.53} & \underline{85.96} & \underline{87.13} & \underline{67.71} \\
        bge-reranker-v2-gemma & 2.5B & 54.2x & 69.82 &
        75.43 & 80.28 & 62.27 & 63.53 & 60.42 & 60.30 & 77.45 & 61.91 & 62.32 &
        57.26 & 72.34 & 70.42 & 71.51 & 74.70 & 83.10 & 79.37 & 80.85 & 63.36 \\
        mxbai-rerank-large-v2 & 1.5B & {52.8x} & 63.26 &
        74.50 & 65.53 & 50.08 & 61.42 & 55.82 & 52.33 & 72.42 & 56.15 & 45.44 &
        53.50 & 71.92 & 66.96 & 69.97 & 63.75 & 57.90 & 79.35 & 79.06 & 62.50 \\
        \midrule
        \rowcolor{gray!10}
        \multicolumn{22}{l}{\emph{Models with 0.5B--1B parameters}} \\
        \model\texttt{-S} & 1B & \textbf{4.6x} & \textbf{74.06} &
        \underline{80.92} & \textbf{85.69} & \textbf{64.22} & \textbf{66.45} & 63.53 & 62.95 & \textbf{82.09} & \textbf{65.29} & 65.65 &
        59.48 & \textbf{78.91} & 72.84 & \textbf{78.13} & \underline{80.12} & \underline{87.39} & \textbf{85.53} & 86.34 & \underline{67.56} \\
        bge-reranker-large & 0.6B & 24.2x & 52.52 &
        55.24 & 62.45 & 44.94 & 50.45 & 42.47 & 42.96 & 68.89 & 46.47 & 38.00 &
        37.96 & 42.19 & 46.41 & 44.29 & 57.98 & 70.03 & 64.18 & 75.16 & 55.26 \\
        bge-reranker-v2-m3 & 0.6B & 24.2x & \underline{73.92} &
        \textbf{81.45} & \underline{84.02} & 63.77 & 63.52 & \underline{64.52} & \underline{64.24} & \underline{82.04} & \underline{64.02} & \underline{66.26} &
        \textbf{61.87} & \underline{78.51} & \underline{73.19} & \underline{76.68} & \textbf{80.41} & \textbf{88.30} & 84.53 & \underline{87.34} & 65.83 \\
        Qwen3-Reranker-0.6B & 0.6B & 28.2x & 73.14 &
        79.84 & 80.54 & \underline{63.81} & \underline{64.75} & \textbf{64.79} & \textbf{65.28} & 80.16 & 62.98 & \textbf{67.27} &
        \underline{61.24} & 76.23 & \textbf{76.58} & 74.50 & 70.25 & 85.81 & \underline{84.78} & \textbf{87.35} & \textbf{70.30} \\
        mxbai-rerank-base-v2 & 0.5B & 19.9x & 59.58 &
        73.65 & 57.16 & 29.12 & 60.86 & 56.97 & 47.74 & 69.56 & 55.81 & 42.68 &
        54.26 & 71.21 & 64.36 & 66.54 & 60.51 & 49.32 & 76.98 & 75.34 & 60.43 \\
        \midrule
        \rowcolor{gray!10}
        \multicolumn{22}{l}{\emph{Models with fewer than 0.5B parameters}} \\
        gte-reranker-base & 0.3B & 8.0x & \underline{69.99} &
        \underline{77.48} & 79.34 & 57.21 & \underline{63.16} & \underline{59.80} & 59.41 & \underline{77.77} & \underline{59.19} & \underline{60.67} &
        57.61 & \underline{73.67} & \underline{73.45} & \underline{72.05} & 74.54 & \textbf{85.78} & 78.93 & \textbf{83.88} & \textbf{65.92} \\
        jina-reranker-v2 & 0.3B & 8.0x & 69.05 &
        73.20 & \underline{81.45} & \textbf{58.43} & 58.52 & 58.32 & \underline{59.64} & 77.44 & 56.40 & 59.96 &
        \underline{59.01} & 71.47 & \textbf{73.88} & 66.47 & \textbf{76.75} & 85.46 & \underline{80.71} & 82.10 & 63.73 \\
        \model\texttt{-N} & 0.27B & \textbf{1.0x} & \textbf{71.11} &
        \textbf{78.76} & \textbf{83.14} & \underline{57.45} & \textbf{65.00} & \textbf{61.54} & \textbf{60.81} & \textbf{79.30} & \textbf{61.03} & \textbf{61.95} &
        \textbf{59.04} & \textbf{74.25} & 72.76 & \textbf{73.77} & \underline{75.74} & \textbf{85.78} & \textbf{82.47} & \underline{83.00} & \underline{64.14} \\
        \bottomrule
    \end{tabular}}
    \vspace{-0.4cm}
\end{table*}

\subsection{Main Results}
\label{sec:main}
Tables~\ref{tab:main_beir_results} and~\ref{tab:main_miracle_results} compare
the reranking performance and efficiency of the \model\ series with competitive baselines on BEIR and MIRACL. We make the following key observations:

\noindent\textbf{Lychee-Reranker achieves strong reranking performance with superior efficiency.}
The \model\ series matches or outperforms strong industrial rerankers from the Qwen, BGE, Jina, and mxbai families.
For example, on BEIR, \model\texttt{-Nano}, \texttt{-Small}, and \texttt{-Large} achieve the best or second-best results on 13/13, 12/13, and 13/13 tasks, respectively, within their respective size groups.
Beyond its strong reranking performance, \model\ is also highly efficient.
For example, \model\texttt{-Small} outperforms Qwen3-Reranker-0.6B at a substantially lower online computation cost (4.6x vs. 28.2x).
\noindent\textbf{Lychee-Reranker demonstrates strong multilingual reranking capabilities.}
Despite limited multilingual training, \model\ achieves competitive performance on MIRACL.
For example, \model\texttt{-Large} outperforms Qwen3-Reranker-8B on MIRACL with a substantially lower online computation cost (29.1x vs. 359.8x).
Within its parameter group, \model\texttt{-L} ranks first in 11 of the 18 languages (\emph{i.e.,} ar, bn, de, en, es, fi, fr, ja, ko, ru, and sw);
six of these languages (\emph{i.e.,} en, es, de, ja, fr, and ru) have reported website usage rates totaling approximately 74.2\% among websites with known content languages~\citep{w3techs2026contentlanguages}.
\noindent\textbf{The FBNL reranking paradigm combines efficiency with expressive relevance modeling.}
Built on the FBNL paradigm and trained on 3.9M instances (Table~\ref{tab:training_data}), the \model\ series achieves performance on BEIR comparable to that of the Qwen3-Reranker series. 
In contrast, the Qwen3-Reranker series follows the conventional reranking paradigm and is trained on 19M high-quality instances~\citep{zhang2025qwen3}.
Meanwhile, \model\ is substantially more efficient, demonstrating that the FBNL reranking paradigm supports expressive relevance modeling while reducing serving costs.
Results and analysis on LMEB~\citep{DBLP:journals/corr/abs-2603-12572} are provided in \S\ref{app:lmeb}.
\begin{figure}[t]
    \centering
    \begin{subfigure}[t]{0.485\linewidth}
        \centering
        \includegraphics[width=\linewidth]{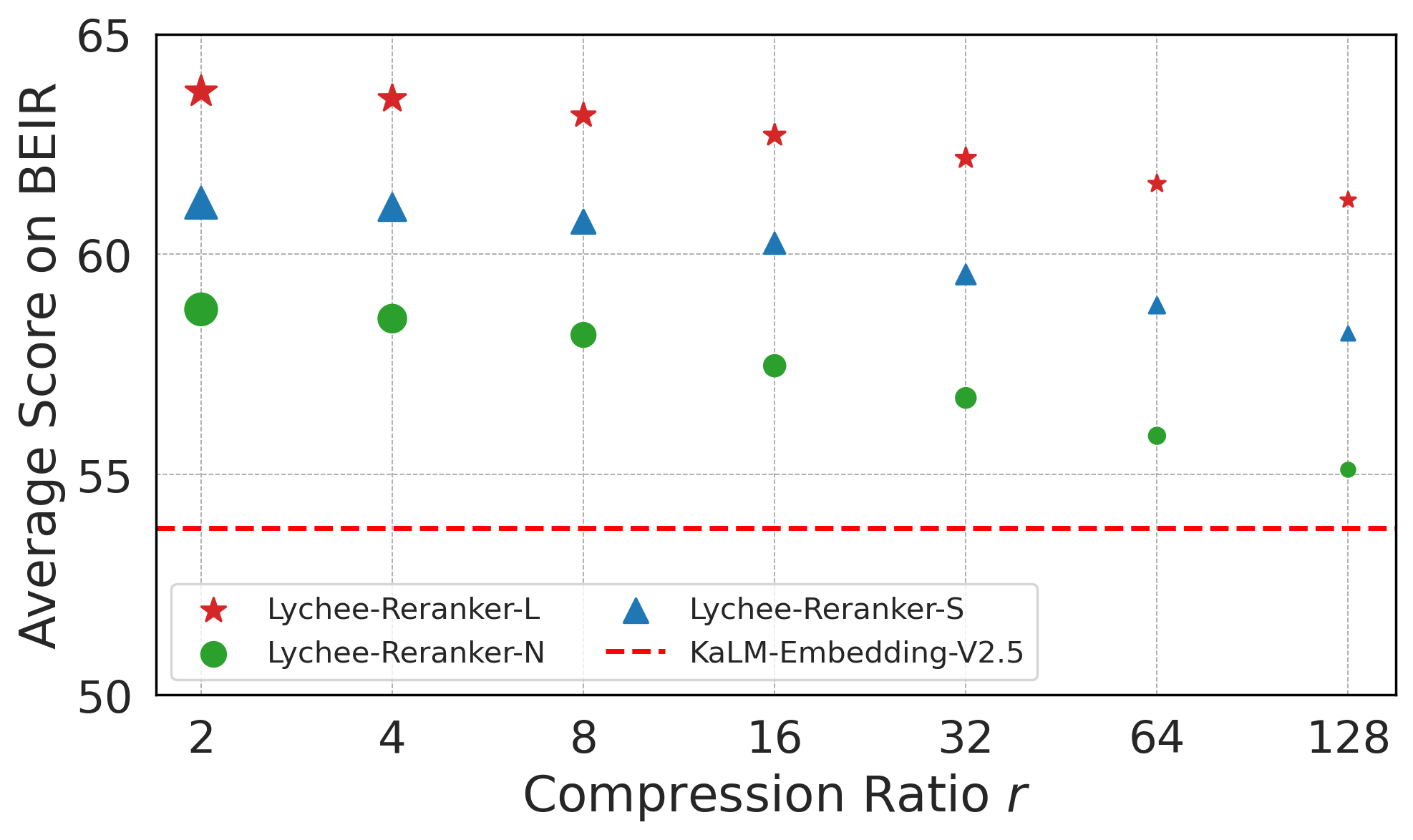}
        \caption{BEIR.}
        \label{fig:beir_mep}
        \vspace{0.05cm}
    \end{subfigure}
    \hfill
    \begin{subfigure}[t]{0.485\linewidth}
        \centering
        \includegraphics[width=\linewidth]{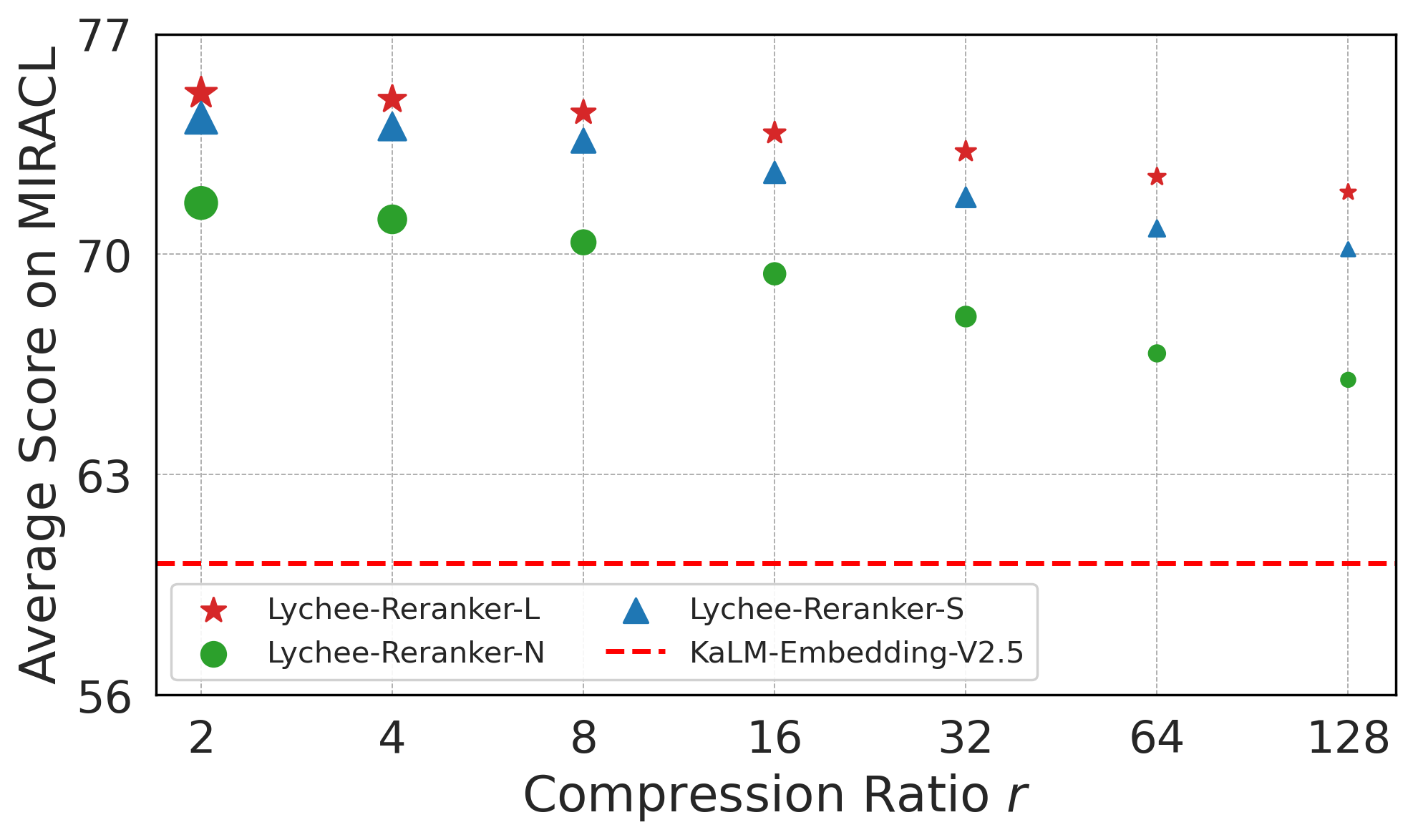}
        \caption{MIRACL.}
        \label{fig:miracle_mep}
        \vspace{0.05cm}
    \end{subfigure}
    \caption{Reranking performance of \model\ on BEIR and MIRACL under different Matryoshka embedding pooling compression ratios. Per-task results are reported in Tables~\ref{tab:beir_mep_results} and~\ref{tab:miracle_mep_results}.}
    \label{fig:mep_scatter}
    \vspace{-0.2cm}
\end{figure}

\subsection{In-depth Analysis}

\subsubsection{Effectiveness of Matryoshka Reranking}
\label{sec:matryoshka}
Figure~\ref{fig:mep_scatter} and Tables~\ref{tab:beir_mep_results}
and~\ref{tab:miracle_mep_results} report performance across MEP
compression ratios on BEIR and MIRACL. They reveal the
compactness--effectiveness trade-off. We make the following observations:

\noindent\textbf{Reranking performance decreases as the compression ratio increases.}
Overall, performance decreases as the compression ratio increases because aggressive pooling discards some useful information.
However, performance degrades only modestly as $r$ increases from 2 to 16, indicating that our models retain most of their reranking effectiveness under moderate compression.
Even at $r=128$, all \model\ variants retain at least 92\% of their average nDCG@10 at $r=2$ on both BEIR and MIRACL.
For practical deployment, we recommend compression ratios from $r=4$ to $r=16$, which offer a favorable trade-off between effectiveness and efficiency.

\noindent\textbf{Larger models are more robust to compression.}
As $r$ increases, passage representations retain less information, which can reduce reranking performance.
Larger models show smaller performance drops at the same compression ratio.
For example, at $r=128$ on BEIR, \model\texttt{-Large} retains 96.1\% of its average nDCG@10 at $r=2$, compared with 95.1\% for \model\texttt{-Small} and 93.8\% for \model\texttt{-Nano}.
Overall, larger rerankers may be better suited to higher compression ratios, partly because their larger hidden dimensions (Table~\ref{tab:model-spec}) may preserve more information per compressed passage token.

\noindent\textbf{Performance gains show diminishing returns as the compression ratio decreases.}
Lower compression ratios preserve more information and generally improve performance, but diminishing gains suggest redundancy in passage representations.
Moderate compression ratios, ranging from $r=2$ to $r=8$, therefore provide substantial efficiency gains while incurring little performance loss.
These findings support moderate compression in practice and confirm the effectiveness of the FBNL design.
\subsubsection{Potential of Test-time Compute Scaling}
Tables~\ref{tab:test_time_compute} and~\ref{tab:test_time_compute_miracl} examine whether selectively allocating more test-time compute to promising candidates can achieve a favorable trade-off between reranking effectiveness and efficiency when the same model is used throughout.
Specifically, we first rerank 100 candidates at a high compression ratio ($r=32$ or $r=64$), then rescore the top-20 candidates with the same model at $r=2$ and return these 20 candidates in descending score order as the final ranked list.
Results on BEIR and MIRACL show that this strategy largely preserves reranking effectiveness while reducing the estimated online computation cost severalfold.
For example, on BEIR, \model\texttt{-Small} achieves an average nDCG@10 of 61.11 under setting (c), compared with 61.17 under setting (a), while reducing the estimated relative serving cost from 8.65x to 2.75x.
Interestingly, settings (c) and (d) outperform setting (a) on some tasks, suggesting that reranking more candidates at a low compression ratio does not necessarily improve reranking quality.
The preliminary reranking stage ($r=32$ or $r=64$) may filter out misleading candidates before the subsequent post-ranking stage ($r=2$).
Overall, high- and low-compression reranking can complement each other, highlighting their practical value for deployment.
\S\ref{app:analysis} provides results on performance scaling with test-time compute.

\begin{table*}[t]
\centering
\caption{Test-time compute scaling results on BEIR, measured by nDCG@10. The reranking and post-ranking columns specify the number of documents ($\mathrm{d}$) and compression ratios ($r$). Cost denotes the estimated relative online computation cost, with the cost of \model\texttt{-Nano} at $r=4$ normalized to 1.0x. Note that all other evaluation settings follow those described in Table~\ref{tab:main_beir_results}.}
\label{tab:test_time_compute}
\vspace{-0.1cm}
\renewcommand{\arraystretch}{1.15}
\setlength{\tabcolsep}{7pt}
\resizebox{\textwidth}{!}{
\begin{tabular}{c c c c c c*{13}{c}}
\toprule
\multirow{2}{*}{\textbf{Models}}
& \multirow{2}{*}{\textbf{Setting}}
& \multirow{2}{*}{\textbf{Cost}}
& \multirow{2}{*}{\textbf{Avg.}}
& \multicolumn{2}{c}{\textbf{Test-Time Compute}}
& \multicolumn{13}{c}{\textbf{BEIR}} \\
\cmidrule(lr){5-6}\cmidrule(lr){7-19}
& & & & \textbf{Reranking} & \textbf{Post-ranking}
& \textbf{AA} & \textbf{CF} & \textbf{CQA} & \textbf{DB} & \textbf{FV} & \textbf{FQA} & \textbf{HQA} & \textbf{MS} & \textbf{NF} & \textbf{NQ} & \textbf{QR} & \textbf{SD} & \textbf{SF} \\
\midrule

& (a) & 1.89x
& \textbf{58.75}
& $100\mathrm{d} \xrightarrow{r=2} 100\mathrm{d}$
& N/A
& 63.87 & \underline{43.23} & \underline{41.08} & \textbf{48.92} & \underline{93.13} & \textbf{47.25} & \textbf{82.80} & \textbf{46.17} & \underline{38.42} & \textbf{69.57} & \textbf{90.11} & \textbf{21.04} & \textbf{78.17} \\

& (b) & 1.0x
& \underline{58.54}
& $100\mathrm{d} \xrightarrow{r=4} 100\mathrm{d}$
& N/A
& \textbf{63.97} & \textbf{43.25} & 40.84 & 47.93 & 92.95 & \underline{47.09} & \underline{82.58} & \underline{45.64} & \textbf{38.72} & 69.30 & 89.92 & 20.80 & \underline{78.03} \\

\rowcolor{gray!12}\cellcolor{white}
& (c) & 0.6x
& 58.51
& $100\mathrm{d} \xrightarrow{r=32} 100\mathrm{d}$
& $20\mathrm{d} \xrightarrow{r=2} 20\mathrm{d}$
& \underline{63.96} & 43.02 & \textbf{41.16} & 48.53 & \textbf{93.18} & 46.26 & 82.55 & 45.51 & 38.11 & \underline{69.56} & \underline{90.08} & 20.81 & 77.89 \\

\rowcolor{gray!12}\cellcolor{white}\multirow{-4}{*}{\model\texttt{-N}}
& (d) & 0.54x
& 58.46
& $100\mathrm{d} \xrightarrow{r=64} 100\mathrm{d}$
& $20\mathrm{d} \xrightarrow{r=2} 20\mathrm{d}$
& 63.89 & 43.01 & 41.05 & \underline{48.81} & \textbf{93.18} & 46.17 & 82.34 & 45.50 & 38.09 & 69.29 & 90.06 & \underline{20.85} & 77.76 \\

\midrule

& (a) & 8.65x
& \textbf{61.17}
& $100\mathrm{d} \xrightarrow{r=2} 100\mathrm{d}$
& N/A
& 68.79 & \textbf{45.98} & \textbf{47.35} & \textbf{50.30} & \textbf{93.50} & 54.08 & \textbf{83.74} & \textbf{47.33} & 39.95 & \underline{73.09} & \underline{90.85} & 22.05 & \underline{78.26} \\

& (b) & 4.58x
& 61.07
& $100\mathrm{d} \xrightarrow{r=4} 100\mathrm{d}$
& N/A
& \textbf{69.43} & 45.63 & 47.06 & 49.77 & 93.36 & 53.62 & 83.55 & \underline{46.90} & 39.89 & \textbf{73.14} & 90.77 & \underline{22.15} & \textbf{78.72} \\

\rowcolor{gray!12}\cellcolor{white}
& (c) & 2.75x
& \underline{61.11}
& $100\mathrm{d} \xrightarrow{r=32} 100\mathrm{d}$
& $20\mathrm{d} \xrightarrow{r=2} 20\mathrm{d}$
& 69.28 & 45.77 & \textbf{47.35} & \underline{50.10} & \underline{93.47} & \textbf{54.42} & \underline{83.66} & 46.79 & \textbf{40.14} & 72.95 & \textbf{90.86} & \underline{22.15} & 77.48 \\

\rowcolor{gray!12}\cellcolor{white}\multirow{-4}{*}{\model\texttt{-S}}
& (d) & 2.49x
& 61.06
& $100\mathrm{d} \xrightarrow{r=64} 100\mathrm{d}$
& $20\mathrm{d} \xrightarrow{r=2} 20\mathrm{d}$
& \underline{69.29} & \underline{45.82} & \underline{47.21} & 49.90 & \underline{93.47} & \underline{54.20} & 83.48 & 46.67 & \underline{40.08} & 72.83 & \underline{90.85} & \textbf{22.22} & 77.80 \\

\midrule

& (a) & 55.05x
& \textbf{63.69}
& $100\mathrm{d} \xrightarrow{r=2} 100\mathrm{d}$
& N/A
& 72.97 & \underline{47.80} & \textbf{52.76} & \textbf{53.29} & \textbf{94.26} & \textbf{59.31} & \textbf{84.39} & \textbf{47.72} & 42.26 & \textbf{76.06} & {90.76} & 24.91 & \textbf{81.49} \\

& (b) & 29.14x
& 63.53
& $100\mathrm{d} \xrightarrow{r=4} 100\mathrm{d}$
& N/A
& 72.90 & \textbf{47.94} & 52.36 & 52.73 & 94.22 & 59.15 & 84.23 & 47.38 & \textbf{42.35} & 75.61 & 90.72 & 24.89 & \underline{81.42} \\

\rowcolor{gray!12}\cellcolor{white}
& (c) & 17.49x
& \underline{63.62}
& $100\mathrm{d} \xrightarrow{r=32} 100\mathrm{d}$
& $20\mathrm{d} \xrightarrow{r=2} 20\mathrm{d}$
& \underline{73.31} & 47.76 & \underline{52.68} & \underline{53.09} & \textbf{94.26} & \underline{59.22} & \underline{84.35} & 47.39 & \underline{42.33} & \underline{75.84} & \textbf{90.79} & \textbf{25.10} & 80.95 \\

\rowcolor{gray!12}\cellcolor{white}\multirow{-4}{*}{\model\texttt{-L}}
& (d) & 15.87x
& 63.60
& $100\mathrm{d} \xrightarrow{r=64} 100\mathrm{d}$
& $20\mathrm{d} \xrightarrow{r=2} 20\mathrm{d}$
& \textbf{73.42} & 47.79 & 52.56 & 52.99 & {94.23} & 59.10 & 84.23 & \underline{47.42} & 42.24 & \underline{75.84} & \textbf{90.79} & \underline{25.06} & 81.17 \\

\bottomrule
\end{tabular}
}
\vspace{-0.2cm}
\end{table*}

\begin{table*}[t]
    \centering
    \caption{Ablation study on the progressive multi-stage training pipeline (\S\ref{sec:multi_stage}). ``Nano'', ``Small'', and ``Large'' denote \model\texttt{-Nano}, \model\texttt{-Small}, and \model\texttt{-Large}, respectively. We report the average nDCG@10 on BEIR and MIRACL after each training stage under a range of compression ratios $r\in\{2,4,8,16,32,64,128\}$. ``Retention'' denotes the percentage of average nDCG@10 retained at $r=128$ relative to $r=2$. Within each model group, the best results are \textbf{boldfaced}, and the second-best results are \underline{underlined}.}
    \label{tab:multistage_ablation}
    \vspace{-0.1cm}
    \renewcommand\arraystretch{1.15}
    \setlength{\tabcolsep}{8pt}
    \resizebox{\textwidth}{!}{\begin{tabular}{llcccccccc|cccccccc}
        \toprule
        \multirow{2}{*}{\textbf{Models}} &
        \multirow{2}{*}{\shortstack[c]{\textbf{Training} \\ \textbf{Stage}}} &
        \multicolumn{8}{c|}{\textbf{BEIR}} &
        \multicolumn{8}{c}{\textbf{MIRACL}} \\
        \cline{3-10}\cline{11-18}
        & & $r=2$ & $r=4$ & $r=8$ & $r=16$ & $r=32$ & $r=64$ & $r=128$ & \textbf{Retention} & $r=2$ & $r=4$ & $r=8$ & $r=16$ & $r=32$ & $r=64$ & $r=128$ & \textbf{Retention} \\
        \hline
        \multirow{3}{*}{\texttt{Nano}}
        & Stage 1 & 57.72 & 57.44 & 56.84 & 56.18 & 55.29 & 54.26 & 53.61 & 92.88\% & 68.93 & 68.33 & 67.48 & 66.16 & 65.03 & 63.39 & 62.68 & 90.93\% \\
        & Stage 2 & \underline{58.51} & \underline{58.28} & \underline{57.82} & \underline{57.20} & \underline{56.34} & \underline{55.47} & \underline{54.62} & \underline{93.35\%} & \underline{71.04} & \underline{70.53} & \underline{69.77} & \underline{68.55} & \underline{67.38} & \underline{66.21} & \underline{65.36} & \underline{92.00\%} \\
        & Stage 3 & \textbf{58.75} & \textbf{58.54} & \textbf{58.17} & \textbf{57.47} & \textbf{56.74} & \textbf{55.88} & \textbf{55.11} & \textbf{93.80\%} & \textbf{71.63} & \textbf{71.11} & \textbf{70.38} & \textbf{69.38} & \textbf{68.02} & \textbf{66.85} & \textbf{66.01} & \textbf{92.15\%} \\
        \hline
        \multirow{3}{*}{\texttt{Small}}
        & Stage 1 & 59.57 & 59.49 & 58.96 & 58.41 & 57.60 & 56.91 & 56.11 & 94.19\% & 72.90 & 72.61 & 72.03 & 70.91 & 69.78 & 68.75 & 67.97 & 93.24\% \\
        & Stage 2 & \underline{61.07} & \underline{60.92} & \underline{60.53} & \underline{59.96} & \underline{59.28} & \underline{58.52} & \underline{57.76} & \underline{94.58\%} & \underline{74.10} & \underline{73.63} & \underline{73.08} & \underline{72.15} & \underline{71.28} & \underline{70.40} & \underline{69.62} & \underline{93.95\%} \\
        & Stage 3 & \textbf{61.17} & \textbf{61.07} & \textbf{60.74} & \textbf{60.25} & \textbf{59.54} & \textbf{58.84} & \textbf{58.20} & \textbf{95.14\%} & \textbf{74.34} & \textbf{74.06} & \textbf{73.61} & \textbf{72.60} & \textbf{71.81} & \textbf{70.82} & \textbf{70.16} & \textbf{94.38\%} \\
        \hline
        \multirow{3}{*}{\texttt{Large}}
        & Stage 1 & 62.30 & 62.04 & 61.61 & 61.14 & 60.51 & 59.84 & 59.29 & 95.17\% & 74.63 & 74.24 & 73.54 & 72.52 & 71.54 & 70.67 & 69.96 & 93.74\% \\
        & Stage 2 & \underline{63.56} & \underline{63.37} & \underline{63.01} & \underline{62.53} & \underline{62.06} & \underline{61.46} & \underline{61.07} & \underline{96.08\%} & \underline{75.03} & \underline{74.83} & \underline{74.40} & \underline{73.74} & \underline{73.02} & \underline{72.31} & \underline{71.89} & \textbf{95.82\%} \\
        & Stage 3 & \textbf{63.69} & \textbf{63.53} & \textbf{63.15} & \textbf{62.70} & \textbf{62.18} & \textbf{61.60} & \textbf{61.23} & \textbf{96.14\%} & \textbf{75.12} & \textbf{74.92} & \textbf{74.50} & \textbf{73.85} & \textbf{73.26} & \textbf{72.46} & \textbf{71.97} & \underline{95.81\%} \\
        \bottomrule
    \end{tabular}}
    \vspace{-0.35cm}
\end{table*}

\subsubsection{Ablation Study on Multi-stage Training}
In \S\ref{sec:multi_stage}, we introduce a multi-stage training pipeline to progressively improve reranking performance.
Table~\ref{tab:multistage_ablation} presents an ablation study of the three stages across diverse compression ratios $r\in\{2,4,8,16,32,64,128\}$.
For Stage~2, we train models using two LoRA configurations: a rank of 64 with an alpha of 32 and a rank of 96 with an alpha of 48.
Since the two configurations perform similarly, Table~\ref{tab:multistage_ablation} reports results for the configuration with a rank of 64 and an alpha of 32.
Across all three model sizes and all seven compression ratios, performance on BEIR and MIRACL improves consistently from Stage~1 to Stage~3, demonstrating the effectiveness of our multi-stage training pipeline.
Concretely, Stage~1 establishes a robust foundation for document reranking, distillation in Stage~2 substantially improves performance, and Stage~3 yields further modest gains.
More importantly, robustness to compression generally improves across the three training stages.
For example, from Stage~1 to Stage~3, the performance retention of \model\texttt{-Nano} at $r=128$ relative to $r=2$ increases from 92.88\% to 93.80\% on BEIR and from 90.93\% to 92.15\% on MIRACL.
Overall, these results show the effectiveness of the multi-stage training pipeline and provide a reference for future research on the FBNL paradigm.
\S\ref{app:in_depth} examines the effect of compression on the model's discriminative ability.
\S\ref{app:mep_ablation} presents an ablation study on MEP.
\S\ref{app:case} presents case studies.

\section{Conclusion}
We present \model, a family of efficient rerankers built on the \emph{fast but not late-interaction} (FBNL) paradigm.
Under the FBNL paradigm, the encoder produces reusable passage representations offline, while the decoder processes instructions and queries, interacting with these representations through merged self-attention and cross-attention to perform expressive relevance modeling.
We introduce Matryoshka embedding pooling (MEP) to compress passage representations along the sequence dimension, reducing storage and serving costs while enabling flexible trade-offs between efficiency and effectiveness.
Together, these designs offer four key advantages: (i) \emph{efficient online serving}, (ii) \emph{expressive relevance modeling}, (iii) \emph{compact passage representations}, and (iv) \emph{adjustable test-time compute}.
Extensive experiments on BEIR, MIRACL, and LMEB show that the \model\ series is competitive with strong industrial rerankers while significantly reducing online computation costs.
In-depth analyses further demonstrate the models' strong robustness to compression, the effectiveness of our multi-stage training pipeline, and the value of test-time compute scaling.
Overall, we hope that \model\ will inspire further innovation within the FBNL paradigm and facilitate the development of scalable retrieval systems.

\section*{AI Use Statement}
AI tools were used to polish the writing and provide feedback on experimental design. The authors reviewed and verified all suggestions and take full responsibility for the final paper.

\bibliography{iclr2027_conference}
\bibliographystyle{iclr2027_conference}

\appendix

\section{Complexity Analysis}
\label{app:complexity}

\subsection{Time Complexity}
\label{app:time_complexity}
\begin{table}[ht]
    \centering
    \caption{Time complexity of reranking $K$ passages. Enc- and dec-based rerankers process each query--passage pair online. By contrast, our enc--dec-based reranker reuses cached passage representations and applies merged self- and cross-attn to the query context and compressed passage representations.}
    \label{tab:time-complexity}
    \small
    \renewcommand{\arraystretch}{1.1}
    \resizebox{1.0\textwidth}{!}{
    \begin{tabular}{lcc}
        \toprule
        Model Type & Interaction Pattern & Time Complexity \\
        \midrule
        Encoder-based reranker & Bidirectional attention & $\mathcal{O}\left(KL\left((|q|+n)^2d+(|q|+n)d^2\right)\right)$ \\
        Decoder-based reranker & Causal attention & $\mathcal{O}\left(KL\left((|q|+n)^2d+(|q|+n)d^2\right)\right)$ \\
        Enc-Dec-based reranker \textbf{(ours)} & Merged self-attn. + cross-attn. & $\mathcal{O}\left(\frac{L}{2}K\left(|q|\left(|q|+\left\lceil\frac{n}{r}\right\rceil\right)d+\left(|q|+\left\lceil\frac{n}{r}\right\rceil\right)d^2\right)\right)$ \\
        \bottomrule
    \end{tabular}
    }
\end{table}

Table~\ref{tab:time-complexity} compares the online time complexity of our encoder--decoder reranker with that of conventional encoder- and decoder-based rerankers.
Let $K$ denote the number of candidate passages to rerank, $|q|$ the query length, $n$ the average passage length, and $d$ the hidden dimension.
Let $L$ denote the total number of Transformer layers in a reranker. For an encoder--decoder reranker, $L$ includes both the encoder and decoder layers. Let $r$ denote the MEP compression ratio.
For a Transformer layer processing a sequence of length $n$, self-attention has a complexity of $\mathcal{O}(n^2 d)$, while feed-forward and projection operations have a complexity of $\mathcal{O}(n d^2)$.
For a fair comparison, we assume that conventional encoder- and decoder-based rerankers have the same total parameter count as \model.
Since \model\ splits its parameters evenly between the encoder and decoder, the decoder used during online reranking has approximately $L/2$ layers.
The encoder is used only to pre-encode passages offline and therefore incurs no online reranking cost.
Conventional rerankers jointly process each query--passage pair during online inference to estimate relevance.
For each candidate, the resulting input sequence has a length of $|q|+n$\footnote{We ignore prompt-template and instruction tokens, as their length is negligible compared with $|q|+n$.}. Therefore, reranking $K$ candidates requires:
\begin{equation}
    \mathcal{O}\left(
        K L \left((|q|+n)^2 d + (|q|+n)d^2\right)
    \right).
\end{equation}
By contrast, \model\ performs only decoder computation online and reuses pre-encoded passage representations.
Through merged self- and cross-attention, the query tokens attend to both the query context and the cached passage representations.
With MEP, each passage is represented by only $\lceil n/r \rceil$ cached tokens, which remain fixed throughout relevance modeling.
Therefore, the online serving complexity of \model\ is:
\begin{equation}
\label{equ:enc_dec_time}
    \mathcal{O}\left(
        \frac{L}{2}K\left(
        |q|\left(|q| + \left\lceil \frac{n}{r} \right\rceil\right)d
        + \left(|q| + \left\lceil \frac{n}{r} \right\rceil\right)d^2
        \right)
    \right),
\end{equation}
where \textbf{the first term} accounts for merged self- and cross-attention between the query and the combined query--passage context. \textbf{The second term} accounts for query-side projections and feed-forward computation, together with key/value projections of the cached passage representations.
Encoding $K$ passages offline incurs a computational cost of $\mathcal{O}\left(K \frac{L}{2}(n^2d + nd^2)\right)$. Since each passage is encoded only once and its resulting representation is reused across queries, this cost is excluded from the online serving complexity.
\begin{figure}[t]
    \centering
    \begin{subfigure}[t]{0.485\linewidth}
        \centering
        \includegraphics[width=\linewidth]{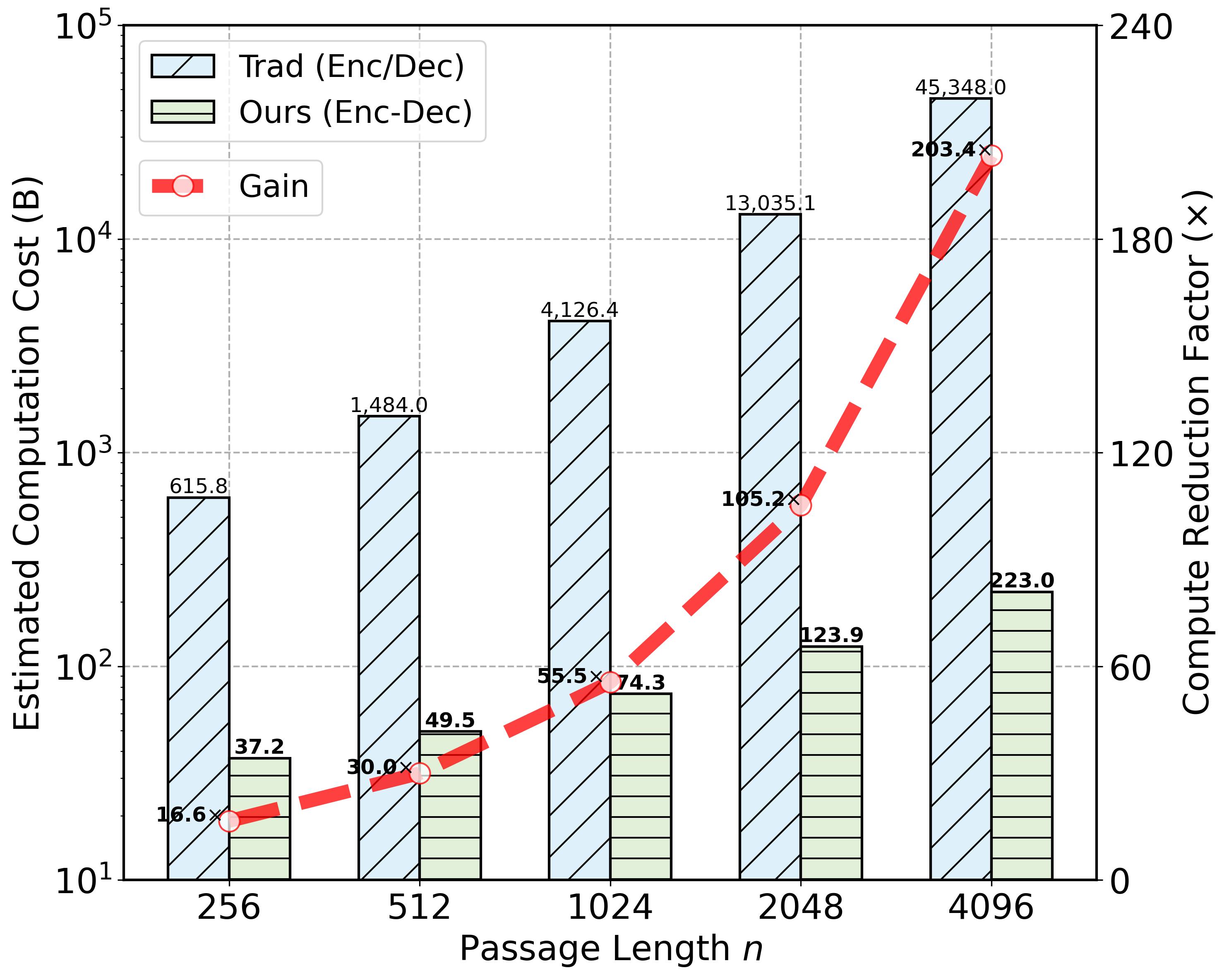}
        \caption{Varying the passage length $n$ while fixing $|q|=32$, $r=16$, $K=100$, $d=640$, and $L=36$.}
        \label{fig:time_complexity_n}
    \end{subfigure}
    \hfill
    \begin{subfigure}[t]{0.485\linewidth}
        \centering
        \includegraphics[width=\linewidth]{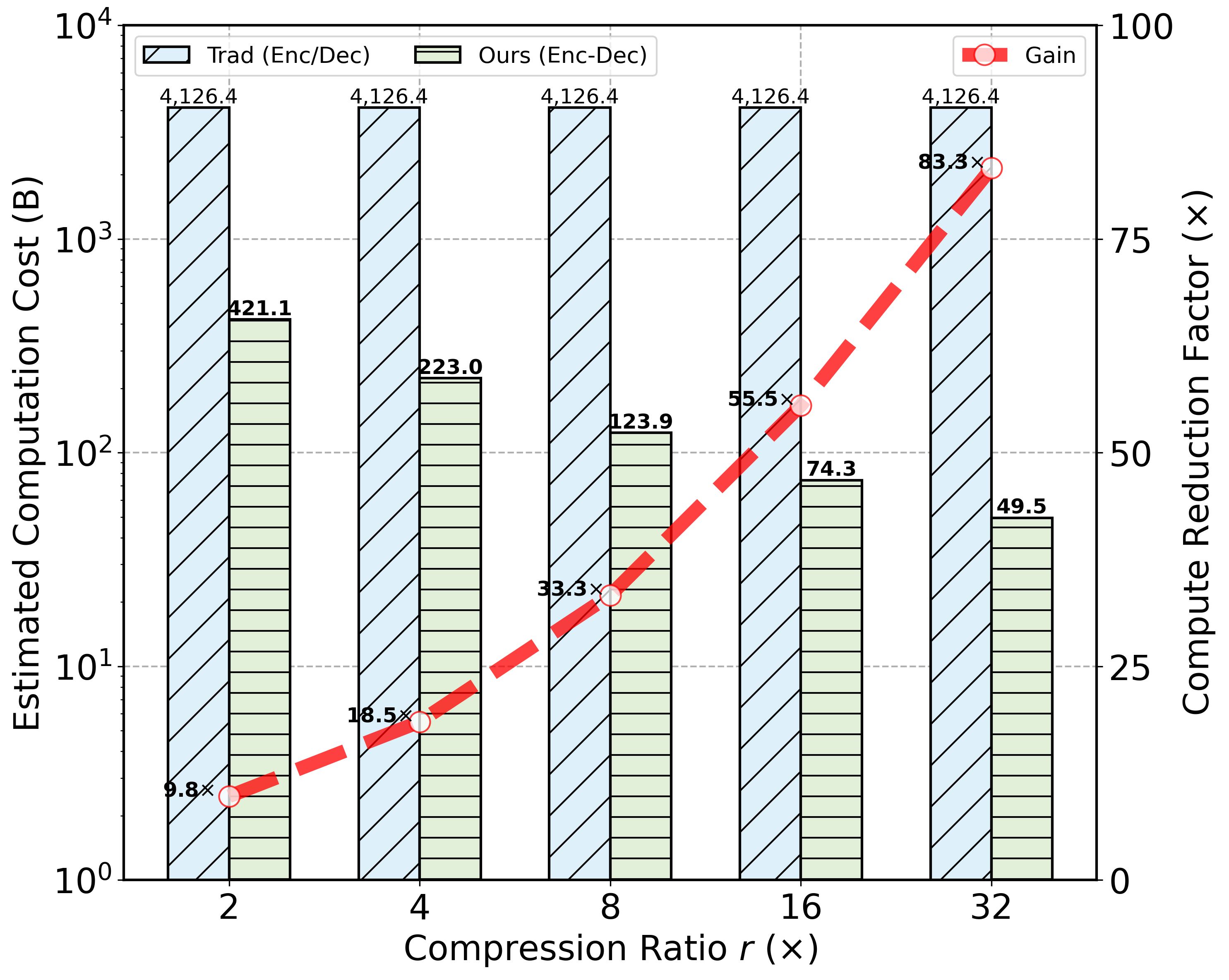}
        \caption{Varying the compression ratio $r$ while fixing $|q|=32$, $n=1024$, $K=100$, $d=640$, and $L=36$.}
        \label{fig:time_complexity_r}
    \end{subfigure}
    \caption{Comparison of online serving time complexity for varying passage lengths $n$ and compression ratios $r$. Bars show the estimated computational cost, and the line shows the factor by which our reranking paradigm reduces computation relative to traditional rerankers. Here, ``B'' denotes billion, ``Trad'' denotes traditional, ``Enc/Dec'' denotes encoder- or decoder-based rerankers, and ``Enc-Dec'' represents our encoder--decoder reranker.}
    \label{fig:time_complexity}
\end{figure}
\textbf{Efficiency Gains.}
Figure~\ref{fig:time_complexity} quantifies serving efficiency under representative settings.
In Figure~\ref{fig:time_complexity_n}, we fix $|q|$, $r$, $K$, $d$, and $L$ and vary the passage length over $n \in \{256,512,1024,2048,4096\}$.
Even at $n=256$, our method reduces the estimated computational cost by a factor of $16.6$.
The efficiency gain increases with passage length, reaching $203.4\times$ at $n=4096$.
In Figure~\ref{fig:time_complexity_r}, we fix $|q|$, $n$, $K$, $d$, and $L$ and vary the compression ratio over $r \in \{2,4,8,16,32\}$.
Even at the low compression ratio of $r=2$, our method reduces the estimated computational cost by a factor of nearly $10$; at $r=4$ and $r=8$, the reduction factor increases to $18.5$ and $33.3$, respectively.
Overall, these results demonstrate that offline passage encoding and MEP substantially reduce online serving overhead, with particularly large benefits for longer passages and higher compression ratios.
\subsection{Space Complexity}
We analyze the storage overhead of caching passage representations in \model\ at scale.
Conventional encoder- and decoder-based rerankers process each query--passage pair online and therefore require no caching of passage representations.
By contrast, our model caches pre-encoded passage representations to avoid recomputing them for each query, thereby introducing additional storage overhead.
Suppose the corpus contains $N$ passages, each with an average length of $n$ tokens, and that the passage representations have a hidden dimension of $d$.
Caching the full passage representations requires:
\begin{equation}
    \mathcal{O}(N n d),
\end{equation}
where each passage has a representation $\mathbf{H}_p \in \mathbb{R}^{n \times d}$.
When reranking a candidate list of size $K$, the online buffer for passage representations requires $\mathcal{O}(K n d)$ space before compression.
This buffer constitutes the main additional space overhead of our model relative to encoder- and decoder-based rerankers.
To reduce this overhead, \model\ introduces \textbf{Matryoshka Embedding Pooling} to compress passage representations along the sequence dimension.
With a compression ratio of $r$, each cached passage representation is compressed to $\mathbf{H}^{(r)}_p \in \mathbb{R}^{\lceil n/r \rceil \times d}$, reducing the total cache size to
\begin{equation}
    \mathcal{O}\left(N \left\lceil \frac{n}{r} \right\rceil d\right).
\end{equation}
The space required by the online candidate buffer is likewise reduced to $\mathcal{O}(K \lceil n/r \rceil d)$.
Thus, MEP balances storage cost and reranking quality: increasing $r$ reduces the cache size nearly linearly while retaining information for fine-grained relevance modeling by the decoder.

\section{Training Data}
\label{app:training_data}
We fine-tune \model\ on retrieval-specific datasets to develop its reranking capabilities.
To improve robustness and generalization, we collect and process large-scale, multilingual, and multi-domain training data spanning diverse retrieval scenarios, including web search, question answering, duplicate-question retrieval, and fact verification.
Our training data is drawn primarily from two public sources: the retrieval subset of the KaLM embedding fine-tuning dataset~\citep{zhao2026kalmembeddingv} and a selected subset of the BGE-M3 training dataset~\citep{chen2024bge}.
For queries with fewer than 16 hard negatives, we use KaLM-Embedding-V2.5~\citep{zhao2026kalmembeddingv} to retrieve the top-100 candidate passages from the corresponding corpus and sample 16 hard negatives from ranks 10--50. Otherwise, we retain the existing hard negatives.
This sampling strategy encourages the model to distinguish relevant passages from hard negatives while avoiding overly easy low-ranked negatives and reducing the risk of selecting false negatives from the highest-ranked candidates.
Each resulting training instance contains 1 query, 1 positive passage, and at least 16 hard negatives.
To provide fine-grained supervision, we use Qwen3-Reranker-8B~\citep{zhang2025qwen3} to assign a teacher score to each query--passage pair. The score is defined as the difference between the logits of the \texttt{yes} and \texttt{no} labels, \textit{i.e.,} $z_{\mathrm{yes}} - z_{\mathrm{no}}$.
Applying the sigmoid function to this score yields the soft teacher label $y$ used in Equation~\ref{equ:kd}.
After preprocessing, the final training set contains approximately 3.9 million instances.
Detailed statistics are provided in Table~\ref{tab:training_data}.

\begin{table}[ht]
    \centering
    \caption{Evaluation hyperparameters used in our experiments. Because bge-reranker-large supports a maximum sequence length of only 512 tokens, we set the maximum combined query and passage length to 512 tokens for this model.}
    \label{tab:eval_hyperparams}
    \renewcommand\arraystretch{1.05}
    \begin{tabular}{lc}
        \toprule
        \multicolumn{1}{l}{\textbf{Hyperparameter}} &
        \multicolumn{1}{c}{\textbf{Value}} \\
        \midrule
        Evaluation metric & nDCG@10 \\
        First-stage retriever & KaLM-Embedding-V2.5 \\
        Number of candidates retrieved per query, $K$ & 100 \\
        Default compression ratio $r$ & $r=4$ \\
        Precision & bf16 \\
        Embedder query max length $|q|$ & 512 \\
        Embedder passage max length $n$ & 512 \\
        Reranker query max length $|q|$ & 512 \\
        Reranker passage max length $n$ & 1024 \\
        \bottomrule
    \end{tabular}
    \vspace{0.075cm}
\end{table}

\begin{table*}[t]
    \centering
    \caption{Model specifications and public checkpoints for the embedding and reranking models evaluated in this work.}
    \label{tab:model_links}
    \renewcommand\arraystretch{1.2}
    \resizebox{\textwidth}{!}{\begin{tabular}{ l | c | c | c | l }
        \toprule
        \multicolumn{1}{l|}{\textbf{Model}} &
        \multicolumn{1}{c|}{\textbf{Params.}} &
        \multicolumn{1}{c|}{\textbf{\#Layers}} &
        \multicolumn{1}{c|}{\textbf{Hidden Dim.}} &
        \multicolumn{1}{|c}{\textbf{Public Model Checkpoints (Link)}} \\
        \midrule
        \multirow{2}{*}{KaLM-Embedding-V2.5} &
        \multirow{2}{*}{0.5B} &
        \multirow{2}{*}{24} &
        \multirow{2}{*}{896} &
        \href{https://huggingface.co/KaLM-Embedding/KaLM-embedding-multilingual-mini-instruct-v2.5}{\nolinkurl{https://huggingface.co/KaLM-Embedding/}} \\
        & & & & \href{https://huggingface.co/KaLM-Embedding/KaLM-embedding-multilingual-mini-instruct-v2.5}{\nolinkurl{KaLM-embedding-multilingual-mini-instruct-v2.5}} \\
        \midrule
        bge-reranker-large & 0.6B & 24 & 1024 & \url{https://huggingface.co/BAAI/bge-reranker-large} \\
        bge-reranker-v2-m3 & 0.6B & 24 & 1024 & \url{https://huggingface.co/BAAI/bge-reranker-v2-m3} \\
        bge-reranker-v2-gemma & 2.5B & 18 & 2048 & \url{https://huggingface.co/BAAI/bge-reranker-v2-gemma} \\
        gte-multilingual-reranker-base & 0.3B & 12 & 768 & \url{https://huggingface.co/Alibaba-NLP/gte-multilingual-reranker-base} \\
        Qwen3-Reranker-0.6B & 0.6B & 28 & 1024 & \url{https://huggingface.co/Qwen/Qwen3-Reranker-0.6B} \\
        Qwen3-Reranker-4B & 4B & 36 & 2560 & \url{https://huggingface.co/Qwen/Qwen3-Reranker-4B} \\
        Qwen3-Reranker-8B & 8B & 36 & 4096 & \url{https://huggingface.co/Qwen/Qwen3-Reranker-8B} \\
        jina-reranker-v2-base-multilingual & 0.3B & 12 & 768 & \url{https://huggingface.co/jinaai/jina-reranker-v2-base-multilingual} \\
        mxbai-rerank-base-v2 & 0.5B & 24 & 896 & \url{https://huggingface.co/mixedbread-ai/mxbai-rerank-base-v2} \\
        mxbai-rerank-large-v2 & 1.5B & 28 & 1536 & \url{https://huggingface.co/mixedbread-ai/mxbai-rerank-large-v2} \\
        \bottomrule
    \end{tabular}}
\end{table*}

\section{Experimental Setup}
\label{app:setup}
\textbf{Benchmarks.} We evaluate reranking performance on retrieval benchmarks covering diverse aspects of reranking capability.
We use BEIR~\citep{thakur2021beir} to assess reranking performance across 13 heterogeneous English retrieval tasks, ranging from question answering on HotpotQA~\citep{yang2018hotpotqa} and duplicate-question retrieval on CQADupStack~\citep{hoogeveen2015cqadupstack} to entity retrieval on DBPedia~\citep{hasibi2017dbpedia} and fact verification on FEVER~\citep{thorne2018fever}.
We exclude Touche2020 and TRECCOVID because their test sets contain only 49 and 50 queries, respectively, making the evaluation results sensitive to individual queries.
We use MIRACL~\citep{zhang2023miracl} to evaluate multilingual reranking performance across 18 languages from diverse linguistic families, including Arabic, Bengali, Hindi, and Swahili.
We further evaluate on the six dialogue memory retrieval tasks from LMEB~\citep{DBLP:journals/corr/abs-2603-12572}, which assess retrieval and reranking capabilities in long-horizon dialogue memory scenarios.
We adopt a two-stage retrieval-then-reranking pipeline for all experiments.
In the first stage, we use KaLM-Embedding-V2.5~\citep{zhao2026kalmembeddingv} as the dense retriever to obtain the top-100 candidate passages for each query.
In the second stage, all rerankers rescore the same top-100 candidates to ensure a fair comparison.
We use nDCG@10 as the main evaluation metric.
Task-specific evaluation instructions are provided in Appendix~\ref{app:inst}.
Overall, these benchmarks provide a comprehensive evaluation across diverse retrieval tasks and domains, multilingual settings, and long-horizon memory scenarios.
\textbf{Baselines.} We compare the \model\ series with representative open-source rerankers whose parameter counts range from 0.3B to 8B.
These include bge-reranker-large, bge-reranker-v2-m3, and bge-reranker-v2-gemma~\citep{li2024llama2vec,xiao2024c,chen2024bge};
gte-multilingual-reranker-base (gte-reranker-base), Qwen3-Reranker-0.6B, Qwen3-Reranker-4B, and Qwen3-Reranker-8B~\citep{zhang2025qwen3,zhang2024mgte};
jina-reranker-v2-base-multilingual (jina-reranker-v2)~\citep{jinaai2025rerankerv2base};
and mxbai-rerank-base-v2 and mxbai-rerank-large-v2~\citep{li2025prorank}.
These baselines cover widely used encoder- and decoder-based rerankers, as well as recent multilingual and LLM-based rerankers, enabling comprehensive comparisons across architectures and model scales.
Unless otherwise specified, \model\ uses the default compression ratio of $r=4$ in all experiments.
Detailed evaluation hyperparameter are provided in Table~\ref{tab:eval_hyperparams}.
Table~\ref{tab:model_links} summarizes the public checkpoints and key specifications of the embedding and reranking models used in our evaluation.
\section{Implementation Details}
\label{app:imple}
The \model\ series is initialized from T5Gemma2 encoder--decoder checkpoints~\citep{DBLP:journals/corr/abs-2512-14856} and fine-tuned using LoRA~\citep{hu2021lora}, with LoRA adapters applied to both the encoder and decoder.
For \model\texttt{-Nano} and \model\texttt{-Small}, LoRA targets all attention and feed-forward projection modules: \texttt{q\_proj}, \texttt{k\_proj}, \texttt{v\_proj}, \texttt{out\_proj}, \texttt{gate\_proj}, \texttt{up\_proj}, and \texttt{down\_proj}. For \model\texttt{-Large}, LoRA targets only the attention projection modules: \texttt{q\_proj}, \texttt{k\_proj}, \texttt{v\_proj}, and \texttt{out\_proj}.
Specifically, \model\texttt{-Nano}, \model\texttt{-Small}, and \model\texttt{-Large} are initialized from the \texttt{t5gemma-2-270m-270m}, \texttt{t5gemma-2-1b-1b}, and \texttt{t5gemma-2-4b-4b} checkpoints, respectively.
All models are trained using bf16 precision, with maximum query and passage lengths of 128 and 512 tokens, respectively.
The training group size is 16, with each training group containing 1 positive passage and 15 negative passages sampled from the negative pool.
For Matryoshka Embedding Pooling (MEP), we use compression ratios $\mathcal{R}=\{2,8,32,128\}$ and assign an equal loss weight of $\lambda_r=1$ to each ratio.
We optimize the models using Adam~\citep{kingma2014adam} ($\beta_1=0.9$, $\beta_2=0.999$), with a weight decay of $1\times 10^{-5}$ and a cosine learning-rate schedule. 
We clip gradients to a maximum gradient norm of 1.0.
We follow the progressive multi-stage training pipeline described in \S\ref{sec:multi_stage}.
All models undergo all three training stages.
We set the learning rate to $2\times 10^{-4}$ for the first two stages and train each model for one epoch in each stage. The third stage applies model soup and involves no additional optimization.
In the distillation stage, Qwen3-Reranker-8B~\citep{zhang2025qwen3} serves as the teacher for all models.
For each model, we train two variants in the second stage: one with a LoRA rank of 96 and an alpha of 48, and the other with a LoRA rank of 64 and an alpha of 32.
In the third stage, we first merge each LoRA adapter into its backbone and then construct a model soup~\citep{cocktail} by averaging the two resulting model checkpoints with equal weights of 0.5.
The per-GPU batch sizes are 2, 2, and 1, and the numbers of gradient accumulation steps are 2, 2, and 4 for \model\texttt{-Nano}, \model\texttt{-Small}, and \model\texttt{-Large}, respectively.
All models are trained on 32 NVIDIA RTX 5090 GPUs, with an effective global batch size of 128.
Training takes approximately 30k steps per epoch.
Table~\ref{tab:training_hyperparams} summarizes the detailed training hyperparameters.
\begin{table}[ht]
    \centering
    \caption{Training hyperparameters for the \model\ series.}
    \label{tab:training_hyperparams}
    \vspace{-0.1cm}
    \renewcommand\arraystretch{1.15}
    \resizebox{1.0\textwidth}{!}{\begin{tabular}{lccc}
        \toprule
        \multicolumn{1}{l}{\textbf{Hyperparameter}} &
        \multicolumn{1}{c}{\textbf{Nano}} &
        \multicolumn{1}{c}{\textbf{Small}} &
        \multicolumn{1}{c}{\textbf{Large}} \\
        \midrule
        Number of stages & 3 & 3 & 3 \\
        Precision & bf16 & bf16 & bf16 \\ 
        Query max length $|q|$ & 128 & 128 & 128 \\
        Passage max length $n$ & 512 & 512 & 512 \\
        MEP compression ratios $\mathcal{R}$ & $\{2,8,32,128\}$ & $\{2,8,32,128\}$ & $\{2,8,32,128\}$ \\ 
        Loss weight $\lambda_r$ for each $r \in \mathcal{R}$ & 1 & 1 & 1 \\ 
        Learning rate (Stages 1--2) & $2\times 10^{-4}$ & $2\times 10^{-4}$ & $2\times 10^{-4}$ \\
        Per-GPU Batch size & 2 & 2 & 1 \\
        Gradient Accumulation steps & 2 & 2 & 4 \\
        Number of GPUs & 32 & 32 & 32 \\
        Effective Global batch size & 128 & 128 & 128 \\
        GPU hours per epoch & $\sim$600 & $\sim$1400 & $\sim$2550 \\
        LoRA configuration & \multicolumn{3}{c}{Stage 1: (rank 96, alpha 48); Stage 2: (rank 96, alpha 48) and (rank 64, alpha 32)} \\
        \shortstack[c]{LoRA target modules} & \multicolumn{2}{c}{\shortstack[c]{\texttt{q\_proj}, \texttt{k\_proj}, \texttt{v\_proj}, \texttt{out\_proj}, \\ \texttt{gate\_proj}, \texttt{up\_proj}, \texttt{down\_proj}}} & \shortstack[c]{\texttt{q\_proj}, \texttt{k\_proj}, \\ \texttt{v\_proj}, \texttt{out\_proj}} \\
        Model soup weights (Stage 3) & $0.5, 0.5$ & $0.5, 0.5$ & $0.5, 0.5$ \\
        Number of epochs (Stages 1--2) & 1 epoch per stage & 1 epoch per stage & 1 epoch per stage \\
        Warmup ratio & 0.1 & 0.1 & 0.1 \\
        Backbone & \href{https://huggingface.co/google/t5gemma-2-270m-270m}{t5gemma-2-270m-270m} & \href{https://huggingface.co/google/t5gemma-2-1b-1b}{t5gemma-2-1b-1b} & \href{https://huggingface.co/google/t5gemma-2-4b-4b}{t5gemma-2-4b-4b} \\
        Teacher model & \href{https://huggingface.co/Qwen/Qwen3-Reranker-8B}{Qwen3-Reranker-8B} & \href{https://huggingface.co/Qwen/Qwen3-Reranker-8B}{Qwen3-Reranker-8B} & \href{https://huggingface.co/Qwen/Qwen3-Reranker-8B}{Qwen3-Reranker-8B} \\
        \bottomrule
    \end{tabular}}
    \vspace{0.075cm}
\end{table}

\begin{table*}[ht]
    \centering
    \caption{Task instructions used for evaluation on the BEIR~\citep{thakur2021beir} benchmark.}
    \label{tab:inst_beir}
    \vspace{-0.1cm}
    \renewcommand\arraystretch{1.15}
    \resizebox{\textwidth}{!}{\begin{tabular}{ll}
        \toprule
        \multicolumn{1}{l}{\textbf{Task Name}} &
        \multicolumn{1}{l}{\textbf{Instruction Template}} \\
        \midrule
        ArguAna (\textbf{AA}) & Given a claim, find documents that refute the claim. \\
        Climate-FEVER (\textbf{CF}) & Given a claim about climate change, retrieve documents that support or refute the claim. \\
        CQADupStack (\textbf{CQA}) & Given a question, retrieve detailed question descriptions from Stackexchange that are duplicates to the given question. \\
        DBPedia (\textbf{DB}) & Given a query, retrieve relevant entity descriptions from DBPedia. \\
        FEVER (\textbf{FV}) & Given a claim, retrieve documents that support or refute the claim. \\
        FiQA2018 (\textbf{FQA}) & Given a financial question, retrieve user replies that best answer the question. \\
        HotpotQA (\textbf{HQA}) & Given a multi-hop question, retrieve documents that can help answer the question. \\
        MSMARCO (\textbf{MS}) & Given a query, retrieve documents that answer the query. \\
        NFCorpus (\textbf{NF}) & Given a question, retrieve relevant documents that best answer the question. \\
        NQ (\textbf{NQ}) & Given a question, retrieve Wikipedia passages that answer the question. \\
        Quora (\textbf{QR}) & Given a question, retrieve questions that are semantically equivalent to the given question. \\
        SCIDOCS (\textbf{SD}) & Given a scientific paper title, retrieve paper abstracts that are cited by the given paper. \\
        SciFact (\textbf{SF}) & Given a scientific claim, retrieve documents that support or refute the claim. \\
        \bottomrule
    \end{tabular}}
\end{table*}
\begin{table*}[ht]
    \centering
    \caption{Task instructions used for evaluation on the MIRACL~\citep{zhang2023miracl} benchmark.}
    \label{tab:inst_miracle}
    \renewcommand\arraystretch{1.15}
    \resizebox{\textwidth}{!}{\begin{tabular}{p{0.52\textwidth}p{0.42\textwidth}}
        \toprule
        \multicolumn{1}{l}{\textbf{Task Name}} &
        \multicolumn{1}{l}{\textbf{Instruction Template}} \\
        \midrule
        \justifying Arabic (\textbf{ar}), Bengali (\textbf{bn}), German (\textbf{de}), English (\textbf{en}), Spanish (\textbf{es}), Persian (\textbf{fa}), Finnish (\textbf{fi}), French (\textbf{fr}), Hindi (\textbf{hi}), Indonesian (\textbf{id}), Japanese (\textbf{ja}), Korean (\textbf{ko}), Russian (\textbf{ru}), Swahili (\textbf{sw}), Telugu (\textbf{te}), Thai (\textbf{th}), Yoruba (\textbf{yo}), Chinese (\textbf{zh}) & Given a query, retrieve documents that answer the query. \\
        \bottomrule
    \end{tabular}}
\end{table*} 
\begin{table*}[ht]
    \centering
    \caption{Task instructions used for evaluation on the LMEB~\citep{DBLP:journals/corr/abs-2603-12572} benchmark.}
    \label{tab:inst_lmeb}
    \renewcommand\arraystretch{1.15}
    \resizebox{\textwidth}{!}{\begin{tabular}{p{0.4\textwidth}p{0.54\textwidth}}
        \toprule
        \multicolumn{1}{l}{\textbf{Task Name}} &
        \multicolumn{1}{l}{\textbf{Instruction Template}} \\
        \midrule
        \justifying ConvoMem, LoCoMo, LongMemEval, MemBench, REALTALK, TMD & We use the task instructions provided by LMEB~\citep{DBLP:journals/corr/abs-2603-12572} without modification. \\
        \bottomrule
    \end{tabular}}
\end{table*} 

\section{Task Instructions}
\label{app:inst}
Tables~\ref{tab:inst_beir}, \ref{tab:inst_miracle}, and \ref{tab:inst_lmeb} summarize the task-specific instructions used to evaluate models on BEIR, MIRACL, and LMEB, respectively.

\section{Performance Scaling with Test-time Compute}
\label{app:analysis}

\begin{figure}[t]
    \centering
    \begin{subfigure}[t]{0.49\linewidth}
        \centering
        \includegraphics[width=\linewidth]{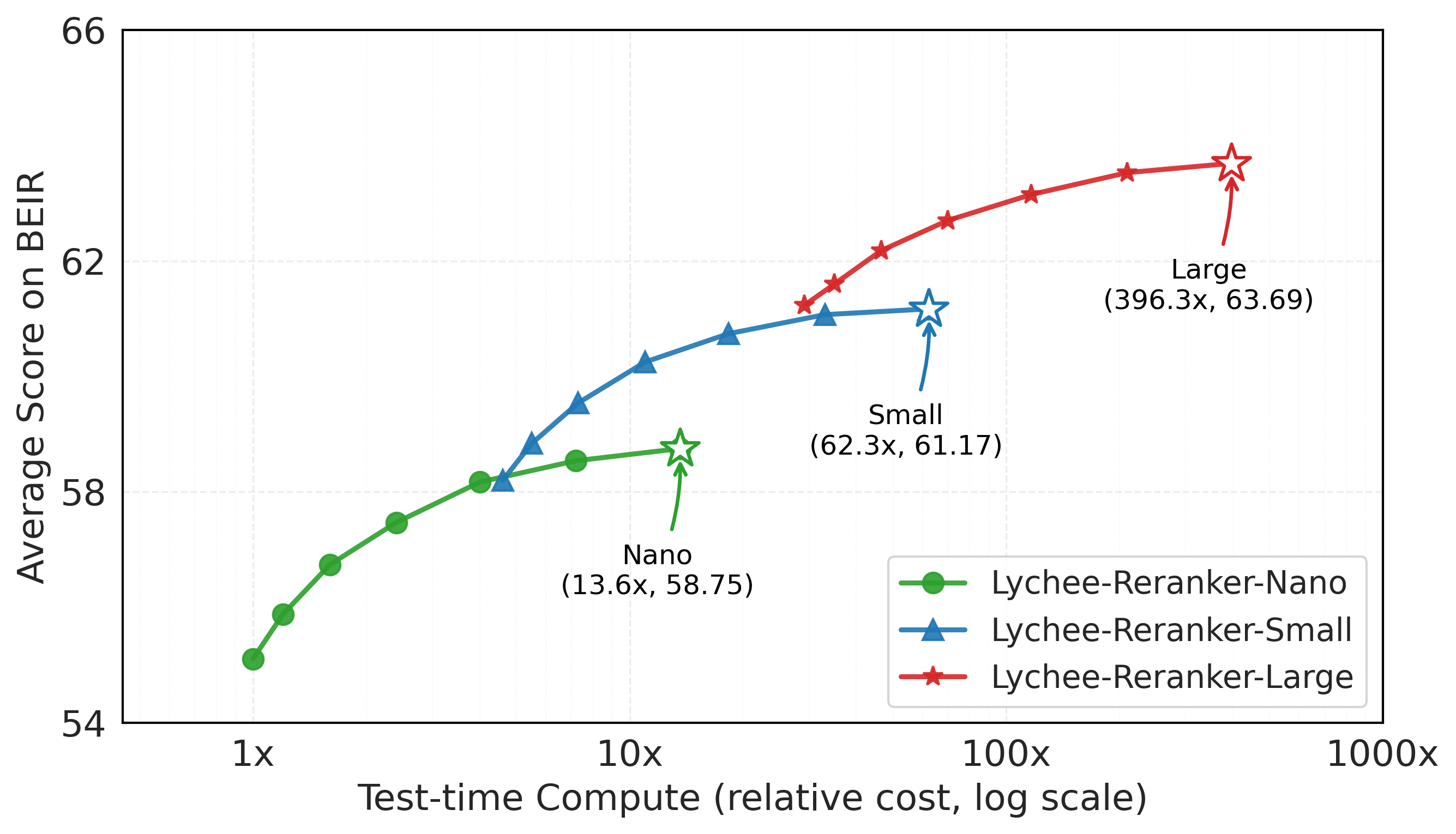}
        \caption{BEIR.}
        \label{fig:beir_scale}
        \vspace{0.05cm}
    \end{subfigure}
    \hfill
    \begin{subfigure}[t]{0.49\linewidth}
        \centering
        \includegraphics[width=\linewidth]{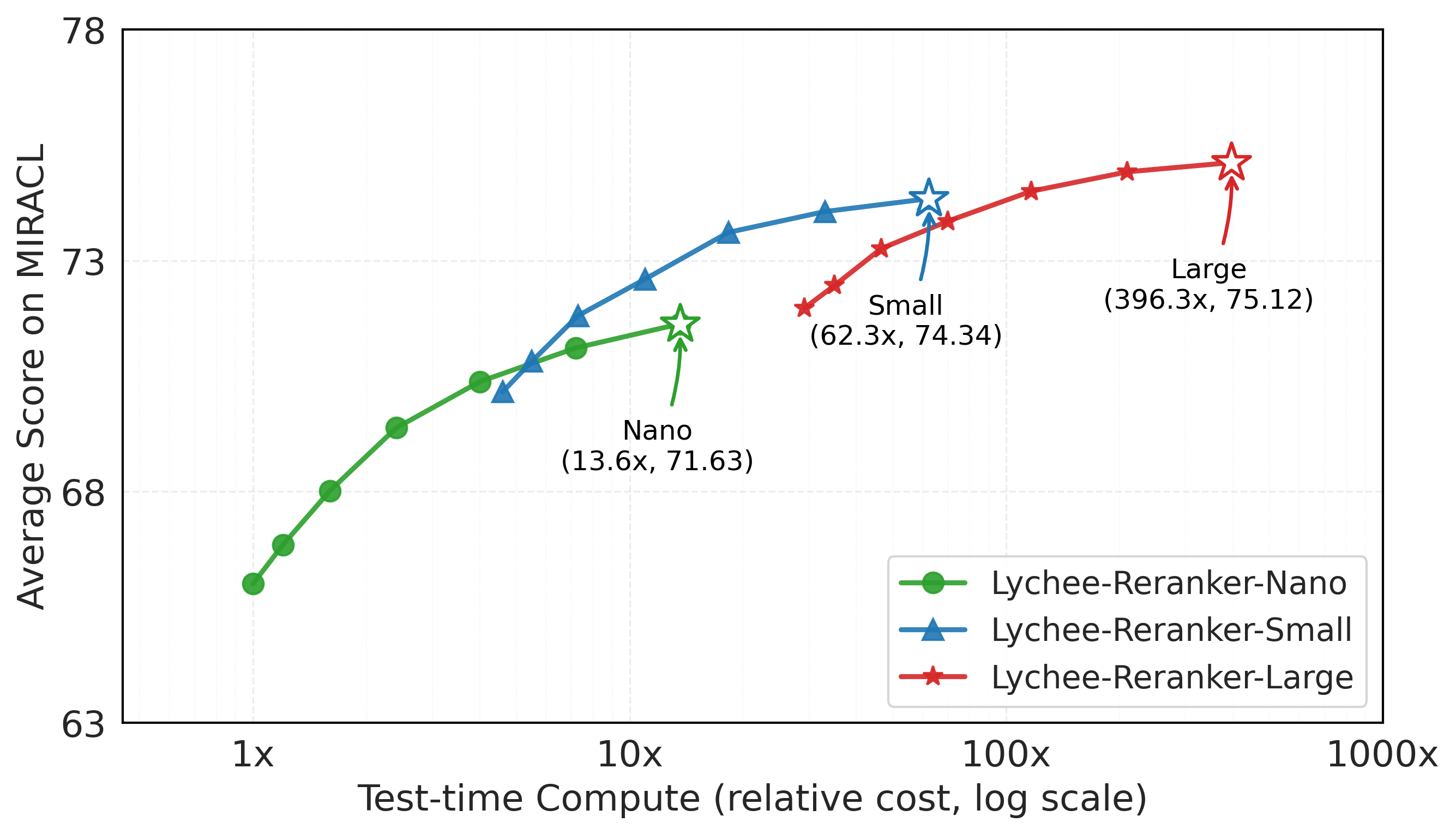}
        \caption{MIRACL.}
        \label{fig:miracle_scale}
        \vspace{0.05cm}
    \end{subfigure}
    \caption{Performance scaling of the \model\ series with test-time compute. Each curve shows how reranking performance changes as computational cost increases. Results are taken from Tables~\ref{tab:beir_mep_results} and~\ref{tab:miracle_mep_results}.}
    \label{fig:cost_perf_scale}
\end{figure}

We further analyze how performance scales with test-time compute across model sizes and compression ratios. 
Figure~\ref{fig:cost_perf_scale} summarizes the trends on BEIR and MIRACL. We draw the following main conclusions:
\noindent\textbf{Whether larger models provide a free lunch at similar cost depends on the application scenario.}
At comparable computational cost, larger models with stronger compression can improve performance on BEIR, but such gains are not consistently observed on MIRACL.
For example, \model\texttt{-Large} at $r=32$ incurs a lower computational cost than \model\texttt{-Small} at $r=2$ (46.6x vs. 62.3x) and achieves higher average nDCG@10 on BEIR (62.18 vs. 61.17), but lower performance on MIRACL (73.26 vs. 74.34).
This suggests that model size and compression ratio should be chosen jointly according to the application scenario, performance requirements, and available compute budget.
\noindent\textbf{Performance exhibits diminishing returns as computation increases.}
For a fixed model size, increasing test-time compute by reducing the compression ratio generally improves reranking performance, although the marginal gains diminish. 
Across both BEIR and MIRACL, using $r=4$ or $r=8$ often provides a favorable trade-off, achieving performance close to that at $r=2$ at substantially lower cost.
As the performance curve flattens, switching to a larger model can offer greater gains than allocating additional compute to the same model.
\noindent\textbf{Smaller models are more sensitive to compression.}
Smaller models exhibit larger performance drops as the compression ratio increases.
For example, as $r$ increases from 2 to 128 on BEIR, the average nDCG@10 of \model\texttt{-Nano} drops by 3.64 points, from 58.75 to 55.11, compared with a 2.46 points drop for \model\texttt{-Large}, from 63.69 to 61.23.
These results suggest that smaller models rely more heavily on retaining passage information, while larger models are more robust to compression.

\begin{table*}[t]
    \centering
    \caption{Reranking results measured by nDCG@10 on the six dialogue memory retrieval tasks in LMEB~\citep{DBLP:journals/corr/abs-2603-12572}. ``N'', ``S'', and ``L'' denote Nano, Small, and Large, respectively. Avg. denotes the average performance across the six tasks. All \model\ variants use a compression ratio of $r=4$. Within each parameter group, the best results are \textbf{boldfaced}, and the second-best results are \underline{underlined}. Other settings follow those in Table~\ref{tab:main_beir_results}.}
    \label{tab:main_lmeb_results}
    \vspace{0.1cm}
    \renewcommand\arraystretch{1.0}
    \setlength{\tabcolsep}{4pt}
    \resizebox{\textwidth}{!}{\begin{tabular}{lccc|cccccc}
        \toprule
        \multicolumn{1}{l}{\textbf{Models}} &
        \multicolumn{1}{c}{\textbf{Size}} &
        \multicolumn{1}{c}{\textbf{Cost}} &
        \multicolumn{1}{c|}{\textbf{Avg.}} &
        \multicolumn{1}{c}{\textbf{ConvoMem}} &
        \multicolumn{1}{c}{\textbf{LoCoMo}} &
        \multicolumn{1}{c}{\textbf{LongMemEval}} &
        \multicolumn{1}{c}{\textbf{MemBench}} &
        \multicolumn{1}{c}{\textbf{REALTALK}} &
        \multicolumn{1}{c}{\textbf{TMD}} \\
        \midrule
        \multicolumn{10}{c}{\textbf{First-stage Retriever}} \\ \midrule
        KaLM-Embedding-V2.5 & 0.5B & -- & 50.80 & 62.74 & 41.88 & 75.18 & 69.59 & 38.60 & 16.82 \\ \midrule
        \multicolumn{10}{c}{\textbf{Second-stage Reranker}} \\ \midrule
        \rowcolor{gray!10}
        \multicolumn{10}{l}{\emph{Models with more than 4B parameters}} \\
        Qwen3-Reranker-8B & 8B & 359.8x & 66.42 & 64.64 & 65.91 & 78.93 & 76.31 & 57.13 & 55.58 \\
        \midrule
        \rowcolor{gray!10}
        \multicolumn{10}{l}{\emph{Models with 1B--4B parameters}} \\
        Qwen3-Reranker-4B & 4B & 157.8x & 62.86 & 66.68 & \underline{63.43} & 53.72 & \underline{78.65} & \underline{55.91} & \textbf{58.79} \\
        \model\texttt{-L} & 4B & \textbf{29.1x} & \underline{65.71} & \underline{70.34} & \textbf{64.80} & \textbf{89.54} & 77.99 & \textbf{56.68} & 34.94 \\
        bge-reranker-v2-gemma & 2.5B & 54.2x & 63.50 & 67.12 & 61.30 & \underline{87.23} & 76.50 & 54.74 & 34.13 \\
        mxbai-rerank-large-v2 & 1.5B & {52.8x} & \textbf{66.08} & \textbf{70.41} & 61.33 & 87.16 & \textbf{81.18} & 54.43 & \underline{41.96} \\
        \midrule
        \rowcolor{gray!10}
        \multicolumn{10}{l}{\emph{Models with 0.5B--1B parameters}} \\
        \model\texttt{-S} & 1B & \textbf{4.6x} & \underline{63.65} & 68.41 & \textbf{62.51} & \textbf{86.66} & 75.41 & \textbf{54.05} & 34.87 \\
        bge-reranker-large & 0.6B & 24.2x & 60.64 & 62.55 & 58.94 & 84.10 & 71.48 & 51.50 & \underline{35.30} \\
        bge-reranker-v2-m3 & 0.6B & 24.2x & 62.38 & \underline{69.15} & 60.20 & 82.39 & \underline{76.82} & \underline{53.90} & 31.84 \\
        Qwen3-Reranker-0.6B & 0.6B & 28.2x & \textbf{63.69} & 67.36 & \underline{60.32} & 84.91 & 75.67 & 53.26 & \textbf{40.63} \\
        mxbai-rerank-base-v2 & 0.5B & 19.9x & 62.47 & \textbf{69.41} & 55.76 & \underline{86.29} & \textbf{79.97} & 49.75 & 33.61 \\
        \midrule
        \rowcolor{gray!10}
        \multicolumn{10}{l}{\emph{Models with fewer than 0.5B parameters}} \\
        gte-reranker-base & 0.3B & 8.0x & 56.08 & \underline{54.78} & 51.01 & 78.94 & 68.76 & 47.97 & \underline{35.02} \\
        jina-reranker-v2 & 0.3B & 8.0x & \underline{59.60} & 54.70 & \underline{56.22} & \underline{86.55} & \underline{72.66} & \textbf{51.53} & \textbf{35.97} \\
        \model\texttt{-N} & 0.27B & \textbf{1.0x} & \textbf{61.07} & \textbf{64.89} & \textbf{58.44} & \textbf{86.66} & \textbf{74.11} & \underline{50.89} & 31.44 \\
        \bottomrule
    \end{tabular}}
\end{table*}
\begin{table*}[t]
    \centering
    \caption{Embedding-based retrieval results (nDCG@10) on the six dialogue memory retrieval tasks in LMEB~\citep{DBLP:journals/corr/abs-2603-12572}. Results are taken from the online \href{https://mteb-leaderboard.hf.space/benchmark/LMEB}{LMEB leaderboard}, accessed September 8, 2026.}
    \label{tab:main_lmeb_emb_results}
    \renewcommand\arraystretch{1.05}
    \setlength{\tabcolsep}{4pt}
    \resizebox{\textwidth}{!}{\begin{tabular}{lcc|cccccc}
        \toprule
        \multicolumn{1}{l}{\textbf{Models}} &
        \multicolumn{1}{c}{\textbf{Size}} &
        \multicolumn{1}{c|}{\textbf{Avg.}} &
        \multicolumn{1}{c}{\textbf{ConvoMem}} &
        \multicolumn{1}{c}{\textbf{LoCoMo}} &
        \multicolumn{1}{c}{\textbf{LongMemEval}} &
        \multicolumn{1}{c}{\textbf{MemBench}} &
        \multicolumn{1}{c}{\textbf{REALTALK}} &
        \multicolumn{1}{c}{\textbf{TMD}} \\
        \midrule
        KaLM-Embedding-Gemma3~\citep{zhao2026kalmembeddingv} & 12B & 56.66 & 64.09 & 42.61 & 80.97 & 73.82 & 44.51 & 33.95 \\ 
        bge-multilingual-gemma~\citep{chen2024bge} & 9B & 59.30 & 66.80 & 56.02 & 81.66 & 72.47 & 48.15 & 30.69 \\ 
        Nemotron-3-Embed-8B-BF16~\citep{babakhin2026nemotron} & 8B & 59.86 & 69.04 & 56.51 & 85.70 & 76.45 & 45.70 & 25.74 \\
        Qwen3-Embedding-8B~\citep{zhang2025qwen3} & 8B & 51.67 & 55.60 & 36.00 & 77.66 & 71.10 & 40.71 & 28.96 \\
        F2LLM-v2-8B~\citep{zhang2026f2llm} & 8B & 55.26 & 66.48 & 42.29 & 81.46 & 70.97 & 40.56 & 29.79 \\
        SFR-Embedding-Mistral~\citep{SFRAIResearch2024} & 7B & 56.38 & 67.84 & 50.05 & 79.75 & 70.19 & 45.27 & 25.17 \\
        gte-Qwen2-7B-instruct~\citep{li2023gte} & 7B & 56.48 & 64.42 & 51.55 & 80.45 & 72.36 & 46.25 & 23.86 \\
        NV-Embed-v2~\citep{lee2025nv} & 7B & 56.02 & 71.05 & 51.24 & 75.76 & 70.83 & 48.09 & 19.13 \\ 
        SFR-Embedding-2\_R~\citep{SFR-embedding-2} & 7B & 55.10 & 69.42 & 52.99 & 67.96 & 69.99 & 46.12 & 24.10 \\
        e5-mistral-7b-instruct~\citep{wang2022text,wang2024improving} & 7B & 55.31 & 68.63 & 46.82 & 78.21 & 68.99 & 42.87 & 26.36 \\ 
        \bottomrule
    \end{tabular}}
    \vspace{-0.2cm}
\end{table*}

\section{Results on LMEB}
\label{app:lmeb}
Agentic memory is increasingly important for LLM-based agents, as systems such as OpenClaw~\citep{openclaw2026}, OpenAI Codex~\citep{openai2025codex}, and Claude Code~\citep{anthropic2025claudecode} retrieve and use stored information to maintain contextual continuity and support long-horizon tasks.
Motivated by this, we further evaluate \model\ on the six dialogue memory retrieval tasks in LMEB~\citep{DBLP:journals/corr/abs-2603-12572}, a benchmark for long-horizon memory retrieval.
The results in Tables~\ref{tab:main_lmeb_results} and~\ref{tab:main_lmeb_emb_results} reveal the following key findings:
\textbf{Reranking is particularly important for memory retrieval.} \model\texttt{-Nano}, \model\texttt{-Small}, and \model\texttt{-Large} improve over the first-stage retriever by 10.27, 12.85, and 14.91 nDCG@10 points on LMEB, respectively, compared with gains of 4.76, 7.29, and 9.75 points on BEIR. 
These consistently larger gains on LMEB highlight the value of reranking for long-horizon memory retrieval.
\textbf{Memory reranking remains challenging.} Across the six dialogue memory reranking tasks, no reranker achieves the best result on all tasks. Each model shows weaknesses on some tasks. 
For example, Qwen3-Reranker-8B achieves an nDCG@10 of 78.93 on LongMemEval, trailing the smaller \model\texttt{-Large} (4B), which scores 89.54.
This indicates the need for further improvements in memory retrieval.
\textbf{Reranking is especially effective for complex queries.} Specifically, TMD contains many temporal queries involving relative or absolute time expressions.
Such queries are challenging for embedding models, but rerankers handle them more effectively.
For example, Qwen3-Reranker-4B achieves an nDCG@10 of 58.79 on TMD, compared with 16.82 for the first-stage retriever.
\textbf{The retrieve-then-rerank pipeline can be more effective than increasing embedding model size.}
As shown in Table~\ref{tab:main_lmeb_emb_results}, bge-multilingual-gemma (9B) and Nemotron-3-Embed-8B-BF16 (8B) achieve average nDCG@10 scores of 59.30 and 59.86, respectively.
In contrast, pairing KaLM-Embedding-V2.5 (0.5B) with \model\texttt{-Nano} (0.27B activated parameters) achieves an average nDCG@10 of 61.07, outperforming both of these larger embedding models.
These results highlight the advantage of the retrieve-then-rerank pipeline over simply increasing embedding model size and illustrate the complementary roles of retrieval and reranking.
\begin{figure}[t]
    \centering
    \includegraphics[width=0.75\linewidth]{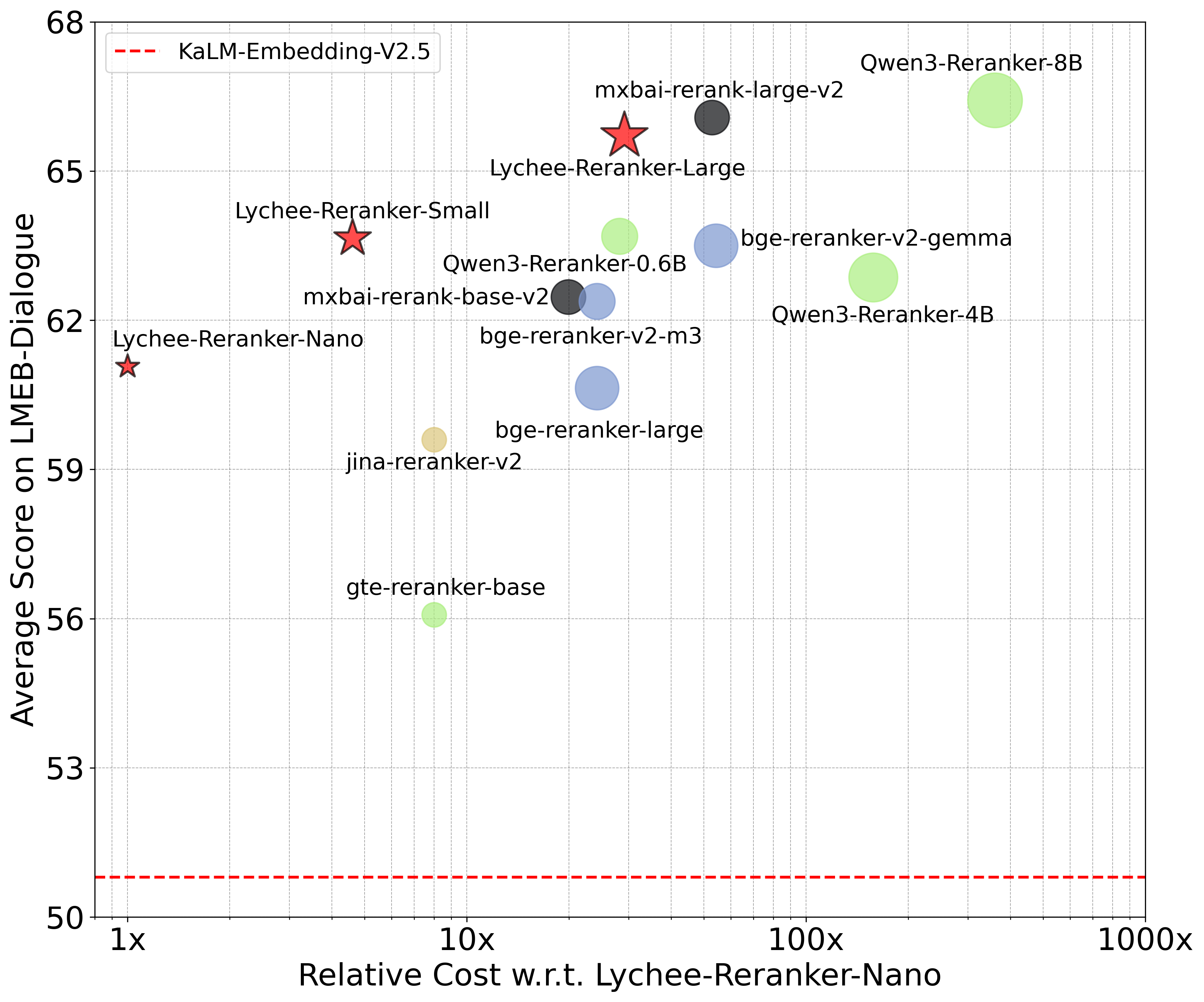}
    \caption{Comparison between the \model\ series and other reranking models on LMEB-Dialogue~\citep{DBLP:journals/corr/abs-2603-12572} in terms of reranking performance and relative online computation cost. The cost is estimated following the analysis in Appendix~\ref{app:time_complexity}; the x-axis is plotted on a logarithmic scale. Marker sizes are proportional to model sizes. The results are taken from Table~\ref{tab:main_lmeb_results}.}
    \label{fig:lmeb_scatter}
\end{figure}
\section{Understanding Reranking Performance Degradation under Compression}
\label{app:in_depth}
\begin{figure}[t]
    \centering
    \begin{subfigure}[t]{1.0\linewidth}
        \centering
        \includegraphics[width=\linewidth]{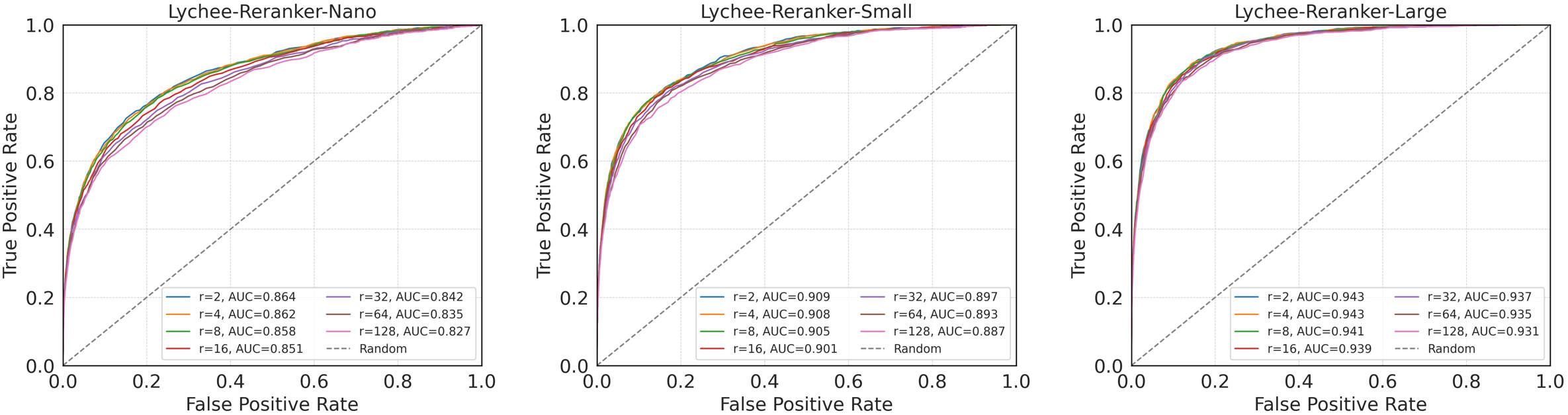}
        \caption{FQA on BEIR.}
        \label{fig:fiqa_auc}
        \vspace{0.05cm}
    \end{subfigure}
    \begin{subfigure}[t]{1.0\linewidth}
        \centering
        \includegraphics[width=\linewidth]{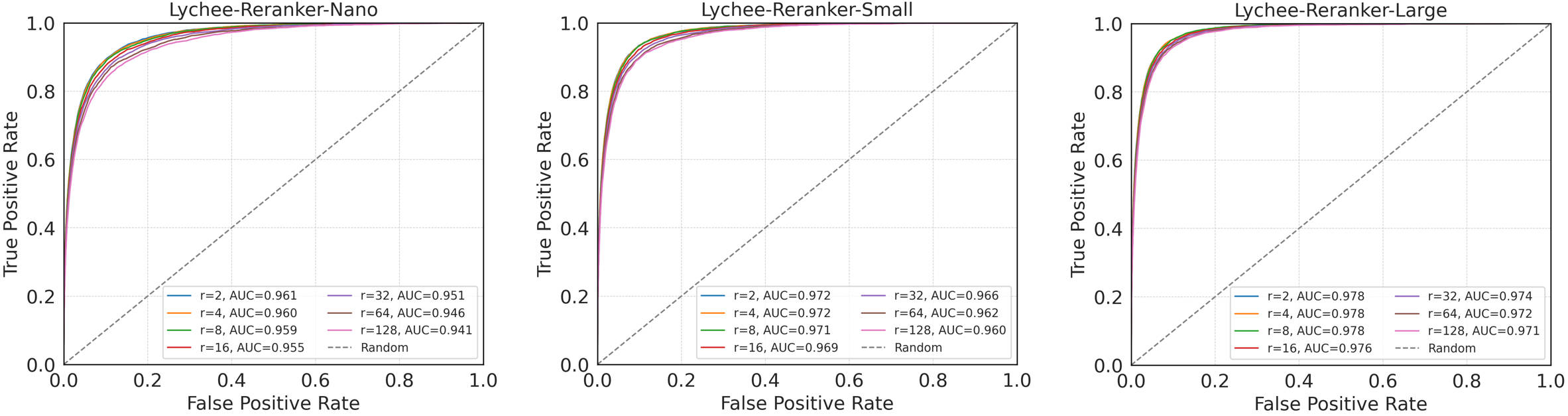}
        \caption{NQ on BEIR.}
        \label{fig:nq_auc}
        \vspace{0.05cm}
    \end{subfigure}
    \begin{subfigure}[t]{1.0\linewidth}
        \centering
        \includegraphics[width=\linewidth]{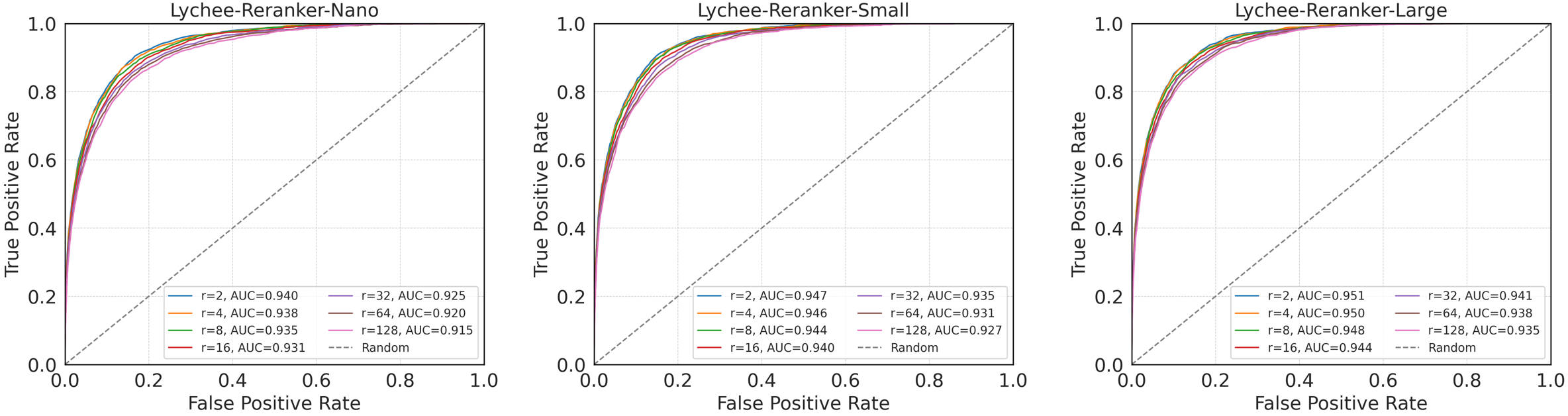}
        \caption{English (en) on MIRACL.}
        \label{fig:en_auc}
        \vspace{0.05cm}
    \end{subfigure}
    \begin{subfigure}[t]{1.0\linewidth}
        \centering
        \includegraphics[width=\linewidth]{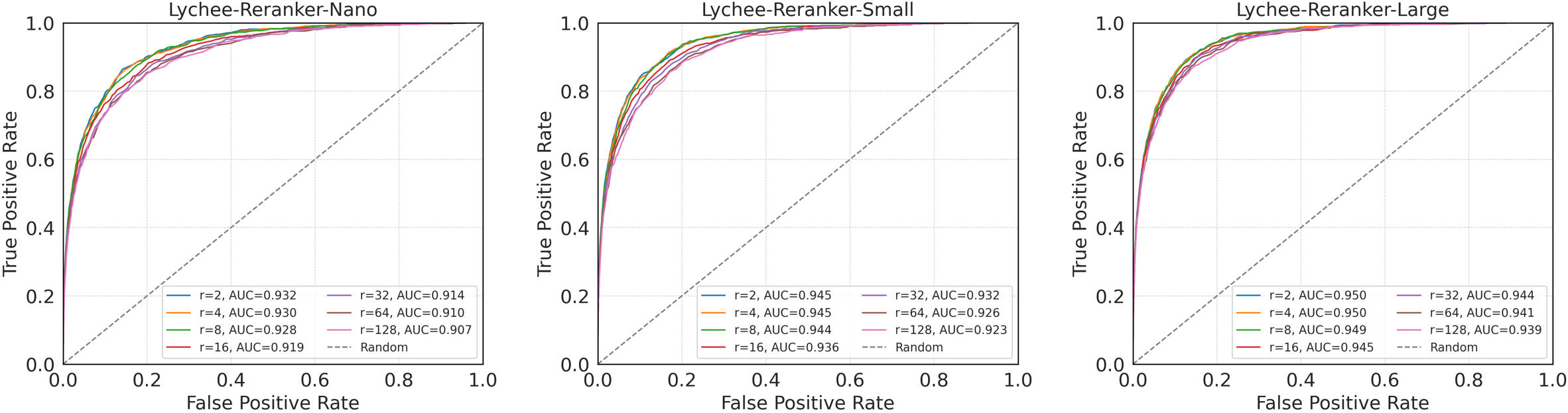}
        \caption{Chinese (zh) on MIRACL.}
        \label{fig:zh_auc}
    \end{subfigure}
    \caption{ROC curves for the \model\ series on representative tasks from BEIR (FQA and NQ) and MIRACL (English and Chinese) under Matryoshka embedding pooling compression ratios $r\in\{2,4,8,16,32,64,128\}$. Curves are computed over the top-100 candidates scored by each reranker. Higher AUC values indicate better discrimination between relevant and irrelevant passages.}
    \label{fig:auc}
\end{figure}

To better understand how compression affects reranking performance, we plot ROC curves on representative tasks using the scores assigned by each reranker to the top-100 candidates for each query (Figure~\ref{fig:auc}).
ROC-AUC measures how well a reranker's relevance scores distinguish relevant passages from irrelevant ones without relying on a specific ranking cutoff.
As $r$ increases, AUC values generally decrease, with more noticeable declines for smaller models.
For example, on FQA, as $r$ increases from 2 to 128, the AUC drops from 0.864 to 0.827 for \model\texttt{-Nano}, from 0.909 to 0.887 for \model\texttt{-Small}, and from 0.943 to 0.931 for \model\texttt{-Large}.
This indicates that higher compression ratios cause a greater decline in smaller models' ability to distinguish relevant passages from irrelevant ones.
The AUC trends suggest two broad groups of compression ratios: AUC values remain relatively stable from $r=2$ to $r=32$, but decline more noticeably at $r=64$ and $r=128$.
These results are consistent with the findings in \S\ref{sec:matryoshka}: \model\ largely preserves its discriminative ability under moderate compression, whereas overly aggressive compression leads to noticeable information loss and performance degradation.
These findings motivate the development of more effective compression methods that better preserve information for relevance modeling at high compression ratios.

\section{Ablation Study on Matryoshka Embedding Pooling}
\label{app:mep_ablation}
\begin{figure}[t]
    \centering
    \begin{subfigure}[t]{0.475\linewidth}
        \centering
        \includegraphics[width=\linewidth]{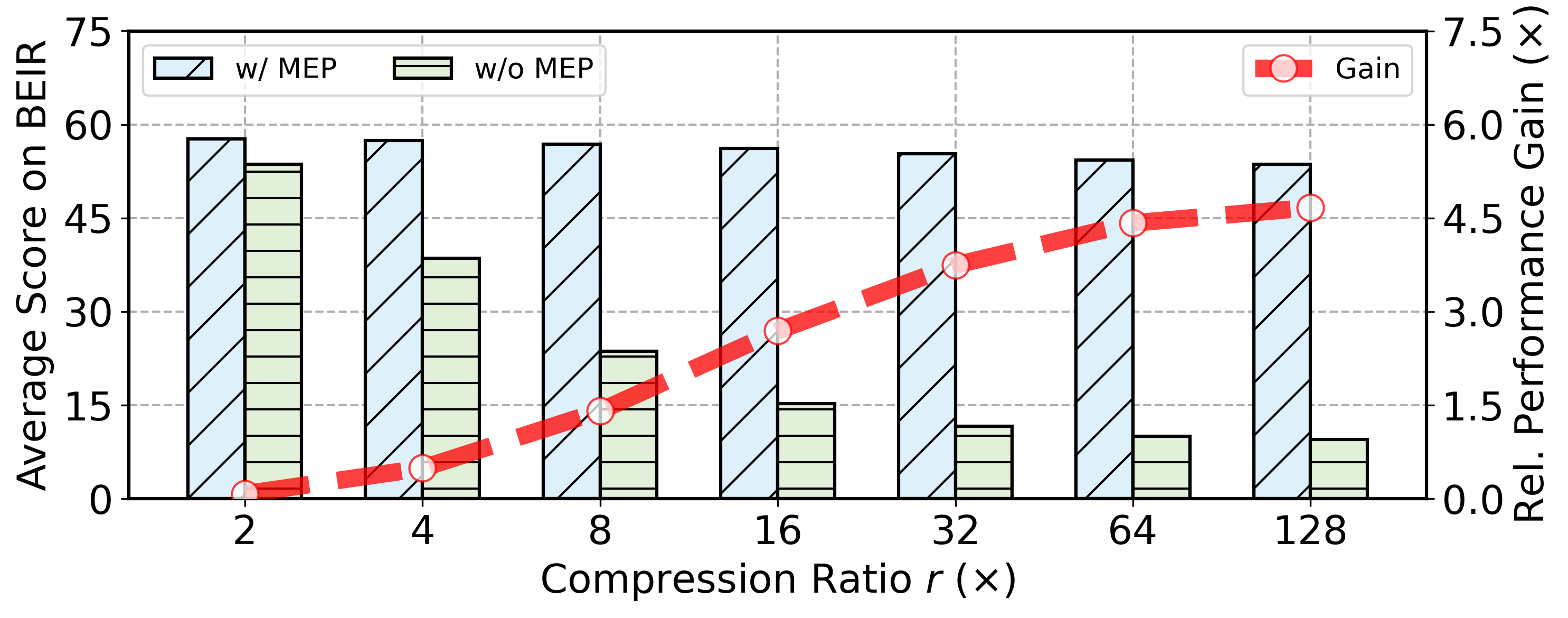}
        \caption{Stage~1: BEIR.}
        \label{fig:stage1_mep_ablation_beir}
    \end{subfigure}
    \hfill
    \begin{subfigure}[t]{0.475\linewidth}
        \centering
        \includegraphics[width=\linewidth]{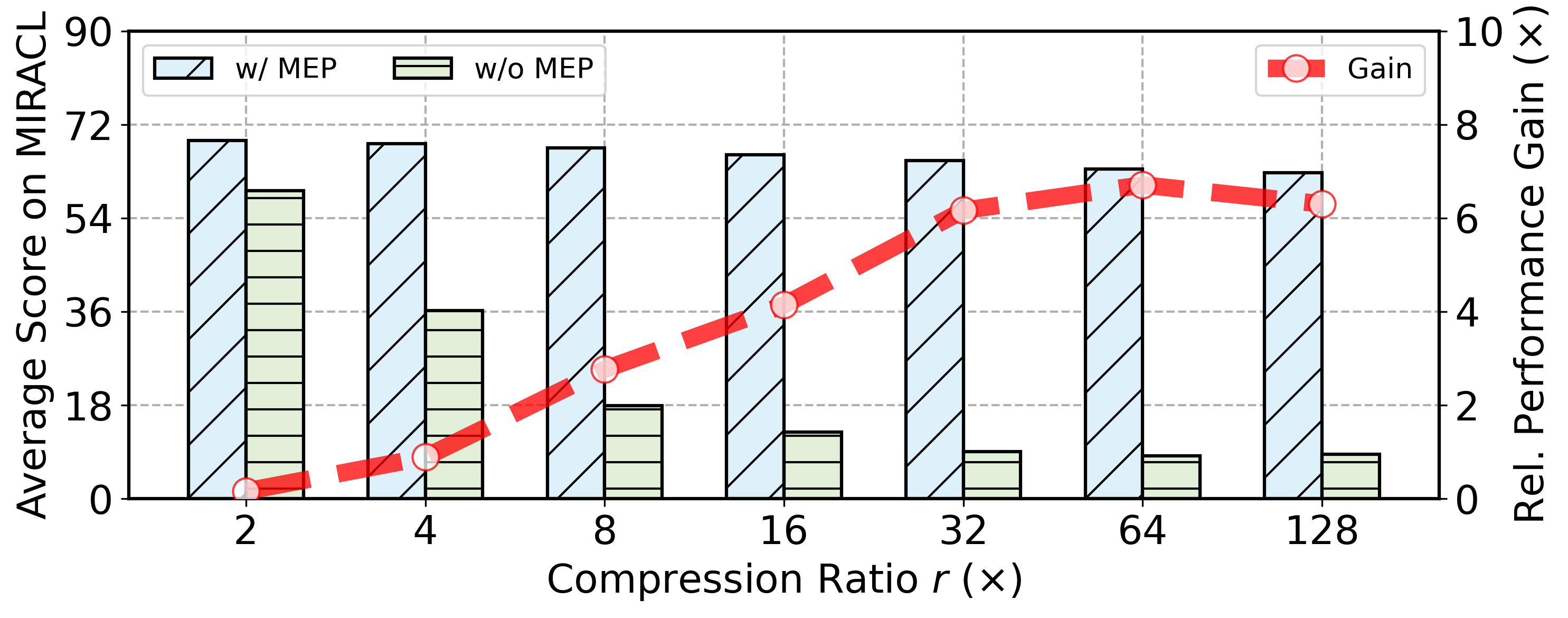}
        \caption{Stage~1: MIRACL.}
        \label{fig:stage1_mep_ablation_miracl}
    \end{subfigure}
    \begin{subfigure}[t]{0.475\linewidth}
        \centering
        \includegraphics[width=\linewidth]{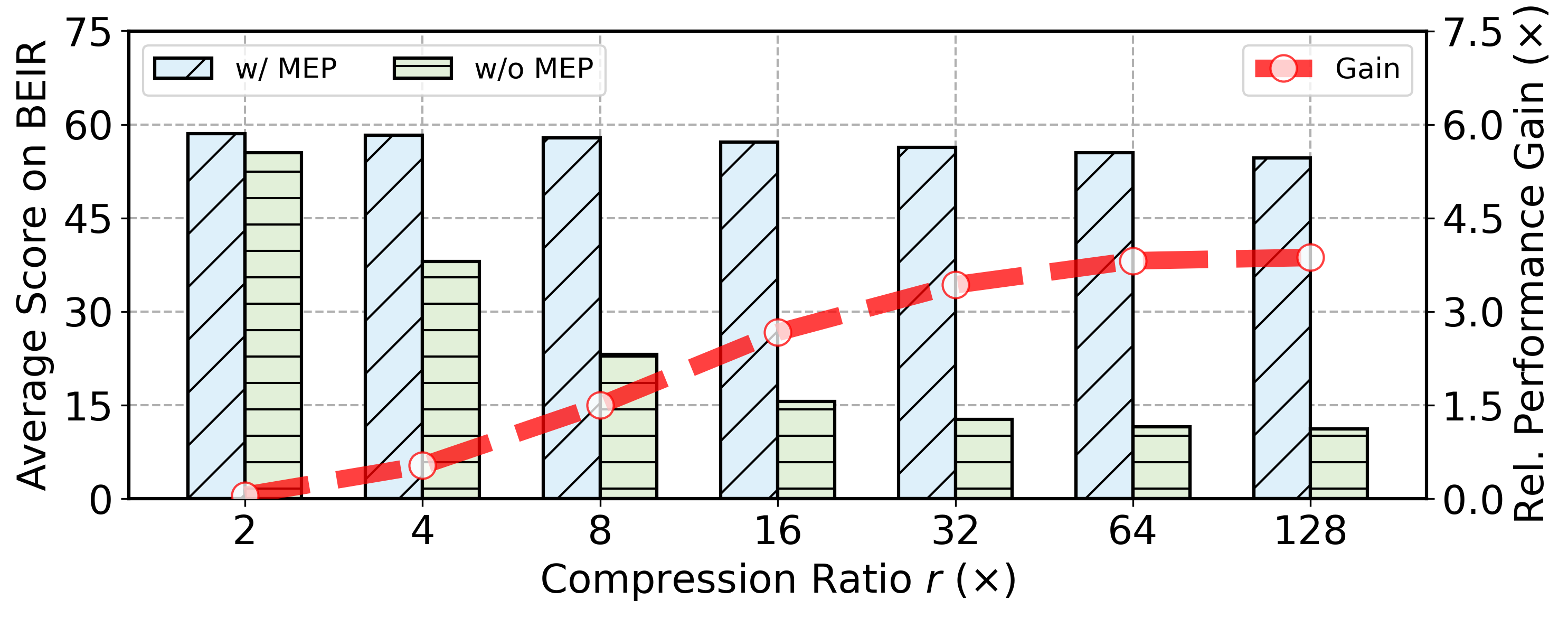}
        \caption{Stage~2: BEIR.}
        \label{fig:stage2_mep_ablation_beir}
    \end{subfigure}
    \hfill
    \begin{subfigure}[t]{0.475\linewidth}
        \centering
        \includegraphics[width=\linewidth]{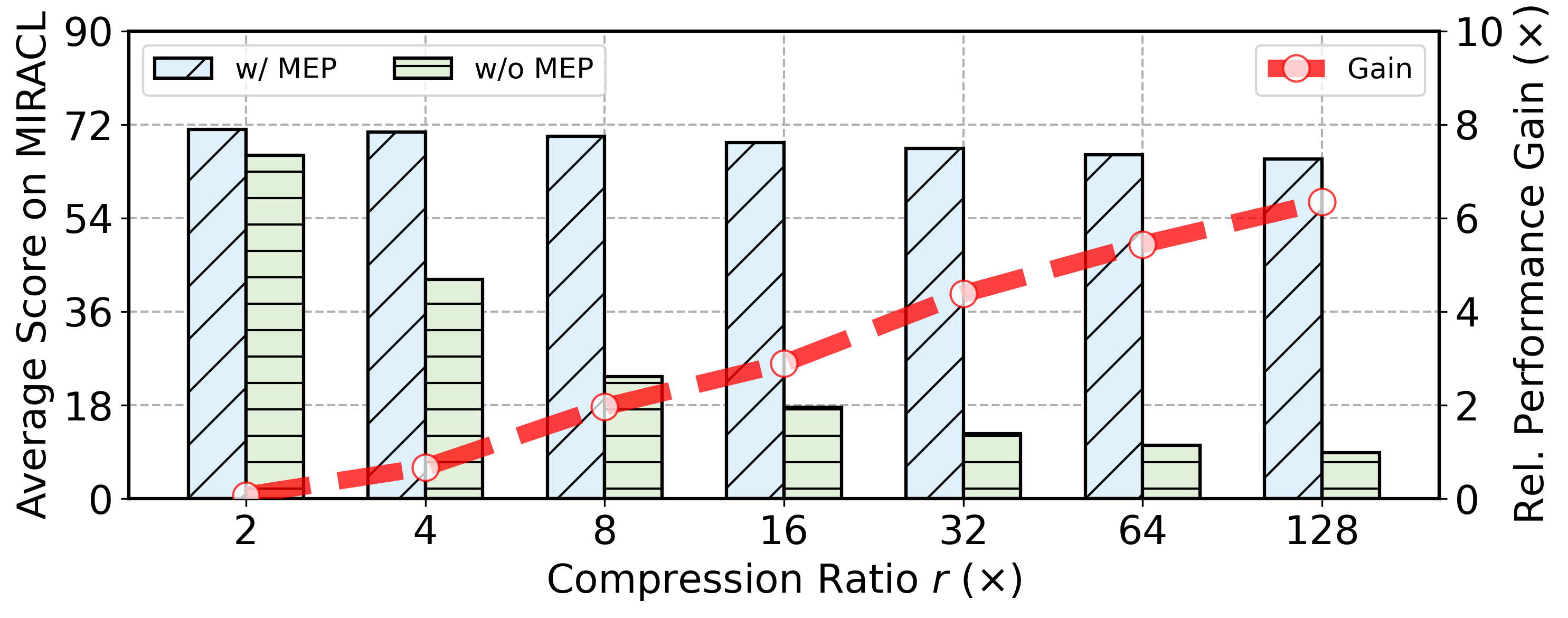}
        \caption{Stage~2: MIRACL.}
        \label{fig:stage2_mep_ablation_miracl}
    \end{subfigure}
    \begin{subfigure}[t]{0.475\linewidth}
        \centering
        \includegraphics[width=\linewidth]{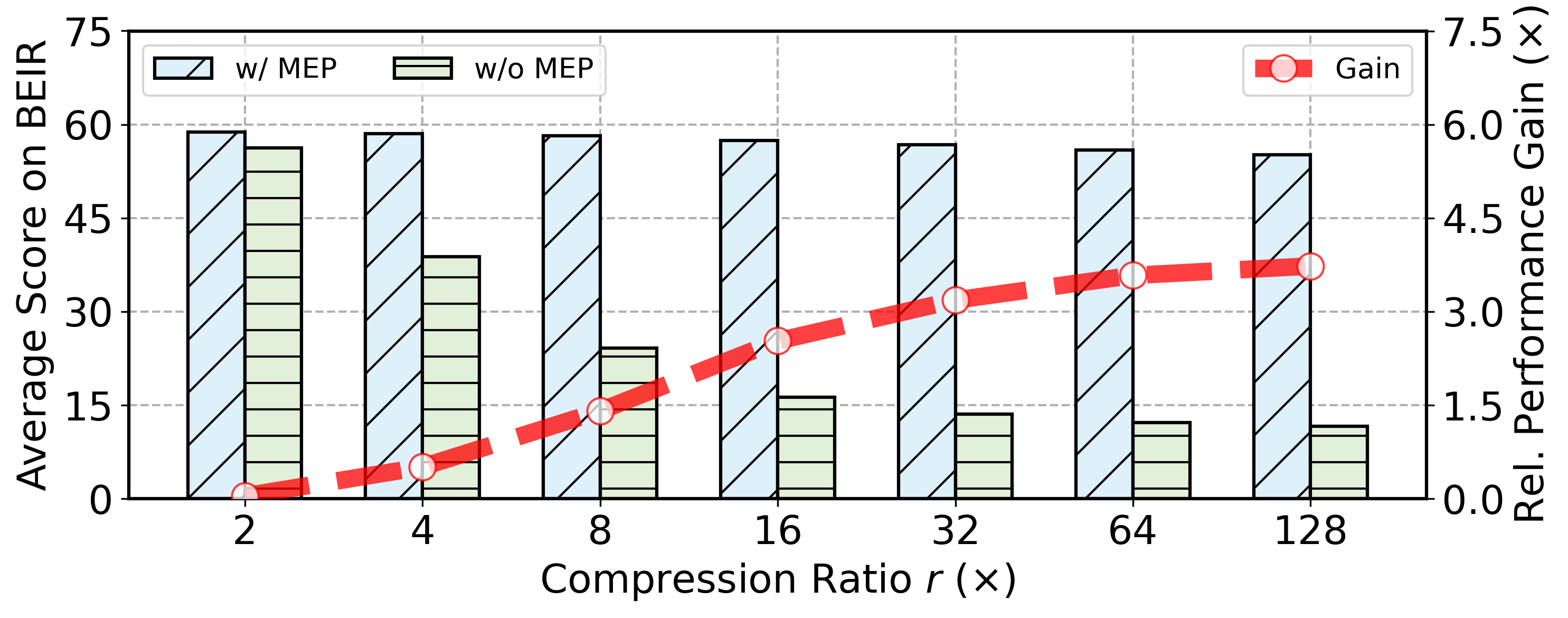}
        \caption{Stage~3: BEIR.}
        \label{fig:stage3_mep_ablation_beir}
    \end{subfigure}
    \hfill
    \begin{subfigure}[t]{0.475\linewidth}
        \centering
        \includegraphics[width=\linewidth]{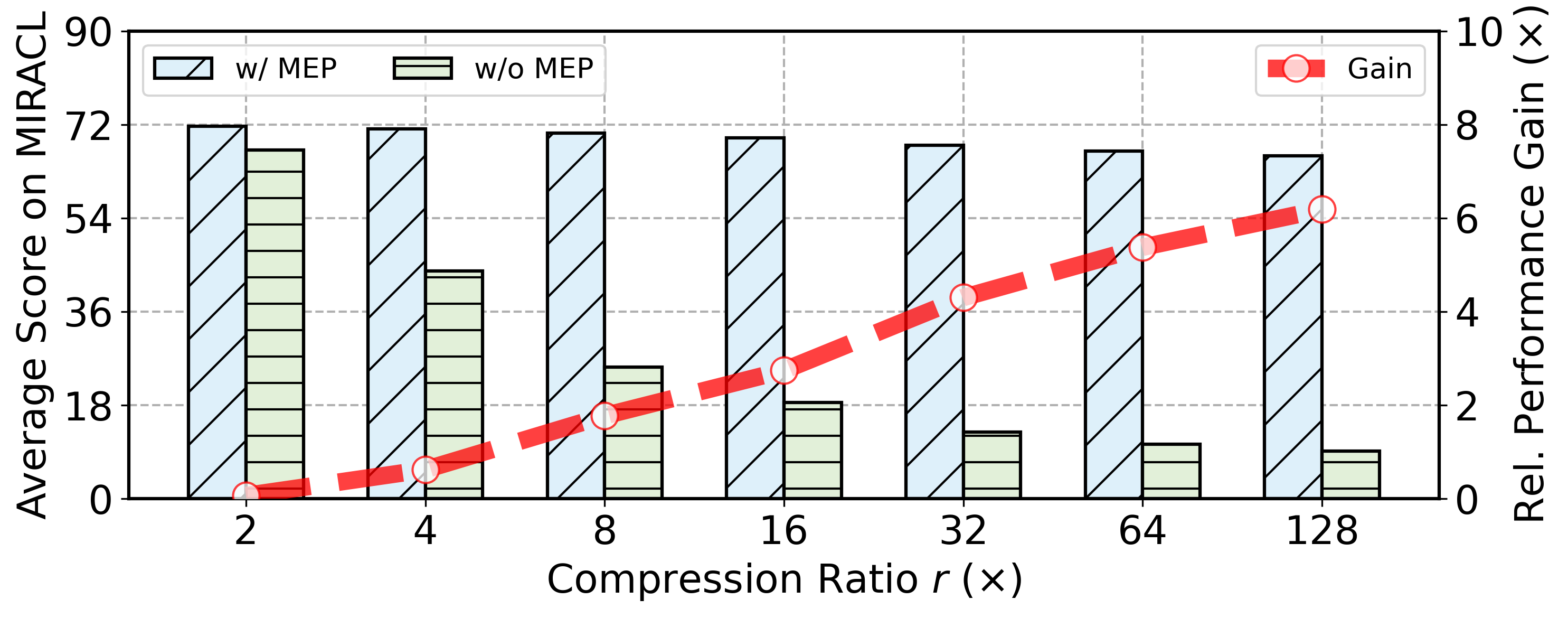}
        \caption{Stage~3: MIRACL.}
        \label{fig:stage3_mep_ablation_miracl}
    \end{subfigure}
    \caption{Ablation study on Matryoshka embedding pooling using \model\texttt{-Nano} on BEIR (left) and MIRACL (right). From top to bottom, the rows show results after Stages~1, 2, and 3, respectively. ``w/ MEP'' denotes training with MEP using compression ratios $\mathcal{R}=\{2,8,32,128\}$, whereas ``w/o MEP'' denotes training only at $r=1$ (no compression). Results for ``w/ MEP'' are taken from Table~\ref{tab:multistage_ablation}. ``Rel.'' denotes the relative performance gain, computed as $(S_{\text{w/ MEP}}-S_{\text{w/o MEP}})/S_{\text{w/o MEP}}$ at the same compression ratio, where $S$ denotes the average nDCG@10 score.}
    \label{fig:mep_training_ablation}
\end{figure}

Figure~\ref{fig:mep_training_ablation} compares the performance of \model\texttt{-Nano} trained with and without MEP on BEIR and MIRACL across different compression ratios.
Results are reported after each stage of the multi-stage training pipeline described in \S\ref{sec:multi_stage}.
For Stage~2, both settings use checkpoints trained with a LoRA rank of 64 and an alpha of 32.
Except for whether MEP is used during training, the two settings use identical model initialization, training data, number of training steps, and all other hyperparameters.
The relative performance gains from training with MEP generally increase with the compression ratio, reaching severalfold improvements over training without MEP at high compression ratios.
Specifically, the model trained without MEP maintains reasonable performance at the low compression ratio $r=2$, but still substantially underperforms the model trained with MEP when both are evaluated at $r=2$.
The performance of the model trained without MEP plummets when the compression ratio increases to $r=4$.
For example, on BEIR after Stage~1, increasing the compression ratio from $r=2$ to $r=4$ reduces the average nDCG@10 of the model trained without MEP from 53.58 to 38.57, a decrease of 15.01 points.
In contrast, the performance of the model trained with MEP declines very slowly as the compression ratio increases.
Overall, these results indicate that training with MEP not only substantially improves the model's robustness to compression but also yields higher peak performance at low compression ratios.

\begin{table*}[ht]
    \centering
    \caption{\model\ training datasets. ``en'', ``zh'', and ``ml'' denote English, Chinese, and multilingual data, respectively. For BGE-M3-data, we primarily use samples with lengths ranging from 0 to 500.}
    \label{tab:training_data}
    \renewcommand\arraystretch{1.45}
    \resizebox{\textwidth}{!}{\begin{tabular}{lcll}
        \toprule
        \multicolumn{1}{l}{\textbf{Source}} &
        \multicolumn{1}{l}{\textbf{Language}} &
        \multicolumn{1}{l}{\textbf{Size}} &
        \multicolumn{1}{l}{\textbf{URL}} \\
        \midrule
        \rowcolor{gray!10}
        \multicolumn{4}{c}{\textbf{KaLM embedding fine-tuning data (retrieval subset)}} \\
        \addlinespace[2pt]
        AdvertiseGen~\citep{shao2019long} & zh & 17,526 & \url{https://huggingface.co/datasets/shibing624/AdvertiseGen} \\
        CHEF~\citep{hu2022chef} & zh & 4,824 & \url{https://github.com/THU-BPM/CHEF} \\
        CodeFeedback~\citep{zheng2024opencodeinterpreter} & en & 49,090 & \url{https://huggingface.co/datasets/m-a-p/CodeFeedback-Filtered-Instruction} \\
        DRCD~\citep{shao2018drcd} & zh & 4,714 & \url{https://huggingface.co/datasets/voidful/DRCD} \\
        Expertqa~\citep{malaviya2024expertqa} & en & 1,252 & \url{https://github.com/chaitanyamalaviya/ExpertQA} \\
        GooAQ~\citep{khashabi2021gooaq} & en & 49,833 & \url{https://github.com/allenai/gooaq} \\
        LCSTS~\citep{hu2015lcsts} & zh & 19,535 & \url{https://huggingface.co/datasets/hugcyp/LCSTS} \\
        MEDI2BGE~\citep{xiao2024c} & en & 71,790 & \url{https://hf.co/datasets/GritLM/MEDI2BGE} \\
        Multi-CPR~\citep{long2022multi} & zh & 234,587 & \url{https://github.com/Alibaba-NLP/Multi-CPR} \\
        OpenOrca~\citep{lian2023openorca} & en & 38,623 & \url{https://huggingface.co/datasets/Open-Orca/OpenOrca} \\
        PAQ~\citep{lewis2021paq} & en & 49,849 & \url{https://huggingface.co/datasets/sentence-transformers/paq} \\
        PubMedQA~\citep{jin2019pubmedqa} & en & 79,954 & \url{https://huggingface.co/datasets/qiaojin/PubMedQA} \\
        RefGPT~\citep{yang2023refgpt} & zh & 49,896 & \url{https://github.com/sufengniu/RefGPT} \\
        SearchQA~\citep{dunn2017searchqa} & en & 9,988 & \url{https://huggingface.co/datasets/kyunghyuncho/search_qa} \\
        T2Ranking~\citep{xie2023t2ranking} & zh & 188,606 & \url{https://huggingface.co/datasets/THUIR/T2Ranking} \\
        THUCNews~\citep{guo2016thuctc} & zh & 19,288 & \url{https://huggingface.co/datasets/SirlyDreamer/THUCNews} \\
        UMETRIP-QA~\citep{wang2022umetripqa} & zh & 2,537 & \url{https://aistudio.baidu.com/datasetdetail/149933} \\
        WebCPM~\citep{qin2023webcpm} & zh & 1,602 & \url{https://github.com/thunlp/WebCPM}  \\
        arxiv\_qa~\citep{titanmldata2023arxivqa} & en & 17,927 & \url{https://huggingface.co/datasets/TitanMLData/arxiv_qa} \\
        aya\_dataset~\citep{singh2024aya} & ml & 26,292 & \url{https://huggingface.co/datasets/CohereLabs/aya_dataset}  \\
        cCOVID-News & zh & 4,727 & \url{https://www.datafountain.cn/competitions/424/datasets} \\
        cMedQA-V2.0~\citep{zhang2018multi} & zh & 88,109 & \url{https://huggingface.co/datasets/wangrongsheng/cMedQA-V2.0} \\
        ccnews~\citep{hamborg2017news} & en & 28,246 & \url{https://edoc.hu-berlin.de/items/ad915f2d-bb2c-4abd-887f-3d50bd3f2516} \\
        CMRC 2018~\citep{cui2019span} & zh & 9,753 & \url{https://huggingface.co/datasets/erhwenkuo/squad-cmrc2018-zhtw} \\
        cord19\_trec-covid~\citep{voorhees2021trec,wang2020cord} & en & 48,517 & \url{https://huggingface.co/datasets/irds/cord19_trec-covid} \\
        CQADupstack~\citep{hoogeveen2015cqadupstack} & en & 7,355 & \url{http://nlp.cis.unimelb.edu.au/resources/cqadupstack/} \\
        csl~\citep{li2022csl} & zh & 19,945 & \url{https://huggingface.co/datasets/neuclir/csl} \\
        dbpedia-entity~\citep{hasibi2017dbpedia} & en & 96,792 & \url{https://github.com/iai-group/DBpedia-Entity/} \\
        dureader~\citep{he2018dureader} & zh & 79,229 & \url{https://huggingface.co/datasets/sentence-transformers/dureader} \\
        dureader-checklist~\citep{tang2021dureader_robust} & zh & 97,764 & \url{https://huggingface.co/datasets/luozhouyang/dureader} \\
        esci~\citep{reddy2022shopping} & en & 26,043 & \url{https://huggingface.co/datasets/tasksource/esci} \\
        fever~\citep{thorne2018fever} & en & 87,216 & \url{https://huggingface.co/datasets/maxzoech/fever} \\
        fiqa~\citep{maia201818} & en & 4,689 & \url{https://huggingface.co/datasets/irds/beir_fiqa_train} \\
        hotpot\_qa~\citep{yang2018hotpotqa} & en & 150,153 & \url{https://huggingface.co/datasets/hotpotqa/hotpot_qa} \\
        law-gpt~\citep{LAWGPT-zh} & zh & 500 & \url{https://huggingface.co/datasets/sentence-transformers/law-gpt} \\
        lawzhidao~\citep{falv5983} & zh & 6,784 & \url{https://www.heywhale.com/mw/dataset/5e953ca8e7ec38002d02fca7/content} \\
        lima-chinese~\citep{zhou2023lima} & zh & 1,991 & \url{https://huggingface.co/datasets/paralym/lima-chinese} \\
        miracl~\citep{zhang2023miracl} & ml & 128,174 & \url{https://huggingface.co/datasets/sentence-transformers/miracl} \\
        mmarco-chinese~\citep{bonifacio2021mmarco} & zh & 379,870 & \url{https://huggingface.co/datasets/unicamp-dl/mmarco} \\
        mr-tydi~\citep{zhang2021mr} & ml & 96,124 & \url{https://huggingface.co/datasets/castorini/mr-tydi} \\
        msmarco-passage~\citep{bajaj2016ms} & en & 174,190 & \url{https://huggingface.co/datasets/Tevatron/msmarco-passage} \\
        msmarco-v2~\citep{bajaj2016ms} & en & 258,617 & \url{https://huggingface.co/datasets/mteb/msmarco-v2} \\
        nfcorpus~\citep{boteva2016full} & en & 10,471 & \url{https://huggingface.co/datasets/BeIR/nfcorpus-generated-queries} \\
        rag-dataset-12000~\citep{neuralbridge2023ragdataset12000} & en & 9,272 &  \url{https://huggingface.co/datasets/neural-bridge/rag-dataset-12000} \\
        retrieval\_data\_llm\_infgrad~\citep{infgrad2024retrievaldata} & zh & 32,551 & \url{https://huggingface.co/datasets/infgrad/retrieval_data_llm} \\
        scifact~\citep{wadden2020fact} & en & 794 & \url{https://huggingface.co/datasets/Tevatron/scifact} \\
        squad\_v2~\citep{rajpurkar2018know,rajpurkar2016squad} & en & 125,816 & \url{https://huggingface.co/datasets/rajpurkar/squad_v2} \\
        triviaqa~\citep{joshi2017triviaqa} & en & 44,442 & \url{https://huggingface.co/datasets/multi-train/emb-triviaqa-train} \\
        webgpt\_comparisons~\citep{nakano2021webgpt} & en & 18,924 & \url{https://huggingface.co/datasets/openai/webgpt_comparisons} \\
        webqa~\citep{li2016dataset} & zh & 4,988 & \url{https://huggingface.co/datasets/suolyer/webqa} \\
        wikipedia-nq~\citep{kwiatkowski2019natural} & en  & 56,377 & \url{https://huggingface.co/datasets/Tevatron/wikipedia-nq} \\
        yahoo-answers~\citep{sentencetransformers2024yahooanswers} & en & 21,724 & \url{https://huggingface.co/datasets/sentence-transformers/yahoo-answers} \\
        quora-question-pairs~\citep{iyer2017first} & en & 83,098 & \url{https://huggingface.co/datasets/AlekseyKorshuk/quora-question-pairs} \\
        arguana~\citep{wachsmuth2018retrieval} & en & 4,065 & \url{https://zenodo.org/records/3973258} \\
        mldr~\citep{chen2024bge} & ml & 41,434 & \url{https://huggingface.co/datasets/sentence-transformers/mldr}\\
        \rowcolor{gray!10}
        \multicolumn{4}{c}{\textbf{BGE-M3-data}} \\
        MSMARCO~\citep{bajaj2016ms} & en & 476,968 & \url{https://huggingface.co/datasets/Shitao/bge-m3-data} \\
        NQ~\citep{kwiatkowski2019natural} & en & 58,554 & \url{https://huggingface.co/datasets/Shitao/bge-m3-data} \\
        HotpotQA~\citep{yang2018hotpotqa} & en & 84,228 & \url{https://huggingface.co/datasets/Shitao/bge-m3-data} \\
        Trivia~\citep{joshi2017triviaqa} & en & 60,283 & \url{https://huggingface.co/datasets/Shitao/bge-m3-data} \\
        \addlinespace[2pt]
        \bottomrule
    \end{tabular}}
\end{table*}

\begin{landscape}
\begin{table*}[p]
    \centering
    \caption{Test-time compute scaling results on MIRACL~\citep{zhang2023miracl}, measured by nDCG@10 across all 18 languages. ``Nano'', ``Small'', and ``Large'' denote \model\texttt{-Nano}, \model\texttt{-Small}, and \model\texttt{-Large}, respectively. The reranking and post-ranking columns specify the number of documents ($\mathrm{d}$) and compression ratios ($r$) used at each stage. Cost denotes the estimated relative online computation cost, with  \model\texttt{-Nano} at $r=4$ normalized to 1.0x. All other evaluation settings follow those in Table~\ref{tab:main_beir_results}.}
    \label{tab:test_time_compute_miracl}
    \renewcommand{\arraystretch}{1.25}
    \setlength{\tabcolsep}{2pt}
    \resizebox{\linewidth}{!}{
    \begin{tabular}{c c c c c c*{18}{c}}
        \toprule
        \multirow{2}{*}{\textbf{Models}}
        & \multirow{2}{*}{\textbf{Setting}}
        & \multirow{2}{*}{\textbf{Cost}}
        & \multirow{2}{*}{\textbf{Avg.}}
        & \multicolumn{2}{c}{\textbf{Test-Time Compute}}
        & \multicolumn{18}{c}{\textbf{MIRACL}} \\
        \cmidrule(lr){5-6}\cmidrule(lr){7-24}
        & & & & \textbf{Reranking} & \textbf{Post-ranking}
        & \textbf{ar} & \textbf{bn} & \textbf{de} & \textbf{en} & \textbf{es} & \textbf{fa}
        & \textbf{fi} & \textbf{fr} & \textbf{hi} & \textbf{id} & \textbf{ja} & \textbf{ko}
        & \textbf{ru} & \textbf{sw} & \textbf{te} & \textbf{th} & \textbf{yo} & \textbf{zh} \\
        \midrule

        & (a) & 1.89x
        & \textbf{71.63}
        & $100\mathrm{d} \xrightarrow{r=2} 100\mathrm{d}$
        & N/A
        & \textbf{79.01} & 83.20 & \underline{58.56} & \textbf{65.74} & \textbf{62.47} & \textbf{61.58} & \textbf{79.82} & \textbf{62.32} & \textbf{62.47}
        & \textbf{59.32} & \textbf{74.32} & \textbf{73.99} & \textbf{74.04} & \textbf{76.10} & \textbf{86.06} & 82.41 & \underline{82.73} & \textbf{65.24} \\

        & (b) & 1.0x
        & 71.11
        & $100\mathrm{d} \xrightarrow{r=4} 100\mathrm{d}$
        & N/A
        & 78.76 & 83.14 & 57.45 & 65.00 & 61.54 & 60.81 & 79.30 & 61.03 & 61.95
        & 59.04 & 74.25 & 72.76 & 73.77 & 75.74 & 85.78 & {82.47} & \textbf{83.00} & 64.14 \\

        \rowcolor{gray!12}\cellcolor{white}
        & (c) & 0.6x
        & \underline{71.35}
        & $100\mathrm{d} \xrightarrow{r=32} 100\mathrm{d}$
        & $20\mathrm{d} \xrightarrow{r=2} 20\mathrm{d}$
        & \underline{78.83} & \underline{83.39} & \textbf{58.59} & \underline{65.25} & \underline{61.90} & \underline{61.31} & \underline{79.68} & 61.97 & \underline{62.37}
        & \underline{59.11} & \underline{74.31} & \underline{73.64} & \underline{73.80} & 75.78 & \underline{86.02} & \textbf{82.55} & 81.81 & 64.02 \\

        \rowcolor{gray!12}\cellcolor{white}
        \multirow{-4}{*}{\texttt{Nano}}
        & (d) & 0.54x
        & 71.18
        & $100\mathrm{d} \xrightarrow{r=64} 100\mathrm{d}$
        & $20\mathrm{d} \xrightarrow{r=2} 20\mathrm{d}$
        & 78.82 & \textbf{83.45} & 57.46 & 65.14 & 61.43 & 61.29 & 79.44 & \underline{62.23} & 62.21
        & 58.82 & 73.97 & 73.17 & 73.75 & \underline{75.85} & 85.92 & \textbf{82.55} & 81.64 & \underline{64.17} \\

        \midrule

        & (a) & 8.65x
        & \textbf{74.34}
        & $100\mathrm{d} \xrightarrow{r=2} 100\mathrm{d}$
        & N/A
        & \textbf{81.29} & \underline{85.66} & 64.79 & \textbf{66.85} & \textbf{64.09} & 63.55 & \underline{82.26} & \textbf{65.47} & \underline{66.02}
        & \underline{59.46} & \textbf{79.28} & \textbf{73.40} & \textbf{78.19} & 80.05 & 87.48 & 85.50 & \textbf{86.44} & \textbf{68.37} \\

        & (b) & 4.58x
        & 74.06
        & $100\mathrm{d} \xrightarrow{r=4} 100\mathrm{d}$
        & N/A
        & 80.92 & \textbf{85.69} & 64.22 & 66.45 & 63.53 & 62.95 & 82.09 & \underline{65.29} & 65.65
        & \textbf{59.48} & 78.91 & 72.84 & \underline{78.13} & \underline{80.12} & 87.39 & \underline{85.53} & \underline{86.34} & \underline{67.56} \\

        \rowcolor{gray!12}\cellcolor{white}
        & (c) & 2.75x
        & \underline{74.18}
        & $100\mathrm{d} \xrightarrow{r=32} 100\mathrm{d}$
        & $20\mathrm{d} \xrightarrow{r=2} 20\mathrm{d}$
        & \underline{81.22} & 85.48 & \textbf{65.21} & \underline{66.59} & \underline{63.54} & \underline{63.71} & \textbf{82.27} & 64.81 & 65.94
        & 59.40 & \underline{79.05} & \underline{73.04} & 78.08 & \textbf{80.16} & \textbf{87.59} & \textbf{85.61} & 86.03 & 67.43 \\

        \rowcolor{gray!12}\cellcolor{white}
        \multirow{-4}{*}{\texttt{Small}}
        & (d) & 2.49x
        & 74.08
        & $100\mathrm{d} \xrightarrow{r=64} 100\mathrm{d}$
        & $20\mathrm{d} \xrightarrow{r=2} 20\mathrm{d}$
        & 81.16 & 85.38 & \underline{65.20} & 66.35 & 63.46 & \textbf{63.79} & 82.06 & 64.90 & \textbf{66.32}
        & 59.17 & 78.99 & 72.80 & 77.99 & 79.84 & \underline{87.54} & 85.46 & 86.03 & 67.00 \\

        \midrule

        & (a) & 55.05x
        & \textbf{75.12}
        & $100\mathrm{d} \xrightarrow{r=2} 100\mathrm{d}$
        & N/A
        & \underline{82.56} & \underline{85.96} & \textbf{66.70} & \underline{67.64} & \underline{63.97} & \textbf{65.24} & \underline{83.37} & \underline{65.90} & 66.45
        & \textbf{59.47} & {80.37} & \textbf{75.11} & \textbf{79.71} & \textbf{81.12} & \textbf{87.96} & 85.89 & \textbf{87.26} & 67.45 \\

        & (b) & 29.14x
        & 74.92
        & $100\mathrm{d} \xrightarrow{r=4} 100\mathrm{d}$
        & N/A
        & \underline{82.56} & 85.80 & 65.75 & \textbf{67.65} & 63.76 & 65.09 & \textbf{83.38} & \textbf{65.99} & 66.30
        & 59.17 & 80.27 & 74.45 & 79.51 & 80.56 & 87.53 & \textbf{85.96} & \underline{87.13} & \textbf{67.71} \\

        \rowcolor{gray!12}\cellcolor{white}
        & (c) & 17.49x
        & \underline{75.05}
        & $100\mathrm{d} \xrightarrow{r=32} 100\mathrm{d}$
        & $20\mathrm{d} \xrightarrow{r=2} 20\mathrm{d}$
        & 82.53 & \textbf{86.10} & \underline{66.54} & 67.45 & \textbf{63.98} & 65.09 & 83.30 & 65.88 & \underline{66.57}
        & \underline{59.45} & \textbf{80.48} & 74.64 & \underline{79.66} & \underline{80.94} & \underline{87.91} & \underline{85.94} & 86.89 & 67.63 \\

        \rowcolor{gray!12}\cellcolor{white}
        \multirow{-4}{*}{\texttt{Large}}
        & (d) & 15.87x
        & 75.01
        & $100\mathrm{d} \xrightarrow{r=64} 100\mathrm{d}$
        & $20\mathrm{d} \xrightarrow{r=2} 20\mathrm{d}$
        & \textbf{82.59} & 85.93 & \underline{66.54} & 67.31 & 63.90 & \underline{65.15} & 83.22 & \underline{65.90} & \textbf{66.60}
        & 59.31 & \textbf{80.48} & \underline{74.70} & 79.47 & 80.85 & \underline{87.91} & 85.89 & 86.88 & \underline{67.65} \\

        \bottomrule
    \end{tabular}}
\end{table*}
\end{landscape}

\begin{landscape}
\begin{table*}[p]
    \centering
    \caption{BEIR~\citep{thakur2021beir} reranking results under different Matryoshka embedding pooling compression ratios $r$, measured by nDCG@10. The 13 task abbreviations follow Table~\ref{tab:inst_beir}. ``CR'' denotes the compression ratio, where a larger value indicates more compressed passage representations. Cost denotes the estimated relative serving computation cost derived from the time complexity analysis in Appendix~\ref{app:time_complexity}. We use $|q|=32$, $n=1024$, $K=1$, and the corresponding compression ratio $r$; obtain $L$ and $d$ from Table~\ref{tab:model-spec}; and normalize the cost of \texttt{Nano} at $r=128$ to 1.0x. Avg. denotes the average performance across the 13 tasks.} 
    \label{tab:beir_mep_results}
    \renewcommand\arraystretch{1.15}
    \setlength{\tabcolsep}{3pt}
    \resizebox{\linewidth}{!}{\begin{tabular}{lccc|ccccccccccccc}
        \toprule
        \multicolumn{1}{l}{\textbf{Model}} &
        \multicolumn{1}{c}{\textbf{CR}} &
        \multicolumn{1}{c}{\textbf{Cost}} &
        \multicolumn{1}{c|}{\textbf{Avg.}} &
        \multicolumn{1}{c}{\textbf{AA}} &
        \multicolumn{1}{c}{\textbf{CF}} &
        \multicolumn{1}{c}{\textbf{CQA}} &
        \multicolumn{1}{c}{\textbf{DB}} &
        \multicolumn{1}{c}{\textbf{FV}} &
        \multicolumn{1}{c}{\textbf{FQA}} &
        \multicolumn{1}{c}{\textbf{HQA}} &
        \multicolumn{1}{c}{\textbf{MS}} &
        \multicolumn{1}{c}{\textbf{NF}} &
        \multicolumn{1}{c}{\textbf{NQ}} &
        \multicolumn{1}{c}{\textbf{QR}} &
        \multicolumn{1}{c}{\textbf{SD}} &
        \multicolumn{1}{c}{\textbf{SF}} \\
        \midrule
        \multirow{7}{*}{\model\texttt{-Nano}} & 2  & 13.6x & 58.75 & 63.87 & 43.23 & 41.08 & 48.92 & 93.13 & 47.25 & 82.80 & 46.17 & 38.42 & 69.57 & 90.11 & 21.04 & 78.17 \\ 
                                                & 4  & 7.2x & 58.54 & 63.97 & 43.25 & 40.84 & 47.93 & 92.95 & 47.09 & 82.58 & 45.64 & 38.72 & 69.30 & 89.92 & 20.80 & 78.03 \\
                                                & 8  & 4.0x & 58.17 & 63.46 & 43.17 & 40.52 & 47.51 & 92.90 & 46.60 & 82.13 & 45.42 & 38.13 & 68.66 & 89.77 & 20.64 & 77.30 \\
                                                & 16 & 2.4x & 57.47 & 62.63 & 42.35 & 40.21 & 46.40 & 92.85 & 45.27 & 81.29 & 44.36 & 37.72 & 67.61 & 89.68 & 20.30 & 76.50 \\
                                                & 32 & 1.6x & 56.74 & 61.31 & 42.16 & 39.36 & 44.80 & 92.45 & 44.44 & 80.17 & 43.82 & 37.28 & 66.70 & 89.64 & 20.12 & 75.35 \\
                                                & 64 & 1.2x & 55.88 & 60.13 & 41.38 & 38.73 & 44.07 & 91.74 & 43.05 & 78.63 & 43.07 & 36.91 & 64.88 & 89.62 & 19.71 & 74.53 \\
                                                & 128 & 1.0x & 55.11 & 58.42 & 40.96 & 38.28 & 43.11 & 91.23 & 42.22 & 77.26 & 42.90 & 36.23 & 63.73 & 89.62 & 19.32 & 73.16 \\
        \midrule
        \multirow{7}{*}{\model\texttt{-Small}} & 2  & 62.3x & 61.17 & 68.79 & 45.98 & 47.35 & 50.30 & 93.50 & 54.08 & 83.74 & 47.33 & 39.95 & 73.09 & 90.85 & 22.05 & 78.26 \\
                                                 & 4  & 33.0x & 61.07 & 69.43 & 45.63 & 47.06 & 49.77 & 93.36 & 53.62 & 83.55 & 46.90 & 39.89 & 73.14 & 90.77 & 22.15 & 78.72 \\
                                                 & 8  & 18.3x & 60.74 & 69.08 & 45.20 & 46.72 & 49.25 & 93.28 & 53.20 & 83.21 & 46.82 & 39.62 & 72.59 & 90.55 & 21.95 & 78.16 \\
                                                 & 16 & 11.0x & 60.25 & 68.52 & 44.78 & 46.07 & 48.55 & 93.08 & 52.63 & 82.73 & 46.13 & 39.31 & 71.76 & 90.49 & 21.67 & 77.49 \\
                                                 & 32 & 7.3x & 59.54 & 67.26 & 44.58 & 45.15 & 46.60 & 92.80 & 51.73 & 81.92 & 45.42 & 39.01 & 70.50 & 90.39 & 21.34 & 77.29 \\
                                                 & 64 & 5.5x & 58.84 & 66.76 & 44.15 & 44.13 & 45.82 & 92.30 & 50.73 & 80.71 & 44.54 & 38.71 & 69.46 & 90.39 & 21.01 & 76.20 \\
                                                 & 128 & 4.6x & 58.20 & 65.88 & 44.07 & 43.37 & 44.88 & 91.65 & 50.00 & 79.70 & 44.53 & 37.86 & 68.54 & 90.39 & 20.78 & 74.98 \\
        \midrule
        \multirow{7}{*}{\model\texttt{-Large}} & 2  & 396.3x & 63.69 & 72.97 & 47.80 & 52.76 & 53.29 & 94.26 & 59.31 & 84.39 & 47.72 & 42.26 & 76.06 & 90.76 & 24.91 & 81.49 \\
                                                 & 4  & 209.8x & 63.53 & 72.90 & 47.94 & 52.36 & 52.73 & 94.22 & 59.15 & 84.23 & 47.38 & 42.35 & 75.61 & 90.72 & 24.89 & 81.42 \\
                                                 & 8  & 116.6x & 63.15 & 71.88 & 47.70 & 52.02 & 52.56 & 94.16 & 58.71 & 83.98 & 47.34 & 42.09 & 75.31 & 90.55 & 24.89 & 79.78 \\
                                                 & 16 & 69.9x & 62.70 & 71.48 & 47.63 & 51.43 & 51.74 & 93.92 & 58.57 & 83.51 & 46.73 & 41.50 & 74.16 & 90.51 & 24.61 & 79.37 \\
                                                 & 32 & 46.6x & 62.18 & 69.97 & 47.59 & 50.64 & 50.95 & 93.56 & 57.31 & 82.94 & 46.18 & 41.39 & 73.28 & 90.45 & 24.54 & 79.49 \\
                                                 & 64 & 35.0x & 61.60 & 69.81 & 47.22 & 49.76 & 50.17 & 93.22 & 56.92 & 82.04 & 45.70 & 40.81 & 72.52 & 90.46 & 24.17 & 77.95 \\
                                                 & 128 & 29.1x & 61.23 & 68.76 & 46.96 & 49.37 & 49.64 & 92.82 & 56.34 & 81.33 & 45.72 & 40.69 & 71.47 & 90.46 & 24.28 & 78.23 \\
        \bottomrule
    \end{tabular}}
\end{table*}
\end{landscape}

\begin{landscape}
\begin{table*}[p]
    \centering
    \caption{MIRACL~\citep{zhang2023miracl} reranking results under different Matryoshka embedding pooling compression ratios $r$, measured by nDCG@10. All 18 language abbreviations follow Table~\ref{tab:inst_miracle}. ``CR'' denotes the compression ratio, where a larger value indicates more compressed passage representations. Cost denotes the estimated relative serving computation cost derived from the time complexity analysis in Appendix~\ref{app:time_complexity}. We use $|q|=32$, $n=1024$, $K=1$, and the corresponding compression ratio $r$; obtain $L$ and $d$ from Table~\ref{tab:model-spec}; and normalize the cost of \texttt{Nano} at $r=128$ to 1.0x. Avg. denotes the average performance across the 18 languages.}
    \label{tab:miracle_mep_results}
    \renewcommand\arraystretch{1.15}
    \setlength{\tabcolsep}{3pt}
    \resizebox{\linewidth}{!}{\begin{tabular}{lccc|cccccccccccccccccc}
        \toprule
        \multicolumn{1}{l}{\textbf{Model}} &
        \multicolumn{1}{c}{\textbf{CR}} &
        \multicolumn{1}{c}{\textbf{Cost}} &
        \multicolumn{1}{c|}{\textbf{Avg.}} &
        \multicolumn{1}{c}{\textbf{ar}} &
        \multicolumn{1}{c}{\textbf{bn}} &
        \multicolumn{1}{c}{\textbf{de}} &
        \multicolumn{1}{c}{\textbf{en}} &
        \multicolumn{1}{c}{\textbf{es}} &
        \multicolumn{1}{c}{\textbf{fa}} &
        \multicolumn{1}{c}{\textbf{fi}} &
        \multicolumn{1}{c}{\textbf{fr}} &
        \multicolumn{1}{c}{\textbf{hi}} &
        \multicolumn{1}{c}{\textbf{id}} &
        \multicolumn{1}{c}{\textbf{ja}} &
        \multicolumn{1}{c}{\textbf{ko}} &
        \multicolumn{1}{c}{\textbf{ru}} &
        \multicolumn{1}{c}{\textbf{sw}} &
        \multicolumn{1}{c}{\textbf{te}} &
        \multicolumn{1}{c}{\textbf{th}} &
        \multicolumn{1}{c}{\textbf{yo}} &
        \multicolumn{1}{c}{\textbf{zh}} \\
        \midrule
        \multirow{7}{*}{\model\texttt{-N}} & 2  & 13.6x & 71.63 & 79.01 & 83.20 & 58.56 & 65.74 & 62.47 & 61.58 & 79.82 & 62.32 & 62.47 & 59.32 & 74.32 & 73.99 & 74.04 & 76.10 & 86.06 & 82.41 & 82.73 & 65.24 \\
                                                & 4  & 7.2x & 71.11 & 78.76 & 83.14 & 57.45 & 65.00 & 61.54 & 60.81 & 79.30 & 61.03 & 61.95 & 59.04 & 74.25 & 72.76 & 73.77 & 75.74 & 85.78 & 82.47 & 83.00 & 64.14 \\
                                                & 8  & 4.0x & 70.38 & 78.23 & 82.94 & 57.16 & 63.88 & 60.98 & 59.52 & 78.61 & 60.92 & 61.33 & 58.29 & 74.09 & 71.80 & 72.50 & 74.78 & 85.40 & 81.82 & 80.78 & 63.79 \\
                                                & 16 & 2.4x & 69.38 & 77.56 & 81.76 & 56.04 & 62.48 & 59.21 & 58.54 & 77.65 & 60.01 & 60.77 & 57.13 & 72.82 & 71.62 & 71.11 & 73.81 & 84.80 & 81.58 & 79.86 & 62.07 \\
                                                & 32 & 1.6x & 68.02 & 76.56 & 80.08 & 52.80 & 61.04 & 57.67 & 56.88 & 76.04 & 59.22 & 59.78 & 56.40 & 71.81 & 69.69 & 70.07 & 72.68 & 84.72 & 80.98 & 77.50 & 60.53 \\
                                                & 64 & 1.2x & 66.85 & 75.71 & 77.93 & 52.47 & 59.60 & 56.36 & 55.60 & 74.94 & 59.14 & 58.30 & 55.85 & 70.10 & 68.17 & 68.37 & 71.63 & 82.99 & 79.92 & 76.88 & 59.40 \\
                                                & 128 & 1.0x & 66.01 & 74.63 & 76.67 & 51.65 & 59.31 & 55.75 & 54.98 & 73.22 & 59.07 & 57.41 & 54.78 & 68.44 & 66.38 & 67.47 & 71.29 & 83.05 & 78.70 & 75.65 & 59.75 \\
        \midrule
        \multirow{7}{*}{\model\texttt{-S}} & 2  & 62.3x & 74.34 & 81.29 & 85.66 & 64.79 & 66.85 & 64.09 & 63.55 & 82.26 & 65.47 & 66.02 & 59.46 & 79.28 & 73.40 & 78.19 & 80.05 & 87.48 & 85.50 & 86.44 & 68.37 \\
                                                 & 4  & 33.0x & 74.06 & 80.92 & 85.69 & 64.22 & 66.45 & 63.53 & 62.95 & 82.09 & 65.29 & 65.65 & 59.48 & 78.91 & 72.84 & 78.13 & 80.12 & 87.39 & 85.53 & 86.34 & 67.56 \\
                                                 & 8  & 18.3x & 73.61 & 80.46 & 85.48 & 63.64 & 65.86 & 62.78 & 63.17 & 81.62 & 64.39 & 64.75 & 59.06 & 78.96 & 72.08 & 77.90 & 79.21 & 87.27 & 85.43 & 85.93 & 66.94 \\
                                                 & 16 & 11.0x & 72.60 & 80.20 & 85.02 & 62.55 & 64.69 & 61.45 & 62.06 & 81.46 & 64.00 & 63.45 & 58.30 & 77.79 & 71.88 & 76.36 & 77.82 & 87.05 & 85.22 & 82.56 & 65.03 \\
                                                 & 32 & 7.3x & 71.81 & 79.70 & 83.90 & 62.24 & 62.90 & 60.76 & 61.54 & 81.05 & 63.01 & 62.92 & 57.44 & 76.90 & 71.27 & 75.55 & 77.74 & 86.59 & 84.39 & 80.80 & 63.92 \\
                                                 & 64 & 5.5x & 70.82 & 79.05 & 82.59 & 61.91 & 61.82 & 58.77 & 61.13 & 79.51 & 61.86 & 61.92 & 56.61 & 76.00 & 70.21 & 74.59 & 77.27 & 86.23 & 83.46 & 79.78 & 62.04 \\
                                                 & 128 & 4.6x & 70.16 & 78.32 & 81.26 & 60.87 & 61.19 & 58.50 & 60.87 & 78.96 & 61.81 & 60.53 & 56.86 & 75.64 & 68.75 & 73.62 & 76.97 & 84.92 & 82.83 & 79.09 & 61.84 \\
        \midrule
        \multirow{7}{*}{\model\texttt{-L}} & 2  & 396.3x & 75.12 & 82.56 & 85.96 & 66.70 & 67.64 & 63.97 & 65.24 & 83.37 & 65.90 & 66.45 & 59.47 & 80.37 & 75.11 & 79.71 & 81.12 & 87.96 & 85.89 & 87.26 & 67.45 \\
                                                 & 4  & 209.8x & 74.92 & 82.56 & 85.80 & 65.75 & 67.65 & 63.76 & 65.09 & 83.38 & 65.99 & 66.30 & 59.17 & 80.27 & 74.45 & 79.51 & 80.56 & 87.53 & 85.96 & 87.13 & 67.71 \\
                                                 & 8  & 116.6x & 74.50 & 82.34 & 85.37 & 65.45 & 67.04 & 63.07 & 64.18 & 82.95 & 65.12 & 65.50 & 58.79 & 79.87 & 73.70 & 78.93 & 80.41 & 87.72 & 86.01 & 87.34 & 67.23 \\
                                                 & 16 & 69.9x & 73.85 & 82.04 & 85.29 & 64.42 & 66.12 & 62.27 & 63.86 & 82.68 & 63.72 & 64.96 & 58.32 & 79.53 & 72.59 & 78.24 & 79.60 & 87.26 & 85.79 & 85.68 & 66.94 \\
                                                 & 32 & 46.6x & 73.26 & 81.83 & 84.31 & 63.14 & 64.97 & 61.02 & 63.17 & 82.18 & 64.53 & 63.73 & 58.06 & 78.62 & 71.97 & 77.88 & 79.71 & 86.70 & 85.15 & 85.63 & 66.10 \\
                                                 & 64 & 35.0x & 72.46 & 81.55 & 82.56 & 63.11 & 63.83 & 60.14 & 62.81 & 81.78 & 62.42 & 62.77 & 56.77 & 77.60 & 70.83 & 76.51 & 79.43 & 86.61 & 84.83 & 85.68 & 65.09 \\
                                                 & 128 & 29.1x & 71.97 & 80.90 & 82.70 & 62.57 & 63.37 & 59.54 & 63.02 & 81.14 & 62.32 & 61.69 & 56.37 & 76.48 & 70.59 & 76.01 & 78.87 & 86.43 & 84.02 & 84.75 & 64.73 \\
        \bottomrule
    \end{tabular}}
\end{table*}
\end{landscape}

\section{Case Study}
\label{app:case}
\begin{figure}[ht]
    \centering
    \begin{subfigure}[t]{0.475\linewidth}
        \vspace{0pt}
        \centering
        \includegraphics[height=0.77\linewidth,keepaspectratio,valign=t]{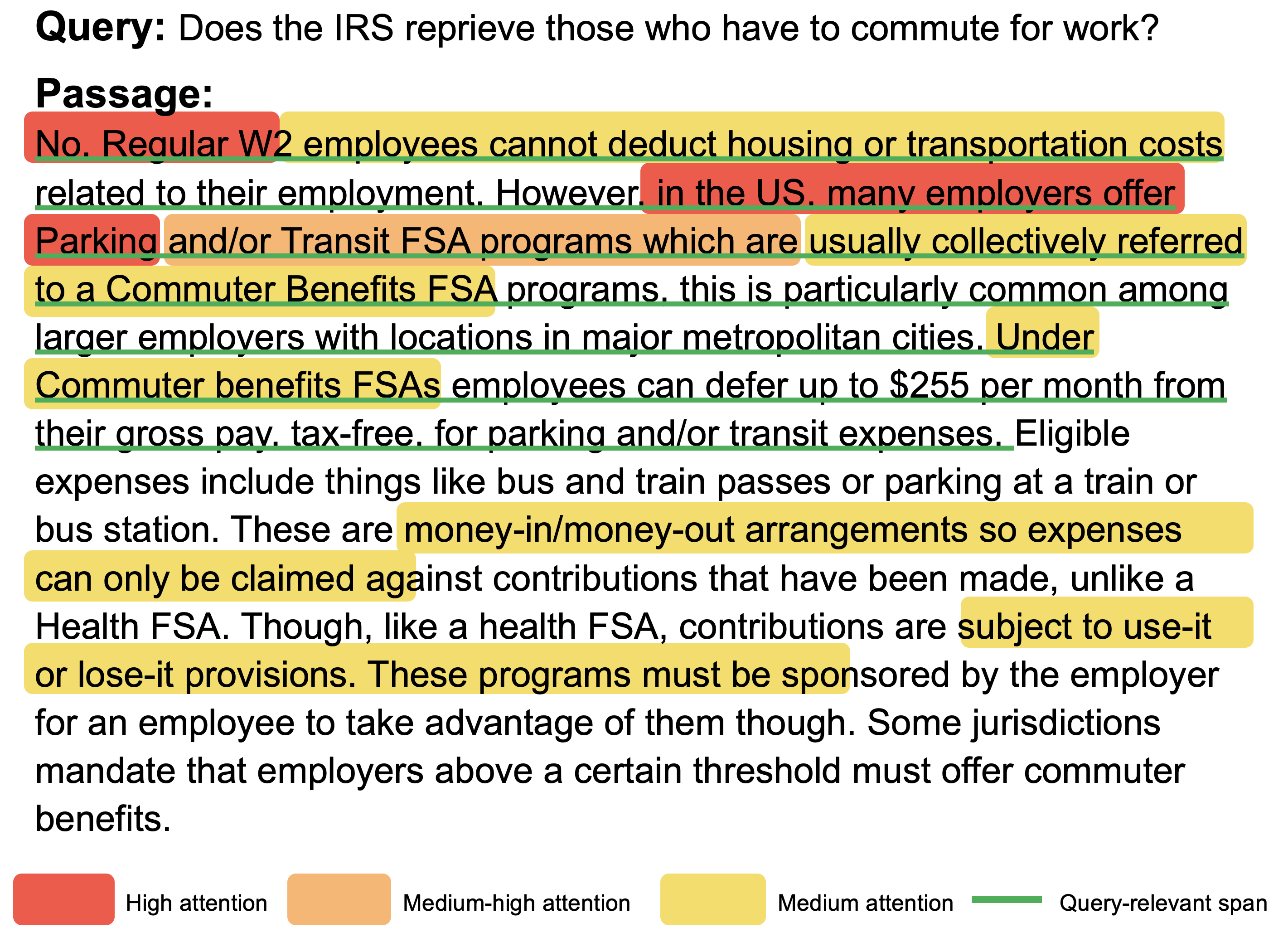}
        \caption{IRS Commuter Benefit}
        \label{fig:case_attention_1}
    \end{subfigure}
    \hfill
    \begin{subfigure}[t]{0.475\linewidth}
        \vspace{0pt}
        \centering
        \includegraphics[height=0.77\linewidth,keepaspectratio,valign=t]{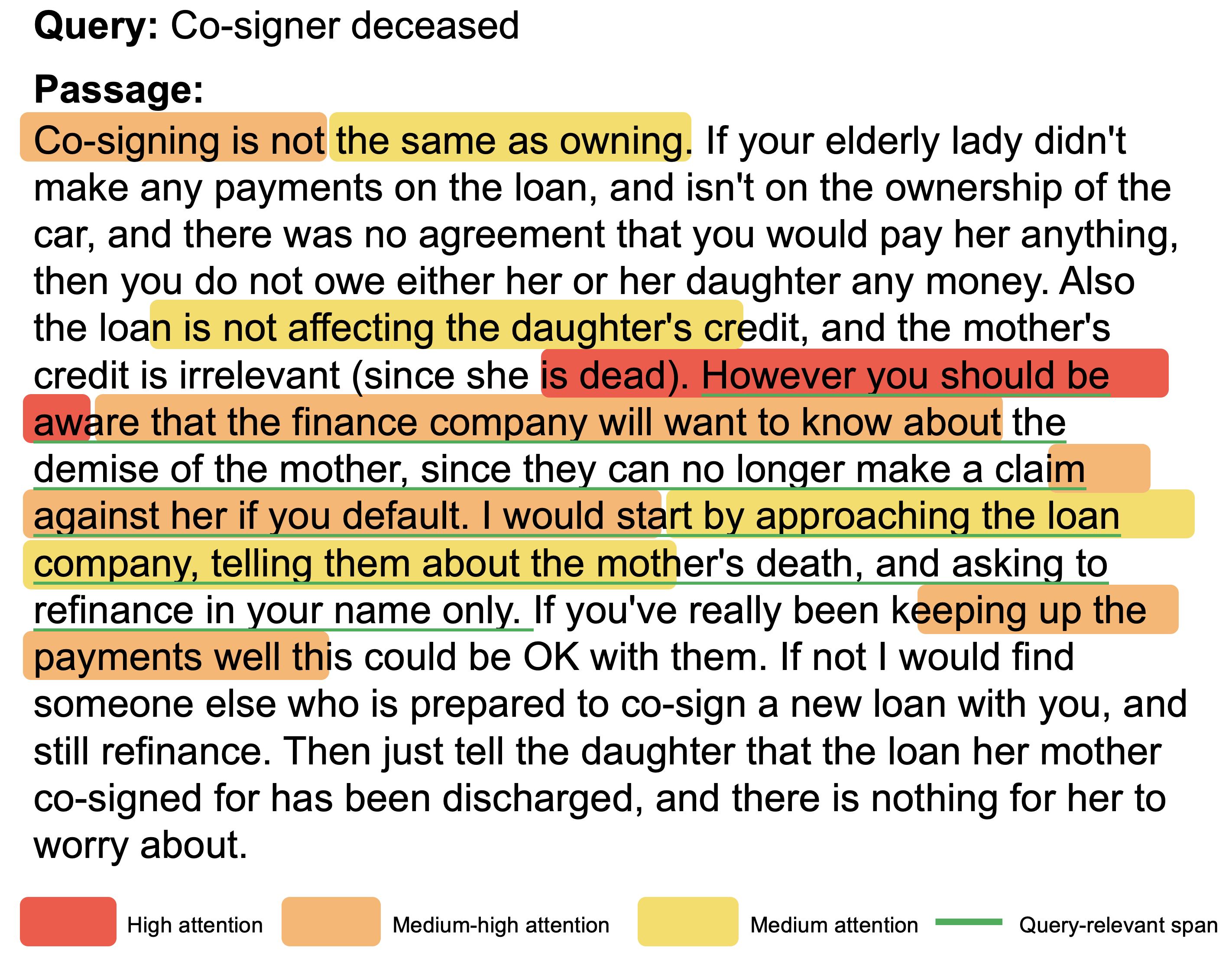}
        \caption{Co-signer Liability After Death}
        \label{fig:case_attention_2}
    \end{subfigure}
    \caption{Decoder-to-encoder attention in \model\texttt{-Small} at a compression ratio of $r=8$ for two query--positive-passage pairs from the FiQA test set in BEIR~\citep{thakur2021beir}.
    Attention weights are extracted from the first prediction position (\textit{i.e.,}
    the last non-padding token in the decoder input).
    We first average attention weights across heads within each decoder layer. For each compressed passage position, we then average the six largest layer-wise attention weights and map the resulting score to the corresponding chunk in the original passage.
    Colors indicate attention strength, and green underlines mark query-relevant spans.
    The resulting high-attention chunks overlap with the query-relevant spans.
    }
    \label{fig:case_attention}
\end{figure}

To examine whether the \textbf{compact passage representations} produced by Matryoshka
embedding pooling retain sufficient information for \textbf{expressive
relevance modeling} (\S\ref{sec:introduction}), we conduct a qualitative analysis of two
query--positive-passage pairs from the FiQA test set in BEIR~\citep{thakur2021beir}.
We use \model\texttt{-Small} with a compression ratio of $r=8$, such that each
position in the compressed passage representation summarizes up to eight
consecutive passage tokens.
This allows us to assess both whether query-relevant information is preserved
after MEP and whether the decoder can recover and use it for relevance
modeling.
\noindent\textbf{Visualization setup.}
We extract decoder-to-encoder attention at the first prediction position,
which corresponds to the last non-padding token in the decoder input. The hidden
state at this position produces the logits for the \texttt{yes} and
\texttt{no} labels, which determine the relevance score, as illustrated in
Figure~\ref{fig:model_framework}.
We first average attention weights across heads within each decoder layer. For each compressed passage position, we then select and average the six largest layer-wise attention weights.
The resulting score is mapped back to the
corresponding chunk in the original passage.
This aggregation highlights strong attention signals while reducing layer-specific
variation.
\noindent\textbf{Qualitative observations.}
Figure~\ref{fig:case_attention} uses color intensity to indicate attention
strength and green underlines to mark query-relevant spans.
In the first example (Figure~\ref{fig:case_attention_1}), the query concerns
tax relief for commuting expenses.
The strongest attention falls on chunks stating that regular W-2 employees cannot deduct transportation costs, as well as those describing employer-sponsored parking and transit FSA programs as a tax-advantaged alternative.
In the second example (Figure~\ref{fig:case_attention_2}), which concerns a deceased co-signer, the strongest attention falls on chunks describing the co-signer's death and the need to notify the finance company. The highlighted evidence also includes the recommendation to refinance the loan.
Together, these regions capture the key facts and practical guidance needed to
establish passage relevance.
\noindent\textbf{Key insights.}
Qualitatively, these examples illustrate the two advantages highlighted in \S\ref{sec:introduction}: \emph{compactness} and \emph{expressiveness}.
MEP compresses the precomputed passage representations by a factor of eight, substantially reducing storage requirements while retaining query-relevant evidence.
The decoder can then selectively attend to and integrate this evidence through merged self-attention and cross-attention, supporting expressive relevance modeling.
Overall, these visualizations provide qualitative evidence that compact passage representations retain information accessible to the decoder.

\end{document}